\documentclass{ecai}

\usepackage{latexsym}
\usepackage{amssymb}
\usepackage{amsmath}
\usepackage{amsthm}
\usepackage{bbm}
\usepackage{booktabs}
\usepackage{enumitem}
\usepackage{graphicx}
\usepackage{color}
\usepackage{natbib}
\usepackage{bm}
\usepackage{dsfont}
\usepackage{amsthm}
\usepackage{amssymb}
\usepackage{doi}
\usepackage{algorithm}
\usepackage{algpseudocode}
\usepackage{caption}
\usepackage{subcaption}
\newtheorem{theorem}{Theorem}

\newcommand{\BibTeX}{B\kern-.05em{\sc i\kern-.025em b}\kern-.08em\TeX}

\begin{document}


\begin{frontmatter}

\paperid{1125} 
\title{Identifying the Best Arm \\in the Presence of Global Environment Shifts}

\author{\fnms{Phurinut}~\snm{Srisawad}}
\author{\fnms{Juergen}~\snm{Branke}}
\author{\fnms{Long}~\snm{Tran-Thanh}} 

\address{University of Warwick, United Kingdom}

\begin{abstract}
This paper formulates a new Best-Arm Identification problem in the non-stationary stochastic bandits setting, where the means of all arms are shifted in the same way due to a global influence of the environment. The aim is to identify the unique best arm across environmental change given a fixed total budget. 
While this setting can be regarded as a special case of Adversarial Bandits or Corrupted Bandits, we demonstrate that existing solutions tailored to those settings do not fully utilise the nature of this global influence, and thus, do not work well in practice (despite their theoretical guarantees).
To overcome this issue, in this paper we develop a novel selection policy that is consistent and robust in dealing with global environmental shifts. We then propose an allocation policy, LinLUCB, which exploits information about global shifts across all arms in each environment. Empirical tests depict a significant improvement in our policies against other existing methods.

\end{abstract}

\end{frontmatter}


\section{Introduction}
A Multi-Armed Bandit (MAB) is an abstract concept of a decision problem, where a decision maker has a choice between different actions (arms), and selecting an action yields a stochastic reward.
Best-arm identification (BAI) \cite{lattimore2020bandit}, a sub-problem in MAB,  aims at identifying the best among all designs/arms without caring about accumulating regret during the exploration. The standard assumption for BAI is that each arm has an underlying reward distribution that is stationary. However, in practice, the reward distributions may change over time. One possible objective in such a non-stationary setting is to track the best arm or minimise the cumulative regret over time, adapting to the environmental changes, and this has been explored extensively in the literature \cite{garivier2011upper,allesiardo2015exp3,cavenaghi2021non,besson2022efficient, hartland2006multi}.  

In this paper, we consider a different problem of identifying the design that works best in expectation, across environments. We furthermore assume that environmental changes affect the underlying reward distributions of all arms in the same additive way, i.e., the mean of the reward distribution of arm $i$ in environment $j$ can be described as $\mu_{ij}=\mu_i+s_j$. We call this problem Multi-Armed Bandits in the Presence of Global Environment Shifts.
 
This is motivated by the fact that environmental changes often influence the reward of different actions in the same way. For example, when aiming to identify the best pricing strategy for a taxi application, customer's willingness to pay may differ from day to day, based on weather or specific events such as concerts or football matches, which may similarly influence the achievable profit for all considered pricing strategies. Or consider advertising on social media, where the click-through rate of different adverts may increase and decrease synchronously over time depending on external effects such as Christmas approaching, the product being discussed in a talk show, or a celebrity wearing the product. This is confirmed by recently published examples of daily empirical means from marketing experiments with uniformly collected data \citep{fiez2024best}. The trend of empirical means of all arms is positively correlated, and their relative gaps are quite well-behaved.

Identifying the best arm under such settings is challenging, because an arm evaluated more often in more favourable environments (positive offset $s_j$) may appear better than an arm that was evaluated more often in less favourable environments, even though the latter is better according to the underlying (environment-independent) expected reward $\mu_i$. Note that this setting can be considered as a special case of Corrupted Bandits \cite{zhong2021probabilistic} and Adversarial Bandits \cite{abbasi2018best} where the adversary can only corrupt rewards of all arms with the same constant $s_j$, and the agent can only observe \emph{when} the adversary attacks the bandits, and otherwise just receives the corrupted feedback. As such, in theory, existing robust BAI algorithms designed for adversarial environments can be applied to our setting.
However, as we will show later in this paper, those algorithms can be less efficient compared to a round-robin exploration since they do not exploit information about global attacks and the notice of corruption. 
As such, we pose the question whether one can design efficient algorithms that work well in under such global environment shifts and perform better than the trivial round-robin policy.

Against this background, this paper proposes a novel method that takes advantage of this special setting by estimating the global shift from rewards across different arms and uses it to design a suitable statistic for an algorithm design.  

\noindent \textbf{Our contribution and organisation:}
As far as we are aware, this is the first paper to consider MAB in the presence of global environment shifts. 
In Section \ref{sec:problem_form}, we provide a formal definition of the considered problem, then discuss related work.  To address the MAB problem with global environment shifts, we transform it into a regression problem in Section~\ref{sec:selection_policy} and explain why its solution is a good choice for the best-arm predictor. In Section~\ref{sec:linlucb}, we propose the LinLUCB allocation policy which applies the confidence bound based on a regression estimator. Numerical experiments in Section~\ref{sec:test} are conducted to understand the effectiveness of the proposed shift estimator in different allocation policies and to examine how our proposed LinLUCB algorithm performs in various problem settings. Finally, a summary  and ideas for future work will be provided in Section \ref{sec:conclusion}

\noindent \textbf{Notation:} Vectors are denoted by lowercase boldface letters and matrices by uppercase boldface letters. In general, we use a superscript of $t$ or $k$ to refer to its value at time step $t$ or its $k^{th}$ value, respectively. For any integer $K$, $[K]$ denotes $\{1,..,K\}$.  A standard basis of $\mathbb{R}^d$ is given by $\{\bm{e}_i(d) \ \text{for} \ i=1,..,d \}$ where the $i^{th}$ coordinate of vector $\bm{e}_i(d)$ is 1, otherwise 0. For a matrix $\bm{A}$, we denote its transpose by $\bm{A}'$. An identity matrix with a size of $d\times d$ is denoted by $\bm{I}_d$. A probability measure is denoted by $\mathbb{P}$. We use $\mathbb{E}[\cdot]$ to refer to the expectation of uni- or multi-variate random variables. $\mathbb{V}[\cdot]$ and $Cov(\cdot,\cdot)$  denote the variance of a random variable and the covariance between two random variables. For a multivariate random variable, $Cov[\cdot]$ denotes its covariance matrix. Denote $\mathbbm{1}[E]$ as an indicator function of event $E$. We denote a discrete uniform distribution and a continuous uniform distribution with parameters of minimum $a$ and maximum $b$ as $\tilde{\mathcal{U}}(a,b)$ and ${\mathcal{U}}(a,b)$, respectively.

\section{Problem Formulation $\&$ Related Work} \label{sec:problem_form}
In this section, we formally define the new K-armed stochastic bandit problem in the presence of global environment shifts. We then review literature related to our setting and show how global environmental shifts negatively affect existing algorithms for finding the best arm.

\subsection{BAI with Global Environment Shifts}
Given a finite discrete set of arms $[K]$, the reward $r_{ij}$ from arm $i$ under the  $j^{th}$ environment is an \textit{i.i.d} random variable, consisting of three components:
$$r_{ij}=\mu_i+s_{j}+\epsilon$$
where $\mu_i \in \mathbb{R}$ is the true quality of arm $i$,  $s_{j}$ represents a global shift on the reward of all arms that depends on the environment $j$, and noise is normally distributed, $\epsilon \sim \mathcal{N}(0,\sigma^2)$. 

We assume that an agent can only observe two things:
\begin{enumerate}
    \item the reward $r_{ij}$ if arm $i$ is chosen during environment $j$, and 
    \item the time of an environmental change.
\end{enumerate}
Note that no information about the environment is available, in particular we cannot directly observe its shift $s_j$, nor do we have features that describe the environment and that could be used, e.g., in contextual MAB. In addition, we suppose that shifts and noise are independent and do not make any specific assumptions about the structure of environment shifts.

The best arm is defined as the arm with the largest expected reward, $i^*=\arg\max_i \mu_i$, which is independent of the environment. Every environment $j$ is assumed to remain valid from time $t=cp_{j-1}$ to $t=cp_j$, i.e., the environment is piecewise stationary. The duration during which an environment is valid may be stochastic, and we do not need to assume an underlying distribution for the length of environments. 
Note, however, that since there is no prior knowledge about $s_{j}$ for each environment, a single observed reward under an environment cannot provide any statistical information about the arm. Such extremely short environment durations would thus have to be ignored in practice. Instead, we here assume that the length of environment $\Delta cp_j:= cp_j - cp_{j-1} \geq 2$ for all $j$.

\begin{figure}[htbp]
    \centering
    \includegraphics[width=0.48\textwidth]{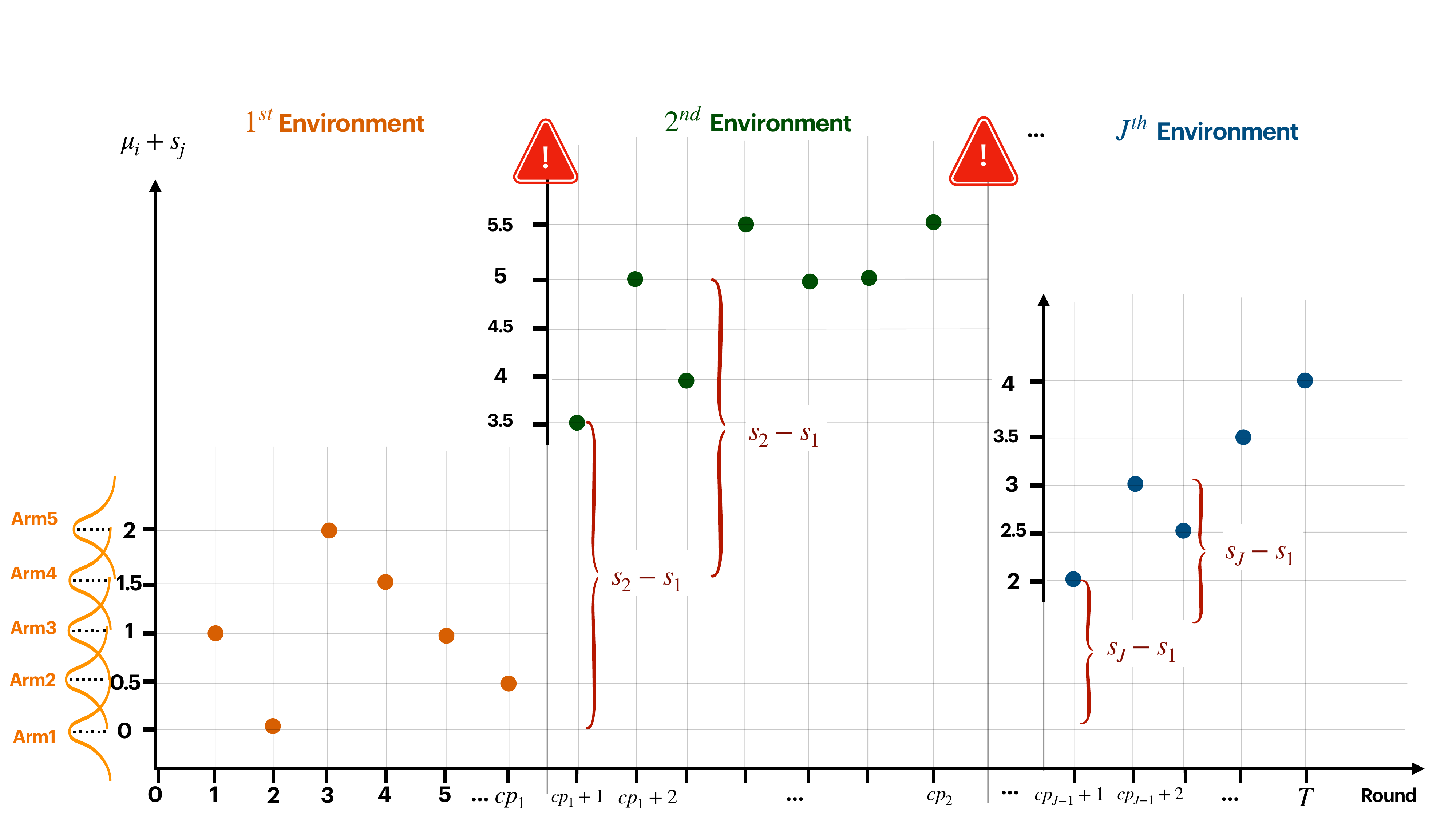}
    \vspace{0.2em}
    \caption{Example of a policy sampling from arms on a BAI problem with global environment shifts.}
    \label{fig:annouced-environment-problem}
\end{figure}

In each time step $t$, we can first observe whether the environment has changed, and then allocate one sample to one of the arms. Our aim is to design an allocation policy that decides which arm to sample next, given all the historical information, and a selection policy that will recommend the best arm at the end of sampling, after having exhausted the available budget of $T$ samples, i.e., we consider a fixed budget setting. A policy $\pi$ is defined as a mapping from sequences of action-reward information, including the environment ordinal, $I^t:=(j^1,i^1,r_{i^1j^1},...,j^{t-1},i^{t-1},r_{i^{t-1}j^{t-1}},j^t)$, to a set of arms $[K]$. Figure \ref{fig:annouced-environment-problem} illustrates an allocation policy in such a setting where the problem has 5 Gaussian arms with different means and the same variance. We use the probability of incorrect selection (PICS) as a performance measure to assess the efficiency of policies. Since the best arm maintains its rank over environmental changes in our setting, the PICS is simply defined by the expectation of 0-1 loss function $\mathcal{L}_{0,1}(\cdot,\cdot)$ as follows 
\begin{align*}
     \mbox{PICS} & :=\mathbb{E}[\mathcal{L}_{0,1}(\hat{i},i^*)]=\mathbb{P}\left(\hat{i} \neq i^* \right) 
\end{align*} where $\hat{i}$ is the arm recommended by the selection policy, and $i^*$ is the true best arm. There also is a so-called fixed-confidence setting aiming to minimise the amount of budget to achieve the specified PICS, but we do not consider such a setting in this paper. 
\vspace{-1em}

\subsection{Related Work}
Our problem formulation shares similarities with (but is different from) many papers on the problem of identifying the best arm in a non-stationary setting. Furthermore, in Section~\ref{sec:effect_on_exist_alg},  we will explain that existing algorithms for stationary settings can identify the best arm in the long run with a high probability under some shift conditions. Therefore, we also review some literature on stationary settings. 

\noindent \textbf{BAI in a stationary  setting}: There is a rich literature on BAI algorithms which assume rewards are \textit{i.i.d}, drawn from a stationary distribution and mostly bounded. In fixed-confidence settings, most algorithms are either elimination-based or confidence-bound-based \cite{jamieson2014best}, such as Exponential Gap Elimination \cite{karnin2013almost}, LUCB \cite{kalyanakrishnan2012pac}, and Lil’UCB \cite{jamieson2014lil}. A simple and efficient algorithm for the fixed-budget setting is Sequential Halving (SH), which divides the total budget into multiple phases and halves the number of candidate arms after a sampling phase ends \cite{karnin2013almost}. There are also some variations of the upper confidence bound (UCB) algorithm applied to this task \cite{bubeck2009pure,audibert2010best}.   

\noindent \textbf{Ranking$\&$Selection (R$\&$S)}: R$\&$S \cite{hong2021review} is a problem class in the stochastic simulation literature, and it is closely related to BAI in MAB, although usually Gaussian reward distributions are assumed \cite{amaran2016simulation}. In a fixed-budget setting, most algorithms are derived either from an equivalent problem of the PICS minimisation with a budget constraint or dynamic programming \cite{shen2021ranking}, such as the OCBA procedure \cite{chen2000simulation}, 0-1 procedure \cite{chick2001new} and Knowledge-Gradient policy \cite{frazier2008knowledge}. These adaptive policies allow batch sampling in multiple phases and work well in practice.

\noindent \textbf{Adversarial Bandits $\&$ Corrupted Bandits}: In general, adversarial settings assume that a sequence of rewards $\{r_i^t\}^T_{t=1}$ for each arm is determined by an adversary \cite{auer2002nonstochastic}, which results in a reward that is not a random variable. There are some variants of adversarial bandits which merge a stochastic structure into the problem formulation. For instance, \textit{corrupted bandits} assume reward distributions can be attacked by an adversary which strives to trick an agent by injecting contaminated information \cite{jun2018adversarial,lykouris2018stochastic}. One can treat our setting as a special case of corrupted bandits where an adversary can instead choose a sequence of global shifts to fool a learner in advance.  Few papers consider BAI tasks in this formulation. For a fixed-budget setting, \cite{zhong2021probabilistic} assume an agent can only observe corrupted rewards $r_i^t=\mu_i+s^t_{i}+\epsilon_i^t$ where an adversary has a bounded total corruption budget, $\sum_{i}^T\max_{i\in[K]}|s_i^t|\leq S$ for some constant $S$ and corrupted rewards are bounded. They propose the PSS algorithm, which is an extension of the SH algorithm \citep{karnin2013almost} with uniform randomisation. Besides, BAI in a corrupted model is studied in a more general way for fixed-confidence settings without any strict assumptions of true reward distribution and contaminated distribution by \cite{altschuler2019best}. In adversarial bandits, the unique best arm over the total time horizon $T$ is possibly undefined without rigorous assumptions. \cite{abbasi2018best} assumes the unique best arm with respect to the highest cumulative rewards exists and studies BAI for the Best-of-Both-world problem. They propose an algorithm P1, in which the probability of sampling each arm $p_i^t$ is generated from a ranking of the inverse-propensity-score (IPS) estimator, and the final recommendation is an arm with the highest IPS estimator. Note that the IPS estimator $\bar{\tilde{r}}^T_i:=(1/T) \sum_{t=1}^T {r}_i^t/p_i^t \cdot \mathbbm{1}[i^t=i]$ is an unbiased estimator of the average reward up to time $T$.   Another way to define the best arm is by assuming the convergence of the reward sequence or $\lim_{t\rightarrow\infty}r_i^t$ exists \cite{jamieson2016non,shen2019universal} and the universal best arm is defined by the highest limit.  To the best of our knowledge, no BAI study in adversarial bandits considers the global structure of change, and in Section \ref{sec:effect_on_exist_alg}, we will empirically show that without exploiting such a structure, these algorithms do not work well in our setting. 

\noindent \textbf{Piecewise-stationary Bandits}: This type of bandit problem is quite relevant to our setting since it allows mean $\mu_i^t$ of the reward distribution to remain stationary within a certain time horizon $\Delta cp_j$ for $j \in [J]$ where $J$ is the number of environmental changes up to time $T$. Similar to adversarial settings, the task of minimising regret is more natural to study. When environments do not change too frequently, and the change is abrupt, there are three general approaches to tackle this setting \cite{garivier2011upper,allesiardo2015exp3,cavenaghi2021non,besson2022efficient, hartland2006multi}:
\begin{enumerate}
    \item Reset strategy if drift is detected
    \item Discounted factors to reduce the importance of rewards received long ago
    \item Sliding window to only evaluate rewards from a desirable time window.
\end{enumerate}
Some works also introduced an evolutionary algorithm and an adaptive allocation strategy to track the best arm under abrupt changes \citep{koulouriotis2008reinforcement}. We are aware of only one study of BAI for piecewise stationary bandits \citep{allesiardo2017non}. Their setting is a generalisation of adversarial settings where an adversary chooses a sequence of reward distributions instead of a sequence of rewards. Some distributions possibly have zero variances. The best arm is defined by $i^*_{PWS}=\arg\max_i \sum^{T}_{t=1} \mu_i^t$. They propose the SER3 algorithm that combines a successive elimination mechanism with randomised round-robin sampling, utilising a criterion derived from Hoeffding’s inequality to eliminate potentially inferior arms until only one best-predicted arm remains. In our paper, drift detection is not required since we assume the agent knows when the change occurs. Besides, our study is a fixed-budget setting, different to the fixed-confidence setting of \cite{allesiardo2017non}. But most importantly, we assume global shifts that affect all arms in the same way, whereas this is not the case in the other publications. 

\noindent \textbf{Linear Bandits}: In Section \ref{sec:selection_policy}, our reward model will be vectorised as a linear function of the index of the arm and of the environment, which is closely related to the linear relationship of feature and reward of linear bandits. For BAI in linear bandits setting, each arm $i$ is represented by a known feature vector $\bm{x}_i \in \mathcal{X} \subset \mathbb{R}^d, |\mathcal{X}|=K$. At time $t$, a noisy reward $r^t$  is assumed to be a linear function of an unknown model parameter $\bm{\theta}^t \in \mathbb{R}^d $; $r^t =\bm{x}^t {\bm{\theta}^t}' + \epsilon^t$. In fixed-budget settings, most of the works assume the unknown parameter is fixed, $\bm{\theta}^t=\bm{\theta}^*$ for all $t$; therefore, the best arm is defined by the highest expected reward mean, $i^*_{LB}=\arg\max_{i} \bm{x}_i {\bm{\theta}^*}'$. \citep{ijcai2022p388} develops the GSE algorithm for which the total budget is evenly split into multiple phases, and a specified number of arms is eliminated after each phase ends. The GSE algorithm applies an adaptive sampling in each phase and uses the least square estimator of $\bm{\theta}^*$ to rank the arms for elimination of the worst. 
\citep{yang2022minimax} proposes the OD-LinBAI algorithm which combines the ideas of the SH algorithm and G-optimal design \cite{lattimore2020bandit}. \citep{alieva2021robust} propose a variant of the SH algorithm equipped with the least square estimator which is robust to moderate levels of misspecification from the linear bandits model. A recent paper \cite{xiong2024b} generalises the assumption of a static model parameter to a non-stationary setting. The goal is to find the optimal arm $i^*$ over the average model parameter $\bar{\bm{\theta}}^T=\sum_{t=1}^T \bm{\theta}^t/T$ at the specified time horizon $T$; $i^*=\arg\max_i \bm{x}_i\bar{\bm{\theta}}^T$. The authors propose the G-BAI algorithm, which samples the next allocation based on G-optimal design and estimates $\theta^t$ from an inverse-propensity score estimator. From the BAI in linear bandits literature, a major difference to linear bandits from our study is that the dimensionality $d$ is fixed, whereas in our setting, the number of dimensions (environments encountered) keeps growing. In order to apply linear bandit algorithms in our setting, since there is no feature about the environment apart from a growing index of environment, a tabular approach and an approach of averaging the model parameter will not be very effective.

\subsection{Effect of Environment Change on Existing Policies} \label{sec:effect_on_exist_alg}
In our setting, the global shift can affect policies in two major ways: 
\begin{enumerate}
    \item the behaviour of the adaptive allocation policy, and
    \item the selection of the best-predicted arm.
\end{enumerate}
We consider a sample mean of reward, which is one of the most commonly used statistics in BAI algorithms such as SH, UCB, and LUCB, including the criteria of the selection policy of round-robin sampling. Denote $\bar{r}_i:=\sum_{j=1}^{J}(\sum_{k=1}^{n_{ij}} r_{ij}^k)/\sum_{j=1}^{J} n_{ij}$ as the sample mean of arm $i$ where $J$ is the latest environment during sampling, $r^k_{ij}$ is the $k^{th}$ reward or arm $i$ in environment $j$, and $n_{ij}$ is the number of samples on arm $i$ under the  $j^{th}$ environment. Under our setting the difference of sample means between arm $i_1$ and $i_2$, $\bar{r}_{i_1}-\bar{r}_{i_2}$ contains the term of $\sum_{j=1}^{J} s_j\left( {n_{i_1j}}/{\sum_{j=1}^{J} {n_{i_1j}}}-{n_{i_2j}}/{\sum_{j=1}^{J} {n_{i_2j}}}\right)$. From such a calculation, the influence of the environment can lead to biased sample means and biased differences if the numbers of samples of each arm under each environment are different. For example, in the case of only one environment change happening or $J=2$, suppose an inferior arm $i_1$ such that $\mu_{i_1}-\mu_{i_2}<0$ has more samples than a superior arm $i_2$ in the second environment $n_{i_12} \geq n_{i_22}$ meanwhile for the first environment they have an equal number of samples $n_{i_11}=n_{i_21}$. If $s_2$ is sufficiently larger than $s_1$ then decision-makers may select an inferior arm due to $\bar{r}_{i_1}-\bar{r}_{i_2} > 0$.  \\
\indent Such a calculation is a main issue for the sample-mean-based final selection if an adaptive allocation policy is used. This phenomenon can also occur in elimination-based algorithms, even when uniform sampling is used, since the change cannot be controlled. We may deduce that the sample mean is not a suitable statistic for both allocation policy and selection policy if no knowledge about the shift is provided. However, if the shift satisfies the conditions in Corollary 2 of \cite{etemadi1983stability}, such as shift is a uniform random variable, existing BAI algorithms that sample all arms sufficiently under different environments will be able to identify the best arm with a high probability. The main reason is the shift term in the sample-mean calculation $\sum_{j=1}^{J} s_j\left( {n_{i_1j}}/{\sum_{j=1}^{J} {n_{i_1j}}}-{n_{i_2j}}/{\sum_{j=1}^{J} {n_{i_2j}}}\right) \rightarrow 0$ as $J\rightarrow \infty$ for all $i \in [K],j \in [J]$. 

\indent Another approach is to use the IPS estimator, which is an unbiased estimator for randomisation-based algorithms in adversarial settings. However, with the same reason as sample mean calculation, insufficient sampling for some arms in some environments can still cause a bias for ranking the IPS estimator since a probability-weighted reward in a favourable environment can be excessive when it is compared to the one in a less favourable environment. Lastly, implementing robust BAI algorithms in contaminated bandits could alleviate the estimator problem, but without exploiting the global shift structure, that algorithm still needs high budgets to identify the best arm.  

\begin{figure}[htbp]
    \centering
    \includegraphics[width=0.5\textwidth,trim={3.5cm 0cm 1.5cm 1.5cm}]{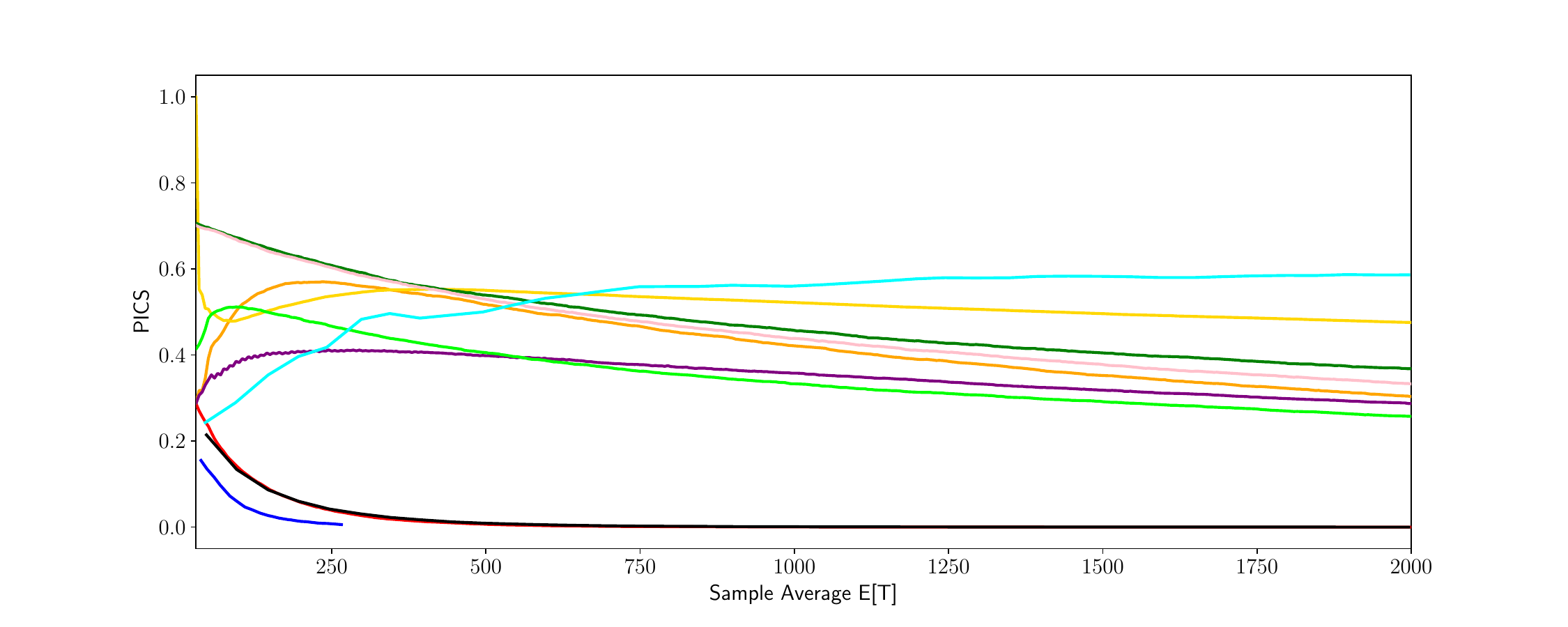}
    \includegraphics[width=0.7\textwidth,trim={3.5cm 1.5cm 2.5cm 1.5cm}]{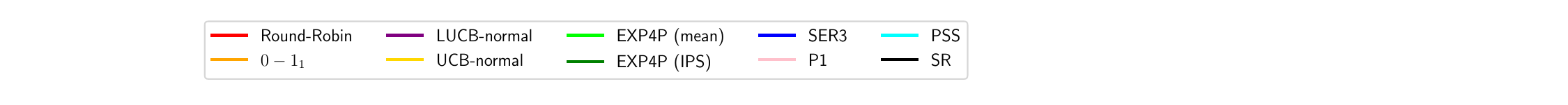}
    \vspace{0.5em}
    \caption{PICS of existing algorithms from $10^5$ replications on the Gaussian configuration of 5 arms where the gaps of ordered arms ($\delta=0.5$) are equally distributed and arms have equal variance ($\sigma=1$). The lengths of environments $j$ are  uniformly distributed, $\Delta cp_{j} \sim \tilde{\mathcal{U}}(2,50)$ and the shift is a random variable, $s_j \sim \mathcal{U}(0,20)$.}
    \label{fig:homo_shift-original_alg}
    \vspace{2em}
\end{figure}

Figure \ref{fig:homo_shift-original_alg} depicts how different existing policies perform under the presence of global shifts when the shift is relatively big in comparison to the gap between optimal arm and suboptimal arm. On the horizontal axis, the sample average refers to the given budget $T$ for each policy except the SER3 algorithm, where it means the average of the required number of samples to achieve different PICSs. The round-robin sampling is executed as a simple baseline. For UCB-based algorithms, LUCB \cite{jamieson2014best} with the sample-mean-based recommendation and UCB \cite{auer2002finite} for minimising cumulative regret with the most-frequency-based recommendation are implemented by using the normal confidence bound in \cite{auer2002finite}. For an algorithm in adversarial settings, the P1 algorithm and the EXP4P algorithm \cite{xu2020regret} with different final recommendations are executed; one is the sample mean, and another is the IPS estimator. For BAI in contaminated bandits, we apply the PSS(2) algorithm with a slight modification by using a randomised round-robin instead of uniform randomisation to ensure each arm is sampled equally. In addition, we mimic such an idea by testing the Successive Rejects (SR) algorithm \cite{audibert2010best} with a randomised round-robin sampling. We also implement the $\text{0-1}_1$ procedure \cite{chick2010sequential} from R$\&$S literature which works well in practice with the Gaussian distribution assumption. The PICS plot of these adaptive policies decreases significantly slower compared to the round-robin sampling when the budget is higher. SR algorithm performs slightly worse than round-robin sampling as sample-mean-based elimination criteria have more risk in this setting. Interestingly, the PICS of the PSS algorithms show a significant difference even if they use the same sampling policy. Two major reasons are that first, eliminating half of the candidate arms in the first phase by using a sample mean has a higher risk of excluding the optimal arm than one-arm elimination, and second the sample mean in the PSS algorithm is computed from rewards in one particular phase which is not sufficient to reduce the influence of shift in the sample mean calculation. The best policy is the SER3 algorithm, which is quite robust to global change, even though the elimination criteria are built on the bounded reward assumption. However, this policy is not quite suitable for use in fixed-budget settings since we need to tune the hyperparameter of the probability of selecting the best to match the limited budget. This result, hence, raises the question of whether there is a better estimator and adaptive policy compared to uniform-exploration-based sampling.

\section{Linear Regression for The Selection Policy} \label{sec:selection_policy}
As explained in Section \ref{sec:effect_on_exist_alg}, even if $s_j$ is bounded, a sample mean of rewards may not be an appropriate statistic for predicting the best arm since different arms may have been evaluated under different environment shifts. In the following, we derive a point estimate by formulating a regression problem.  
\subsection{Ordinary Least Square (OLS) Estimator} \label{sec:regression}
Since we are only interested in identifying the best arm, without loss of generality, we assume that $s_1=0$. Considering the stated problem as a regression model, a reward matrix, given a total number of evaluations $N$ across $J$ environments, can be rewritten in two ways as follows 
\begin{equation} \label{eq:hybrid_linear_model}
    \bm{r} =\bm{A} \bm{\mu} +\bm{B} \bm{s} + \bm{\epsilon}
\end{equation}  
\begin{equation} \label{eq:regression}
   \bm{r} =\bm{X} \bm{\theta} + \bm{\epsilon}
\end{equation}
where
\begin{itemize}
    \item $\bm{r}:=[r^1 \ r^2 \ \ldots \ r^N ]'$ is a column vector containing all rewards obtained from $N$ evaluations
    \item $\bm{\epsilon}:=[\epsilon^1 \ \epsilon^2 \ \ldots \ \epsilon^N ]'$ is a corresponding noise vector where $\bm{\epsilon} \sim \mathcal{MN}(\bm{0},\sigma^2 \bm{I}_N)$. 
    \item $\bm{A}$ is a coefficient matrix in which each row $\bm{a}_t$ is a vector of the standard basis of $\mathbb{R}^{K}$ referring to the chosen arm $i^t$. i.e., $\bm{a}^t=\bm{e}'_{K}(i^t)$
    \item Similarly, $\bm{B}$ is a coefficient matrix referring to the environment ordinal $j^t$, i.e., each row $\bm{b}_t=\bm{e}'_{J-1}(j^t-1)$ for $j\geq2$.
    \item $\bm{\mu} :=[\mu_1 \ \ldots \  \mu_K ]'$ is a $K-$dimensional column vector containing the actual means of all arms.
    \item $\bm{s} :=[ s_2 \ \ldots \ s_J ]'$ is a $(J-1)-$dimensional column vector containing actual shifts relative to the first environment.
    \item $\bm{X}= \left[\bm{A} \ \bm{B} \right]$ is a coefficient block matrix. Similarly, the $(K+J-1)-$dimensional joint parameter vector $\bm{\theta}= \left[\bm{\mu}' \ \bm{s}' \right]'. $ 

\end{itemize}
Note that the dimensions of $J-1$ and $K+J-1$ are due to the zero-valued shift $s_1$ assumption; therefore, such a shift will not be estimated. Model \eqref{eq:hybrid_linear_model} is a hybrid linear model similar to the model in \cite{li2010contextual}. One difference is the dimension of our parameters $\bm{s}$, which grows by 1 when transitioning to a new environment, but the values of parameters in previous environments are unchanged. 

To find the solution to the regression problem, the second model \eqref{eq:regression} is easier to solve. Based on a least squares method, we can derive a unique solution;
\begin{equation*} \label{eq:ols_solution}
    \hat{\bm{\theta}}=(\bm{X}'  \bm{X})^{-1} \bm{X}' \bm{r}. 
\end{equation*}
Note that such an estimator is unbiased ($\mathbb{E}[\hat{\bm{\theta}}]=\bm{\theta}$) which means that the estimated mean and shift are also unbiased; $\mathbb{E}[\hat{\mu}_i]=\mu_i, \mathbb{E}[\hat{s}_j]=s_j$. In addition, the distribution of OLS estimators is $\hat{\bm{\theta}}\sim \mathcal{MN}(\bm{\theta}, \sigma^2 (\bm{X}'  \bm{X})^{-1})$, provided that $\bm{X}$ is fixed, due to Gaussian noise assumption. By using block matrix inversion, we can separate the solution for each parameter as follows
\begin{eqnarray*}
    \hat{\bm{\mu}}&=& \left(\bm{A'}(\bm{I}-\bm{H}_{\bm{B}} )\bm{A}\right)^{-1} \bm{A'} \left(\bm{I}-\bm{H}_{\bm{B}}\right)\bm{r} \label{eq:ols_solution_mean}\\
   \hat{\bm{s}}&=& \big(\bm{B'}(\bm{I}-\bm{H}_{\bm{A}} )\bm{B}\big)^{-1} \bm{B}' \left(\bm{I}-\bm{H}_{\bm{A}}\right)\bm{r} \label{eq:ols_solution_shift}
\end{eqnarray*}
where $\bm{H}_{\bm{A}}:=\bm{A}\big(\bm{A}'\bm{A}\big)^{-1}\bm{A}'$ and $\bm{H}_{\bm{B}}:=\bm{B}\big(\bm{B}'\bm{B}\big)^{-1}\bm{B}'$. In addition, the covariance of both estimators can be computed by 
\begin{eqnarray*}
    Cov[\hat{\bm{\mu}}]&= \sigma^2[\bm{A}'(\bm{I}-2\bm{H}_{\bm{B}}+\bm{H}_{\bm{B}}\bm{H}_{\bm{A}}\bm{H}_{\bm{B}})\bm{A}]^{-1} \label{eq:separated_ols_covar_mean}\\
    Cov[\hat{\bm{s}}]&= \sigma^2[\bm{B}'(\bm{I}-2\bm{H}_{\bm{A}}+\bm{H}_{\bm{A}}\bm{H}_{\bm{B}}\bm{H}_{\bm{A}})\bm{B}]^{-1} \label{eq:separated_ols_covar_shift}. 
\end{eqnarray*}

In the case that a common variance $\sigma^2$ is not known, an unbiased estimator for such variance can be calculated from the following formula 
\begin{equation*} \label{eq:est_var_unbiased}
\hat{\sigma}^2=\dfrac{\sum_{j=1}^J\sum_{i=1}^K\sum_{k=1}^{n_{ij}} \left(r^k_{ij}-\hat{\mu}_i-\hat{s}_j\right)^2 }{N - (K+J-1)}.
\end{equation*}

\subsubsection*{Consistency of mean estimator}

The key challenge of this work is whether the mean estimator can guarantee the correct ranking in the long run since the dimension of the parameter keeps growing. Due to our assumption that $s_1=0$, the correlation between estimated parameters is likely not to vanish, leading to an inconsistent mean estimator. Nevertheless, the estimated ranking is more crucial to identify the best arm; we therefore consider the difference between two mean estimators (ranking) instead.  

\begin{theorem} \label{thm:consistency_ols}
    For any policy under which the OLS estimator is valid and all arms are sampled infinitely often, or $N_i:=\sum_{j=1}^J n_{ij} \rightarrow \infty$ for all $i$, assume that $J \rightarrow \infty$ and there exist constants $v^*$, $w^*$ such that  $0<v^*\leq\mathbb{V}[\hat{s}_j], Cov(\hat{s}_j,\hat{s}_m)\leq w^* < \infty$ for all $j\neq m$.
    \begin{itemize}
        \item[1)] The mean estimator $\hat{\mu}_i$ is not consistent.
        \item[2)] If $J \in o(N)$ and $N_1,N_2 \in \Theta(N)$, the difference in mean estimators between those two arms, $\hat{\mu}_1-\hat{\mu}_2$, is consistent. 
    \end{itemize}
\end{theorem}
The above theorem implies that when all arms are sampled infinitely often, the mean estimator does not converge to its true value, but it can be used to identify the best arm since the ranking still converges to the actual one when the environment change grows sublinearly, and the number of samples for each arm grows linearly. In addition, the consistency of difference holds empirically without such additional assumptions.  Besides, the mean estimator and the difference estimator benefit from robustness against environmental shifts since their consistency does not depend on the shift magnitude. Due to the page limit, the proof and discussion are provided in Supplementary Materials \ref{sec:appendix_proof}. 

\subsection{Requirements for Regression} \label{sec:require_reg}
 Merely merging an OLS estimator with an allocation policy may lead to an ill-posed problem due to a singularity of matrix $X^T X$. In linear bandits, the singularity problem is alleviated by e.g. adding a regularisation term in the regression loss function \citep{li2010contextual,hoffman2014correlation} or applying a dimensionality reduction technique \citep{yang2022minimax,ijcai2022p388}. However, in our setting, these approaches are not helpful if all arms are not observed in the same environment or the loss function can be partitioned and optimised separately since mean estimators are not comparable. Therefore, in general, any allocation policy that applies regression and is not aware of environmental change cannot be directly implemented. In addition, evaluating only one arm in one environment can lead to unchanged mean estimators since an estimated shift in such an environment can be varied arbitrarily. To ensure the existence and uniqueness of the regression solution, there are a few requirements for allocation policies.

\subsubsection*{Initialisation for Regression}
\begin{figure}[h!]
    \centering
    \includegraphics[width=0.475\textwidth]{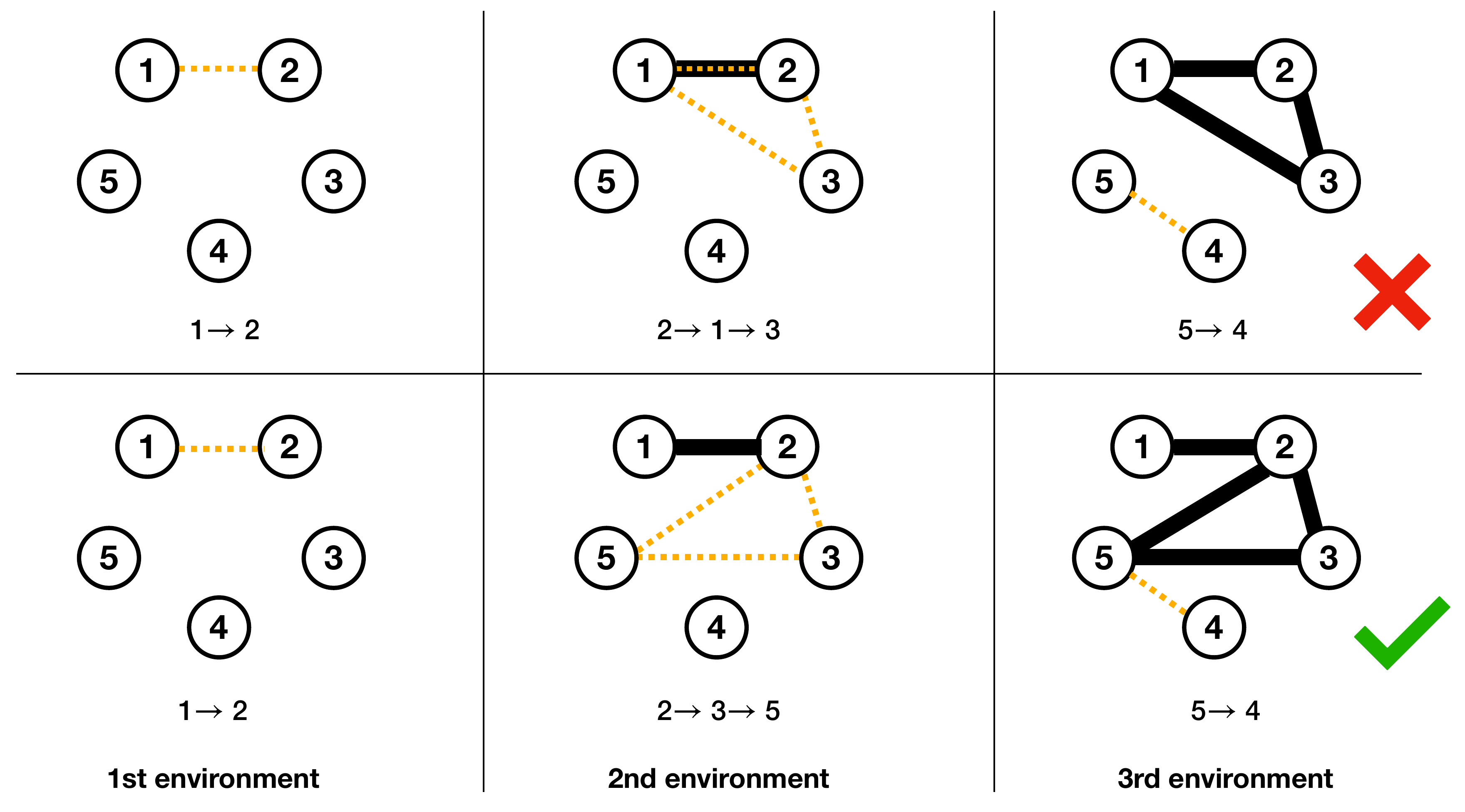}
    \caption{Representative graph structure illustrates how an allocation policy produces the evolution of the graph at the end of each environment.}
    \label{fig:tree regression}
    \vspace{1.5em}
\end{figure}

Disconnected evaluations across different environments can cause an ill-posed optimisation problem. For instance, given a 5-arms setting, if a policy evaluates arms $\{1,2\} $ under the first environment, arms $\{1,2,3\} $ under the second environment and arms $\{4,5\}$ under the third environment, then parameters of the loss function will be separated into two partitions for arms 1, 2, 3 and arms 4, 5. The information share of regression parameters, in fact, can be represented by a graph where the vertices are arms, and the undirected edge between two vertices exists if the corresponding arms have been sampled under the same environments. From the mentioned example,  we can represent it with two sub-graphs where one is $1 - 2 - 3 - 1$, and another is $4 -5$ as in the top row of Figure \ref{fig:tree regression}. So the estimators of arm 1 and arm 4 are not comparable. The regression approach requires a connected graph connecting all arms to fully share information - if the graph is disconnected, it is impossible to rank solutions from different arms relative to each other. The initialisation phase is crucial for every allocation policy to generate at least a tree structure.

 For the initialisation phase, the \textit{randomised round-robin sampling} is modified to evaluate the last arm chosen under the previous environment at the start of the next environment if all arms cannot be observed within one environment. The pseudocode is provided in Algorithm \ref{alg:random_rr}. For example, sequentially evaluate arm $1 \rightarrow 2$ (in the $1^{st}$ environment) $ \rightarrow 2 \rightarrow 3 \rightarrow5$ (in the $2^{nd}$ environment), and $ \rightarrow 5 \rightarrow 4$ (in the $3^{rd}$ environment) as in the bottom row of Figure \ref{fig:tree regression}. With such initialisation, the estimators of arm 1 and arm 4 can be quantitatively compared through arm 2 and then arm 5.

\begin{algorithm}[t] 
    \caption{Randomised round-robin sampling for initialisation} \label{alg:random_rr}
    \begin{algorithmic}[1]
    \Require Number of initial samples per arm $n_0\geq2$
    \State  Set the initial ordinal of environment $j=1$ 
    \State Set an initial arm set $S=[K]$ and shuffle.
    \State Set arm $i^1$ as the first indexed arm in $S$.
    \For{$t=1,...,n_0K$}
        \If{EnvChange==True}
            \State Set the ordinal of the environment: $j\leftarrow j+1$
            \If{BuildTreeSuccess==False}
                \State Play arm $i^t \leftarrow i^{t-1}$ 
            \Else \State Play arm $i^t$ from $S$ in an order following  $i^{t-1}$
            \EndIf
        
        \Else
            \State Play arm $i^t$ from $S$ in an order following $i^{t-1}$
        \EndIf
        \State Obtain a reward $r^t$
        \State Remove arm $i^t$ from $S$ if its number of samples $N_{i^t}=n_0$
        \If{The last indexed arm in $S$ is played}
            \State Shuffle $S$
            \State BuildTreeSuccess $\leftarrow$ True

        \EndIf
    \EndFor  
    \end{algorithmic}
\end{algorithm} 

 \subsubsection*{Evaluating two first distinct arms when the environment changes}
Even if an environment length is relatively short, some allocation policies may evaluate only one arm under the same environment. This would not provide any valuable information since the estimated shift under such an environment can be any arbitrary value subject to the value of the estimated mean. In other words, there are no updates in estimator values if only one arm is evaluated in one environment. In order to avoid such an issue, evaluating at least two distinct arms once the environment changes is imperative. 

\section{LinLUCB Allocation Policy} \label{sec:linlucb}
Given a normal distribution of OLS estimator for the actual means $\hat{\bm{\mu}}$ at time $t$, the upper confidence bound of the actual mean of arm $i$ can be defined as
\begin{align*}
    \text{UCB}_i^t &= \hat{{\mu}}_i + \gamma^t_i \sqrt{{\bm{a}^t}' \sigma^2[\bm{A}'(\bm{I}-2\bm{H}_{\bm{B}}+\bm{H}_{\bm{B}}\bm{H}_{\bm{A}}\bm{H}_{\bm{B}})\bm{A}]^{-1} \bm{a}^t}
\end{align*}
where $\gamma^t$ is an exploration rate at a time step $t$. We propose a new variant of the LUCB algorithm modified from  \cite{kalyanakrishnan2012pac} for our linear model in Algorithm~\ref{alg:LinLUCB-normal}. The LUCB algorithm was originally designed for a PAC subset selection in a fixed-confidence setting where sampling two arms every time step is allowed. However, we found its potential to be implemented in a fixed-budget setting, especially in our setting. The LUCB algorithm ensures that at least two arms are evaluated in every environment, allows for an adaptive budget $T$, and is optimal in a two-armed setting with the worst environment length of 2. The algorithm starts by executing  Algorithm~\ref{alg:random_rr} for the initialisation phase and then alternating samples, the greedy arm and the most potentially best arm from the rest, while guaranteeing that the two first samples in the new environment are distinct. At time $t$, the greedy arm is indexed based on the highest mean estimator $l^t:= {\arg\max}_{i \in [K]} \hat{\mu}_i$, in which ties are broken arbitrarily, then the rest of arms are filtered to find the highest UCB arm $u^t:= {\arg\max}_{i \in {[K]\setminus \{l_t\}}} \text{UCB}_i^t$. In this part, since our setting does not allow the sampling of two arms in one time step, we mimic the batch sampling by sequentially selecting $l^{t}$ and $u^{t}$ instead of using the interleaving strategy. If there is an environment change and the choice of second sampling in such a new environment is the same as the choice of first sampling, we can swap the sampling order of $l^{t}$ and $u^{t}$ to ensure two first choices of sampling are different. Motivated by the UCB1-normal algorithm from \cite{auer2002finite}, we use $\gamma_i^t=\sqrt{16 \ln(t)/\sum_{j=1}^J n_{ij}}$ as an exploration rate. Finally, the selection policy chooses the highest OLS mean estimator as the best-predicted arm. The pseudocode of LinLUCB policy is provided in Supplementary Materials \ref{sec:appendix_linlucb}. 

\section{Empirical Evaluation} \label{sec:test}
In order to understand how environmental change influences different policies on various configurations, we conduct numerical experiments for the proposed algorithm and modified versions of some existing policies. We chose the examined problem settings from \citep{chick2010sequential} since it was a seminal paper developing a policy for PICS minimisation for Gaussian rewards. Two configurations are \textit{monotone decreasing means} (MDM) configuration  and \textit{slippage} configuration (SC), with a modification by adding random shifts $s_j \sim \mathcal{U}(0,20)$. For the MDM configuration, rewards for alternatives $i=1,...,K$ are
$$r_{ij} \sim \mathcal{N}\left(\delta(i-1) + s_j,\sigma^2 \right),$$ while for the SC configuration, rewards are
$$r_{ij} \sim \mathcal{N}\left( s_j, \sigma^2 \right) \ \text{for $ 1 \leq i < K$}, \quad r_{Kj}  \sim \mathcal{N}\left(\delta + s_j, \sigma^2 \right).$$
 We use $\text{PICS}$ as the performance measure estimated by the fraction of replications selecting the true best alternative correctly. For a fair comparison, all procedures in all time steps share the same set of potential observations by controlling random seeds. The PICS convergence plots below are generated using $10^5$ replications. We set the value of parameters in the problem as $\delta=0.5$ and $\sigma=1$. 
 
 \subsection{Comparison against standard policies}
   From the proof of Theorem \ref{thm:consistency_ols}, the uncertainty of the estimated shift plays a vital role in the convergence of the mean OLS estimator. Since the environment length has a significant influence on the shift estimation, we test the performance of our proposed LinUCB policy against other existing policies in the following environmental change scenarios with 5 arms, additional results on different scenarios can be found in the Supplementary materials \ref{sec:appendix_supplement}.
\begin{itemize}
    \item \textbf{General scenario,}  where $\Delta cp_j \sim \tilde{\mathcal{U}}(2,10K)$: The duration of the stationary phase of the environment may vary from very short to relatively long, leading to different challenges in estimating.
    \item \textbf{Cannot-sample-all-arms scenario,} where $\Delta cp_j \sim \tilde{\mathcal{U}}(2, K-1)$: The environment is very short and policies cannot explore all arms in one environment 
\end{itemize} 

\noindent The following list  describes the tested policies:
 
\begin{itemize}
    \item \textbf{Round-robin}: round-robin sampling with $\bar{r}_i$ as a selection policy
    \item \textbf{0-1$_1$}: procedure proposed in \citep{chick2010sequential} and $\bar{r}_i$ as a selection policy
    
    \item \textbf{SER3}: the elimination-based algorithm from \cite{allesiardo2017non} for fixed-confidence piecewise-stationary bandits where prior knowledge about the optimal gap $\mu_1-\max_{i\neq1} \mu_i$ is provided 
    
    \item \textbf{LinLUCB} : Our proposed method (Section $\ref{sec:linlucb}$)
\end{itemize}

\begin{figure}[h]
     \begin{subfigure}{0.24\textwidth}
         \centering
         \includegraphics[width=\textwidth,trim={2cm 0cm 0cm 1.5cm}]{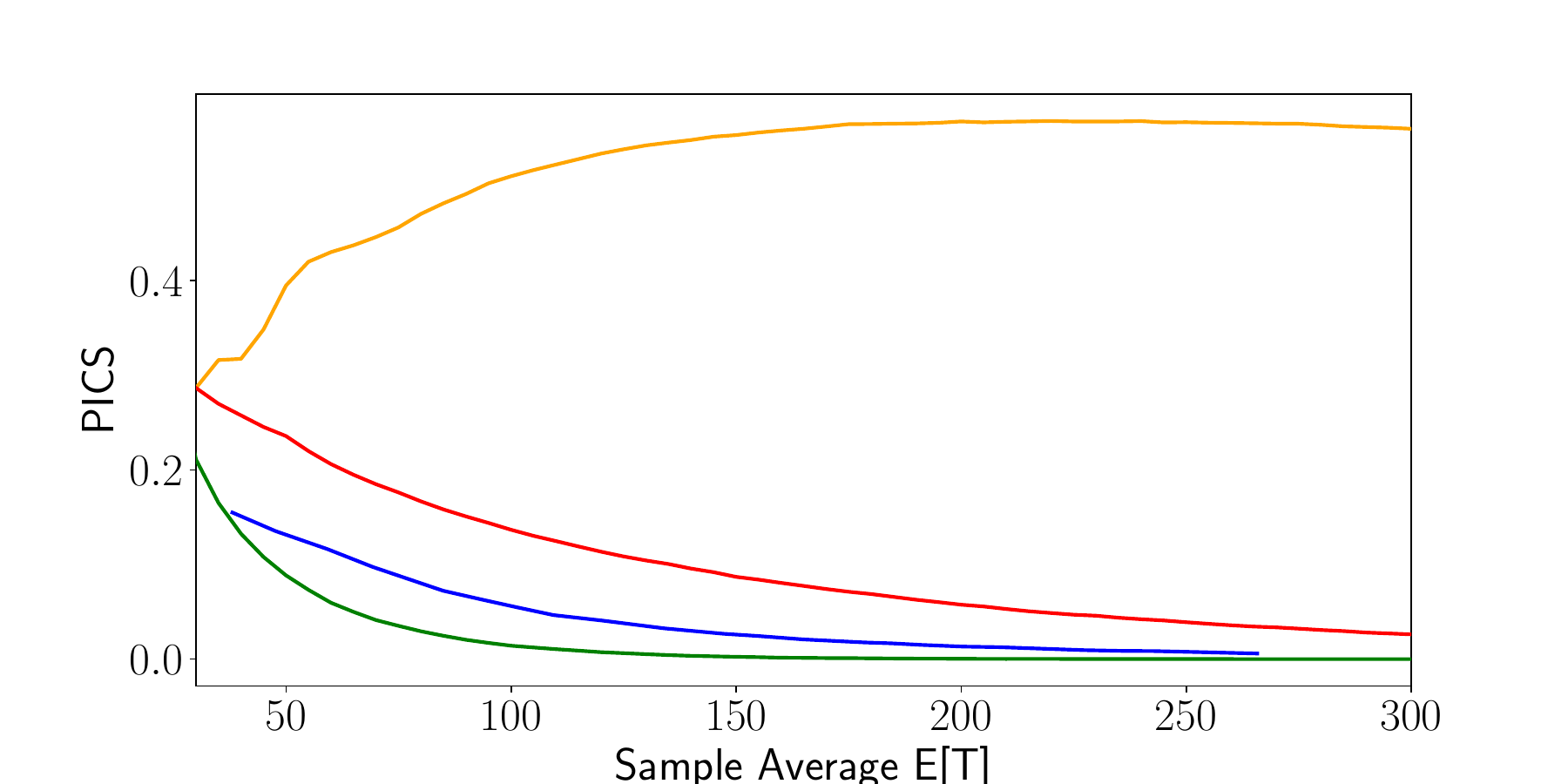}
         \caption{MDM, General}
         \label{fig:mdm5_gen_compare}
     \end{subfigure}
          \hfill
          \begin{subfigure}{0.24\textwidth}
         \centering
         \includegraphics[width=\textwidth,trim={2cm 0cm 0cm 1.5cm}]{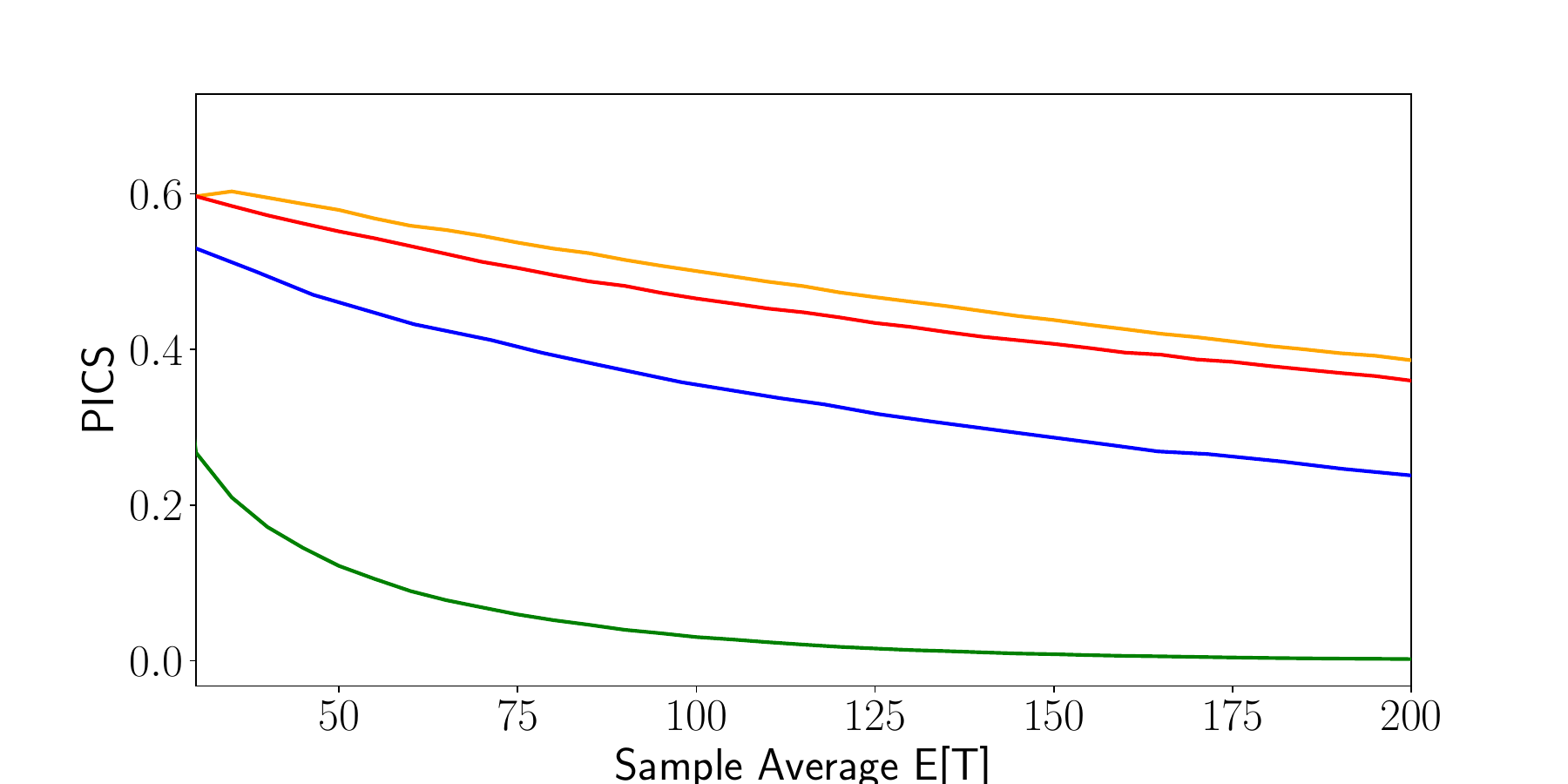}
         \caption{MDM, Cannot-sample-all-arms}
         \label{fig:mdm5_cannot_compare}
     \end{subfigure}

     \vspace{1em}
     \begin{subfigure}{0.24\textwidth}
         \centering
         \includegraphics[width=\textwidth,trim={2cm 0cm 0cm 1.5cm}]{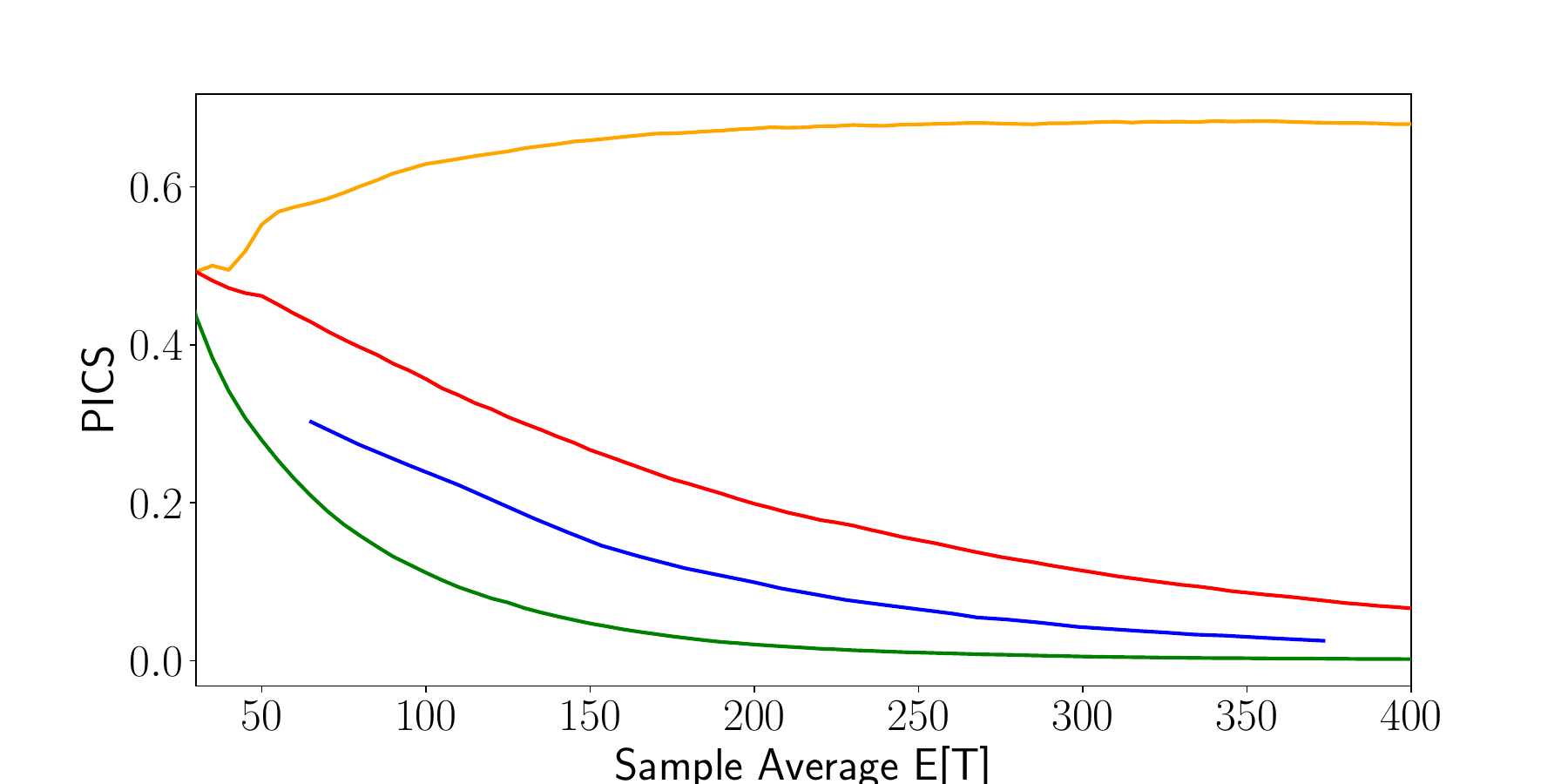}
         \caption{SC, General}
         \label{fig:sc5_gen_compare}
     \end{subfigure}
     \hfill
     \begin{subfigure}{0.24\textwidth}
         \centering
         \includegraphics[width=\textwidth,trim={2cm 0cm 0cm 1cm}]{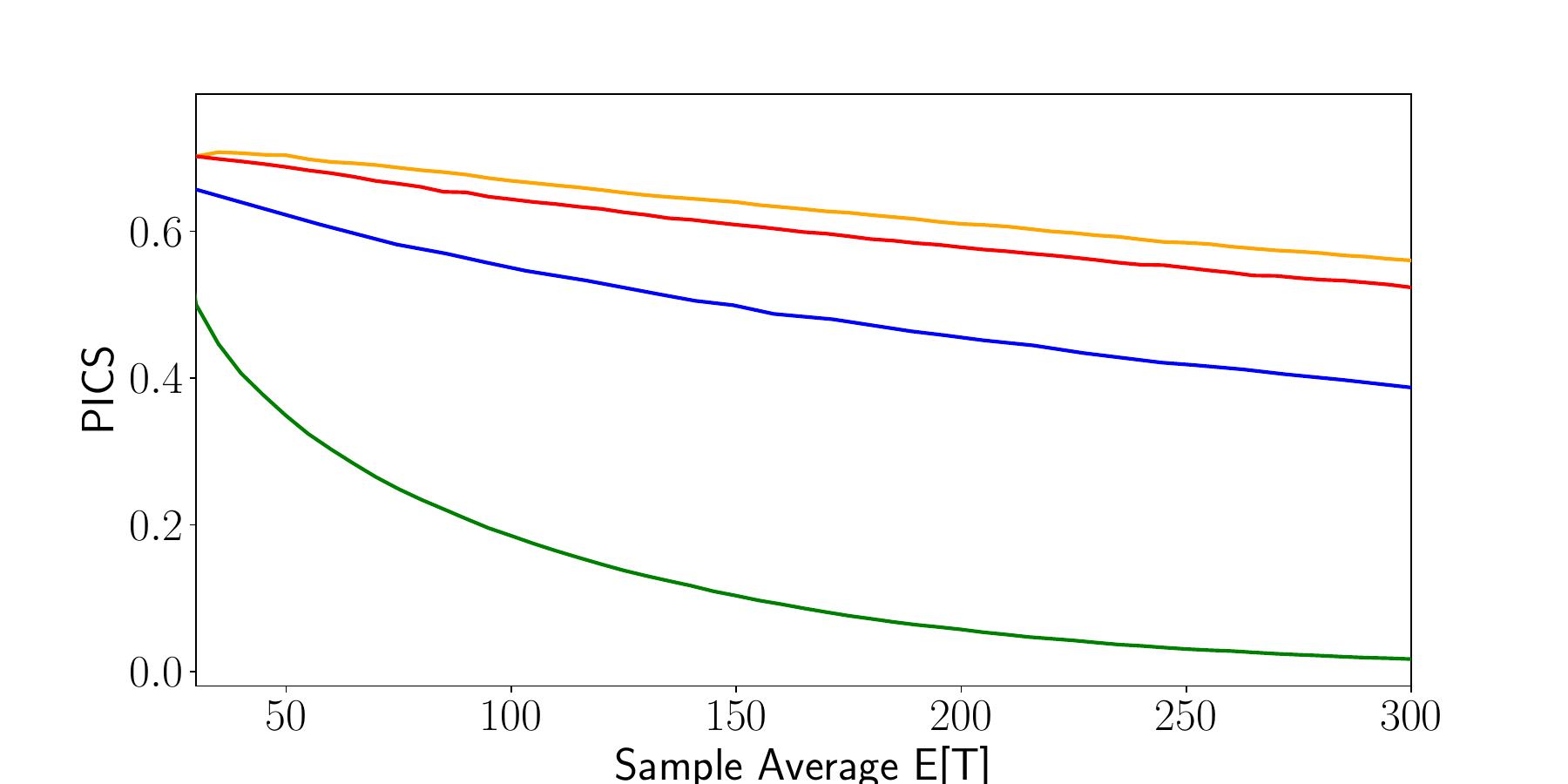}
         \caption{SC, Cannot-sample-all-arms}
         \label{fig:sc5_cannot_compare}
     \end{subfigure}
     \vspace{0.1em}

     \centering
     \includegraphics[width=0.3\textwidth]{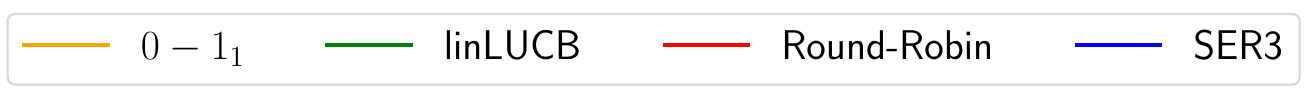}
     \vspace{0.7em}
     
    \caption{The performance of LinLUCB and benchmark policies}
    \label{fig:5arm_compare}
\end{figure}

\noindent Following \cite{chick2010sequential}, 0-1$_1$ and LinLUCB first perform an initialisation phase with $n_0=6$ samples with a round-bin sampling and Algorithm \ref{alg:random_rr}, respectively. As shown in Figure~\ref{fig:5arm_compare}, LinLUCB significantly outperforms other policies in all configurations. With short environment durations (Cannot-sample-all-arms), shift estimation has more uncertainty, and consequently we observe a slower decaying PICS compared to the General setting for both MDM and SC configurations. 
For the 0-1$_1$ policy, the General setting seems actually more difficult because an imbalance of samples per environment can strongly bias the sampling strategy and the selection policy. Sampling from several environments can reduce the dominating effect of a few environmental shifts, resulting in better PICS in quickly changing environments (compare Figure~\ref{fig:mdm5_cannot_compare} with \ref{fig:mdm5_gen_compare} and Figure \ref{fig:sc5_cannot_compare} with \ref{fig:sc5_gen_compare} including the initial worsening in all cases). But even in the Cannot-sample-all-arms scenario, 0-1$_1$ performs worse than Round-Robin.

\begin{figure}[h!]
     \begin{subfigure}{0.24\textwidth}
         \centering
         \includegraphics[width=\textwidth,trim={2cm 0cm 1cm 1cm}]{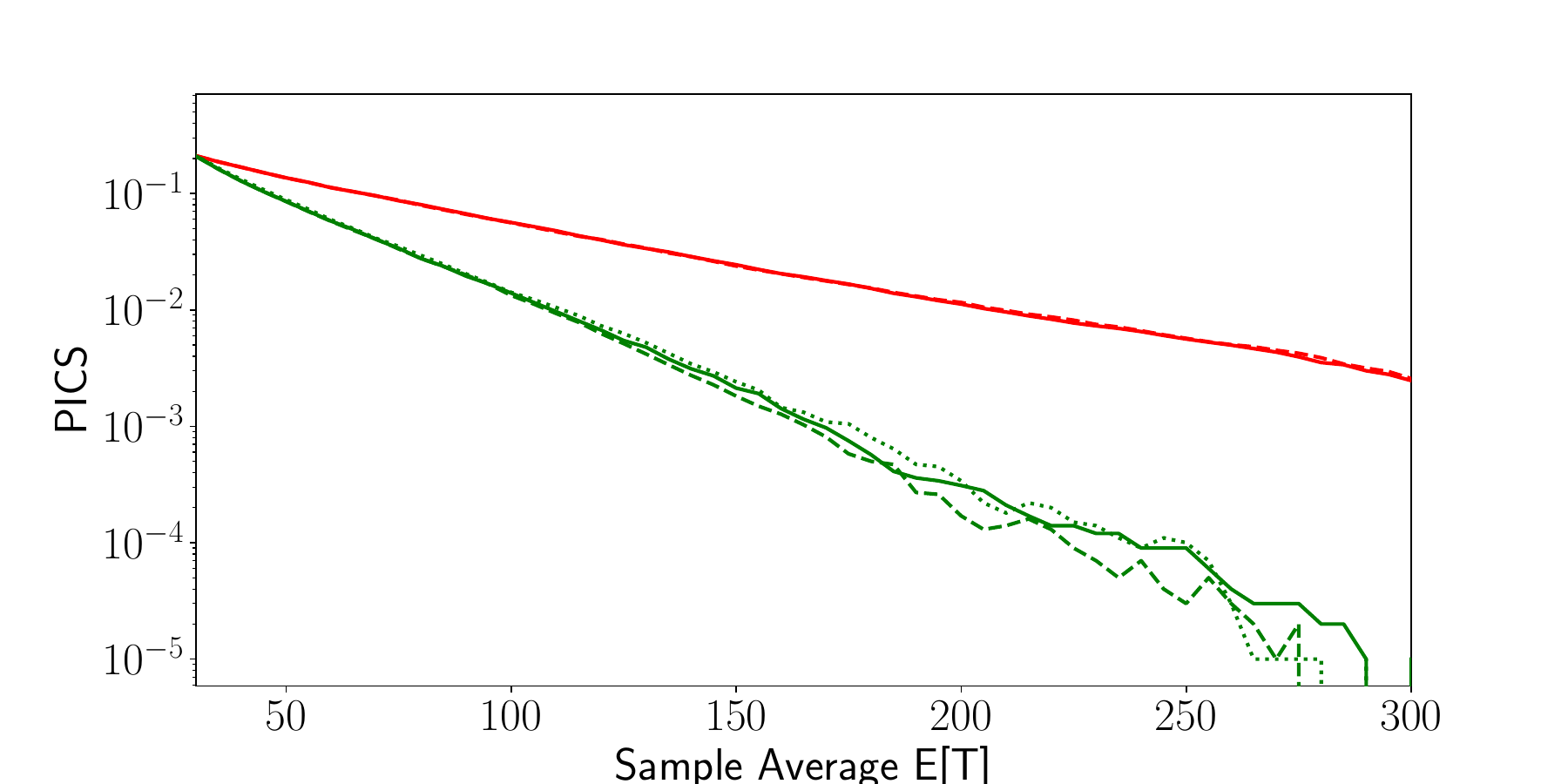}
         \caption{MDM, General}
         \label{fig:reduce_mdm5_gen}
     \end{subfigure}
     \hfill
     \begin{subfigure}{0.24\textwidth}
         \centering
         \includegraphics[width=\textwidth,trim={2cm 0cm 1cm 1.5cm}]{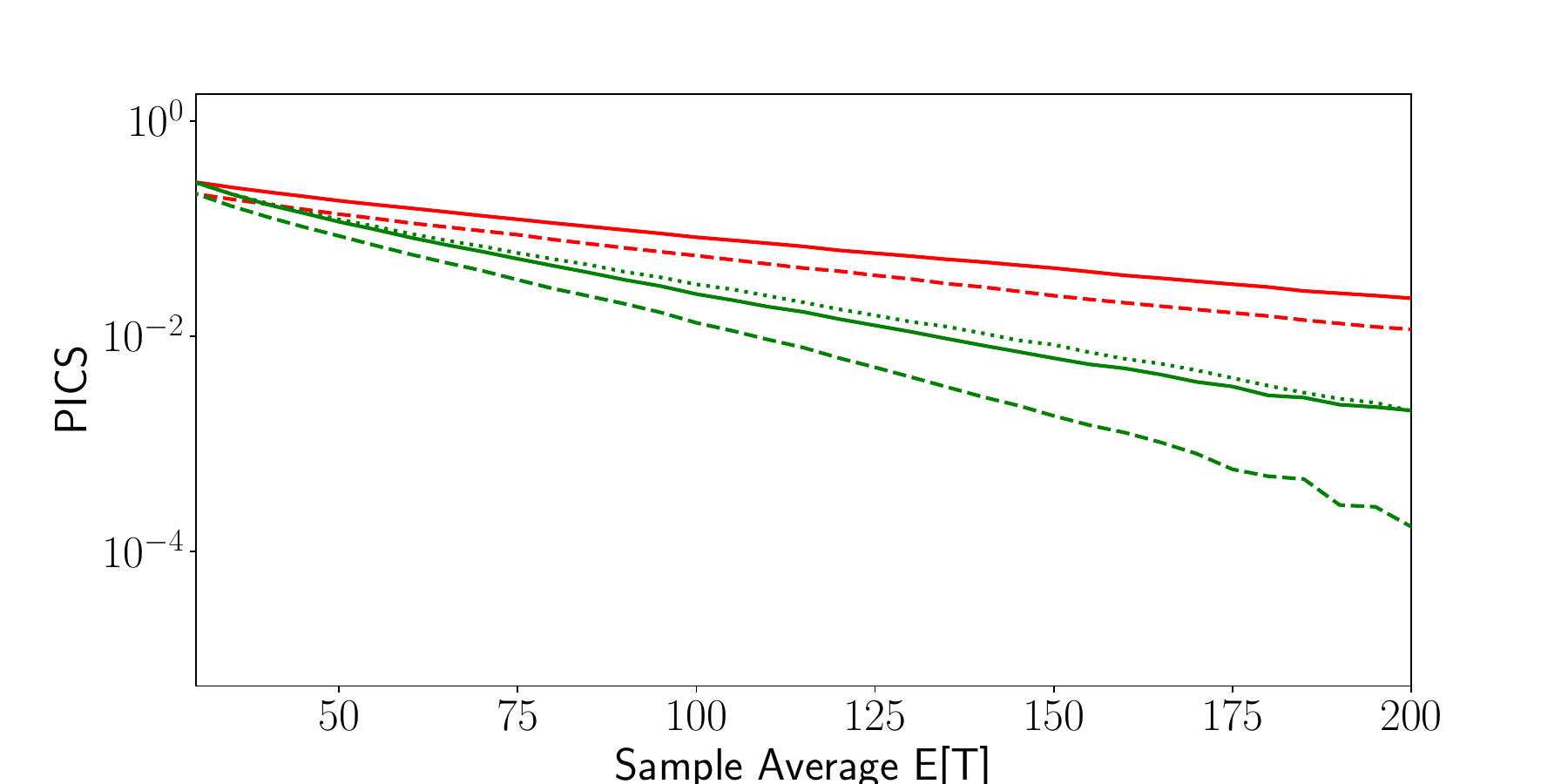}
         \caption{MDM, Cannot-sample-all-arms}
         \label{fig:reduce_mdm5_cannot}
     \end{subfigure}
     
    \vspace{0.5em}
    \begin{subfigure}{0.24\textwidth}
         \centering
         \includegraphics[width=\textwidth,trim={2cm 0cm 1cm 1cm}]{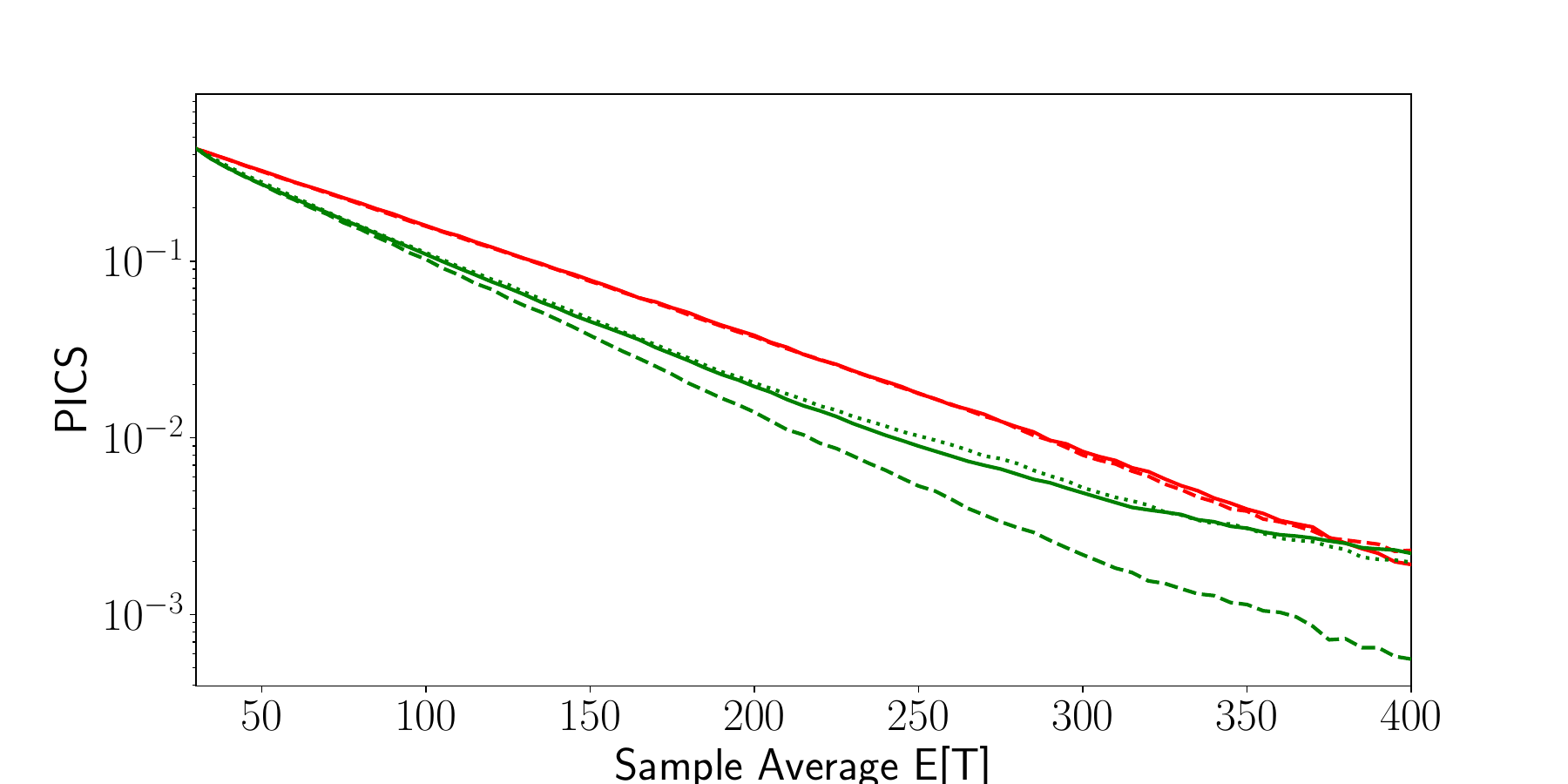}
         \caption{SC, General}
         \label{fig:reduce_sc5_gen}
     \end{subfigure}
     \hfill
     \begin{subfigure}{0.24\textwidth}
         \centering
         \includegraphics[width=\textwidth,trim={2cm 0cm 1cm 1.5cm}]{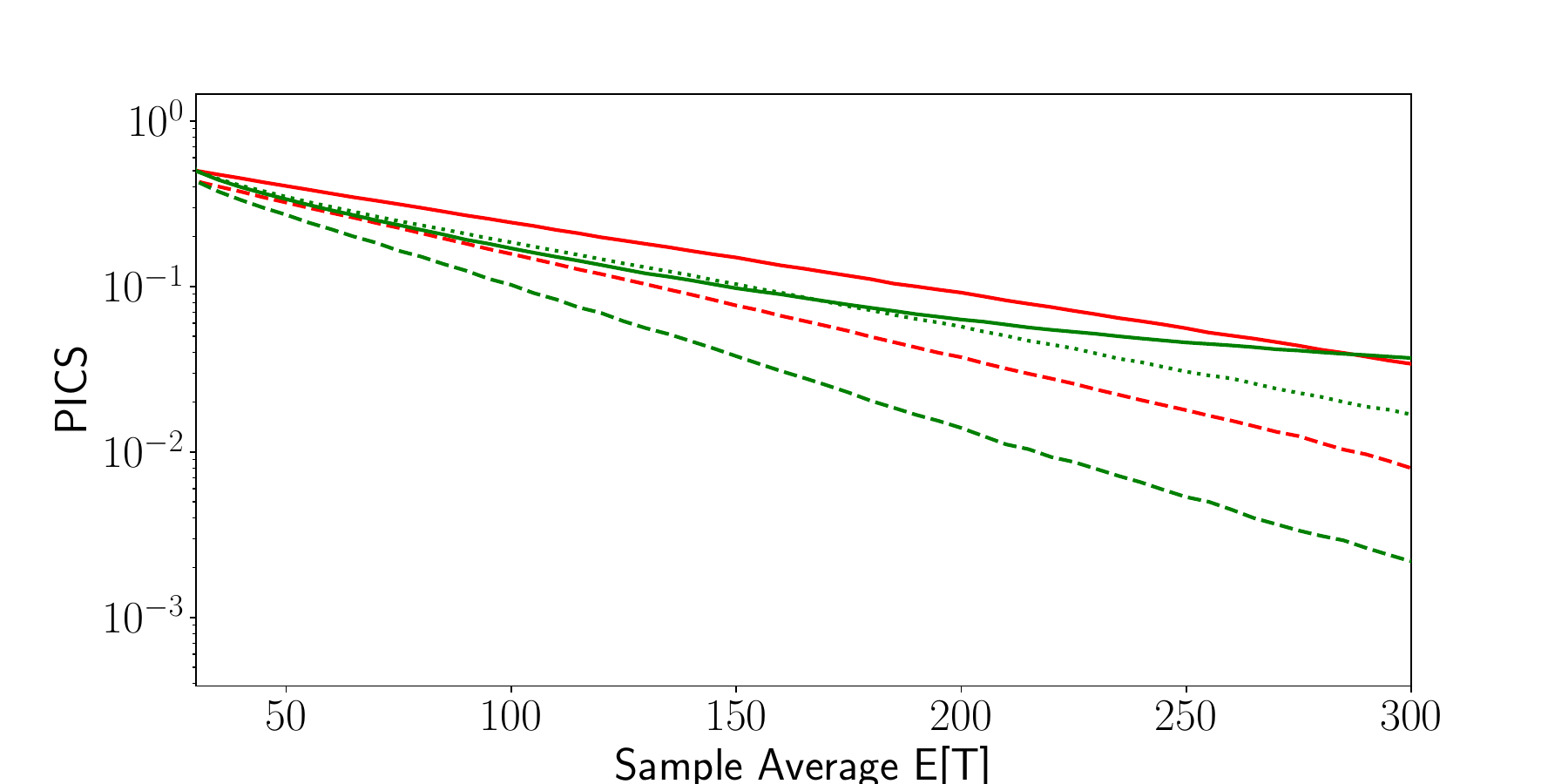}
         \caption{SC, Cannot-sample-all-arms}
         \label{fig:reduce_sc5_cannot}
     \end{subfigure}
     \vspace{0.2em}

     \centering
     \includegraphics[width=0.45\textwidth]{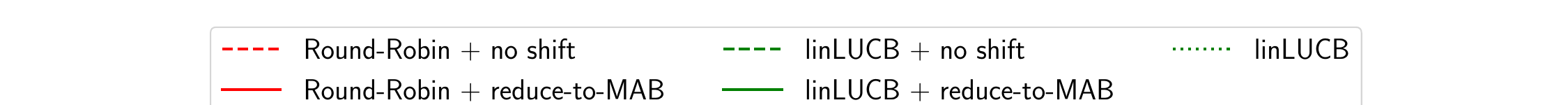}
     \vspace{0.7em}
     
    \caption{Comparison of Reduce-to-MAB strategies and the corresponding performances in a stationary environment}
    \label{fig:reduce-to-MAB}
\end{figure}

 \subsection{Reduce-to-MAB strategy}
In order to gauge the benefit of shift estimation, we test an alternative approach by applying any existing policy designed for a stationary environment, and once a change occurs, we simply subtract the OLS shift estimators from the respective rewards ($r_{ij}^{new}=r_{ij} - \hat{s}_j$) to approximately reduce the problem to a standard MAB problem without shifts (\textit{Reduce-to-MAB strategy}). One can suppose that all modified rewards are Gaussian with the expectation of $\mathbb{E}[r_{ij} - \hat{s}_j]=\mu_i$. All requirements for regression, however, are applied, and all required statistics in any such policies are replaced by statistics calculated from all subtracted rewards instead of original rewards. Note that the sample mean of such modified rewards is equivalent to the OLS mean estimator.

  We implemented \textbf{Round-Robin} and \textbf{LinLUCB} with the \textbf{Reduce-to-MAB strategy} and compared these policies in our test scenarios and, for comparison, under idealised conditions without any environmental shifts. Note that for LinLUCB in a stationary environment, the corresponding UCB$_i$ becomes $\bar{r}_i+\sqrt{16 \ln(t) \tilde{\sigma}^2/(\sum_{j=1}^J n_{ij})^2}$ where $\tilde{\sigma}^2={\sum_{i=1}^K\sum_{k=1}^{n_{i1}} \left(r^k_{i1}-\bar{r}_i \right)^2 }/{(N-K)}$ is an unbiased estimator for $\sigma^2$. Meanwhile, for LinLUCB with the Reduce-to-MAB strategy, the calculation of $\bar{r}_i$ and $\tilde{\sigma}^2$ for UCB$_i$ is instead computed from rewards with the shift estimators subtracted. We also executed the proposed LinLUCB (Section \ref{sec:linlucb}) to investigate the benefit of including the uncertainty of the OLS estimator in the UCB computation. The relatively small gaps in Figure \ref{fig:reduce-to-MAB}  between the policies and their respective performance in a stationary environment demonstrate the effectiveness of shift estimation. Not surprisingly, the gaps are smaller in long-duration environments (General) than in short-duration environments (Cannot-sample-all-arms). Round-robin sampling shows the smallest gap between its variants with and without the OLS estimator. This may be because the adaptive sampling strategy of LinLUCB is susceptible to estimation errors of the shifts, whereas round-robin sampling is not affected. Comparing the variants of LinLUCB, a small advantage of Reduce-to-MAB strategy can only be observed for a very small budget, as may be seen in Figure \ref{fig:reduce_mdm5_cannot} and \ref{fig:reduce_sc5_cannot}. This phenomenon occurs because the value of the exploration term in UCB of the Reduce-to-MAB variant drops faster than of the proposed LinLUCB due to the denominator. Moreover, the variance estimator $\tilde{\sigma}^2$ underestimates its true value in a non-stationary environment and leads to less exploration of the Reduce-to-MAB one. 
   
\section{Conclusion and Discussion} \label{sec:conclusion}
In this paper, we formulate a new setting for fixed-budget best-arm identification in which an environment can globally shift the rewards of all arms in the same way. A selection policy based on ordinary linear regression is proposed to ensure an unbiased and consistent best-arm predictor where the number of environments keeps increasing. We also propose LinLUCB, an algorithm which integrates an error from the mean and shift estimator into the sample allocation decision, constructing a confidence bound that naturally arises from the covariance matrix of the OLS estimator. Empirically, the LinLUCB algorithm is effective in dealing with piecewise stationary environments with global shifts. Besides, our numerical experiments demonstrate the benefits of exploiting the OLS estimator and how the distribution of the duration of stationary periods of the environment affects the performance of policies. Simply using the shift estimates produced by our OLS estimator to reduce the problem to a standard MAB setting works well if the length of environments is sufficiently long. Still, LinLUCB works at least as good and mostly better in all tested cases.

The paper opens several interesting avenues for future work. For instance, the strong assumption of global environmental influence may be relaxed. Also, in real-world contexts, environmental changes are often continuous with smooth transitions rather than piecewise stationary, making it significantly harder to detect the changing point. Lastly, the extension to heterogeneous noise for different arms and different environments will be useful for a more general study.


\begin{ack}
Phurinut would like to acknowledge the Royal Thai Government scholarship sponsored by The Institute for the Promotion of Teaching Science and Technology.
\end{ack}


\bibliography{mybibfile}
\newpage
\appendix
\section*{Supplementary Materials}
\section{Pseudocode of LinLUCB policy} \label{sec:appendix_linlucb}

\begin{algorithm}[h] 
    \caption{LinLUCB with OLS estimator} \label{alg:LinLUCB-normal}
    \begin{algorithmic}[1]
    \Require Total budget $T$, $n_0$ 
    \State  Execute Algorithm \ref{alg:random_rr} until time $t=n_0K$
    \State Set the current environment ordinal $j=J_0$. 
    \State Fit regression model by Eq.\ \eqref{eq:regression}
    \State RecentChange $\leftarrow$ False
    \While{$t<=T$}
    \State $t\leftarrow t+1$
        \If{EnvChange}
            \State Increase the ordinal of the environment: $j\leftarrow j+1$
            \State Play arm $i^t \leftarrow l^{t-1}$  
            \State $t\leftarrow t+1$
            \State Play arm $i^{t} \leftarrow u^{t-2}$
            \State RecentChange $\leftarrow$ False
        \Else
            \If{RecentChange $\wedge (l^{t-1}=i^{t-1})$}
                \State RecentChange $\leftarrow$ False
                \State Play arm $i^{t} \leftarrow u^{t-1}$
                \State $t\leftarrow t+1$
                \If{EnvChange}
                    \State Increase the ordinal of the env.: $j\leftarrow j+1$
                    \State RecentChange $\leftarrow$ True
                \EndIf
                \State Play arm $i^t \leftarrow l^{t-2}$
            \Else
                \State RecentChange $\leftarrow$ False
                \State Play arm $i^t \leftarrow l^{t-1}$ 
                \State $t \leftarrow t+1$
                \If{EnvChange}
                    \State Increase the ordinal of the env.: $j\leftarrow j+1$
                    \State RecentChange $\leftarrow$ True
                \EndIf
                \State Play arm $i^{t} \leftarrow u^{t-2}$
            \EndIf
        \EndIf
    \EndWhile
    \end{algorithmic}
    \Return Best predicted arm $\hat{i}=\arg\max_i \hat{\mu}_{i}$.
\end{algorithm} 

LinLUCB algorithm (Algorithm \ref{alg:LinLUCB-normal}) starts by executing  Algorithm~\ref{alg:random_rr} for the initialisation phase and then alternating samples, the greedy arm and the most potentially best arm from the rest, while guaranteeing that the two first samples in the new environment are distinct. After initialisation phase, at time $t$, the greedy arm is indexed based on the highest mean estimator $l^t:= {\arg\max}_{i \in [K]} \hat{\mu}_i$, in which ties are broken arbitrarily, then the rest of arms are filtered to find the highest UCB arm $u^t:= {\arg\max}_{i \in {[K]\setminus \{l_t\}}} \text{UCB}_i^t$. To mimic a batch sampling of the original LUCB policy \cite{kalyanakrishnan2012pac}, without loss of generality, we generally select $l^t$ and then $u^t$ in the next time step. After exhausting the total budget $T$, the highest mean estimator is returned.  \\
\indent In Algorithm \ref{alg:LinLUCB-normal}, EnvChange refers to whether a current environment changes in the system. To guarantee the first two samples from different arms, we can separate the conditions into three scenarios:
\begin{itemize}
    \item[1.] (Lines 7-12) If the current environment changes before a round of alternating sampling, sample $l^t$ and then $u^t$,
    \item[2.] (Lines 14-22) If there is a change in one step before a round of alternating sampling and the greedy arm $l^t$ is similar to the arm sampled previously, sample $u^t$ and then $l^t$,
    \item[3.] (Lines 24-31) Otherwise, sample $l^t$ and then $u^t$.
\end{itemize}

\section{Analysis of Consistency of the OLS estimator} \label{sec:appendix_proof}

\noindent \textbf{Proof of Theorem \ref{thm:consistency_ols}}
\begin{proof}
         Denote the number of samples of arm $i$ as $N_i:=\sum_{j=1}^J n_{ij}$. An estimator $\hat{\alpha}$ of $\alpha$ is called weakly consistent if for all $\epsilon>0$, $\lim_{N_i \rightarrow \infty} \mathbb{P}[|\hat{\alpha}-\alpha|>\epsilon]=0$. Since shifts are independent of noise, we consider the variance of the OLS estimator conditional on a sequence number of evaluations for an alternative under each environment $\{n_{ij}\}_{j=1}^J$ and a sequence of shifts $\{s_{j}\}_{j=1}^J$, i.e., $\mathbb{V}[\hat{\mu}_i | \{n_{ij}\},\{s_{j}\}]$ and $\mathbb{V}[\hat{\mu}_{i_1}-\hat{\mu}_{i_2} | \{n_{i_{1}j}\},\{n_{i_{2}j}\},\{s_{j}\}]$ (simply write $\mathbb{V}[\hat{\mu}_i]$ and $\mathbb{V}[\hat{\mu}_{i_1}-\hat{\mu}_{i_2}]$, respectively). 
         
         Due to the Gaussian distribution of the OLS estimator, it suffices to show that $\mathbb{V}[\hat{\mu}_1] \nrightarrow 0$ and $\mathbb{V}[\hat{\mu}_1-\hat{\mu}_2 ] \rightarrow 0$ without loss of generality. Instead of investigating the matrix formula of the mean estimator variance, we consider the estimated mean derived from minimising the following loss function, which corresponds to the stated regression problem;
     \begin{align*}
         L(\tilde{\bm{\mu}},\tilde{\bm{s}})&= \sum_{i=1}^N 
         \left[ \sum_{k=1}^{n_{i1}} (r^k_{i1}-\tilde{\mu}_{i})^2+ \sum_{k=1}^{n_{i2}}(r^k_{i2}-\tilde{\mu}_{i}-\tilde{s}_{2})^2 +... \right. \\
         & \left. +\sum_{k=1}^{n_{iJ}}(r^k_{iJ}-\tilde{\mu}_{i}-\tilde{s}_{J})^2 \right]. 
        \end{align*}
          
 Setting the gradient equal to zero, the mean estimator is equivalent to the following formula given the shift estimator $\hat{\bm{s}}$, 
      \begin{align*}
  \hat{{\mu}}_{i} &= \dfrac{n_{i1}\bar{r}_{i1} + n_{i2}(\bar{r}_{i2}-\hat{s}_{2}) + ... +n_{iJ}(\bar{r}_{iJ}-\hat{s}_{J}) }{ N_i}   \\
  &= \bar{r}_{i}  -\sum_{j=2}^J \left( \dfrac{n_{ij}}{N_i}\right)
\hat{s}_j
 \end{align*}
 where $\bar{r}_{i}  \sim \mathcal{N}\left(\mu_i+\dfrac{\sum_{j=1}^{J} n_{ij}s_{j}}{N_i},\dfrac{\sigma^2}{N_i}\right)$. \\ This implies that $Cov(\bar{r}_i,\hat{s}_j) \rightarrow 0$ as $J \rightarrow \infty$ for all $i \in [K]$,$j \in [J]$ due to $\mathbb{V}[\bar{r}_i] \rightarrow 0$ as $N_i \rightarrow \infty$ \\
 
 To prove the convergence of variance, we suppose $0<v^*\leq\mathbb{V}[\hat{s}_j], Cov(\hat{s}_j,\hat{s}_m)\leq w^* < \infty$ for some constant $v^*, w^*$ and all $j\neq m$. Implicitly, we assume that the variance of the shift estimator $\hat{s}_j$ and the covariance of any two shift estimators cannot vanish. Besides, all shift estimators are positively correlated.    \\
 
 1) The variance of the mean estimator can be computed as
  \vspace{-0.5em}
 \begin{align*}
    \mathbb{V}[\hat{\mu}_1 ]&=\sum_{j=2}^J  \mathbb{V}\left[\dfrac{n_{1j}}{N_i}\hat{s}_j\right]-\sum_{j=2}^J 2Cov\left(\bar{r}_1,\dfrac{n_{1j}}{N_i}\hat{s}_j \right)\\
    &+\sum_{j=2}^{J-1} \sum_{m>j}^J 2Cov\left(\dfrac{n_{1j}}{N_i}\hat{s}_j,\dfrac{n_{1m}}{N_i}\hat{s}_m \right)+\mathbb{V}[\bar{r}_1]
\end{align*}
 \begin{align*}
    \mathbb{V}[\hat{\mu}_1 ]&=
    \sum_{j=2}^J \dfrac{n_{1j}^2}{N_i^2} \mathbb{V}[\hat{s}_j]-\sum_{j=2}^J 2\dfrac{n_{1j}}{N_i} Cov\left(\bar{r}_1,\hat{s}_j \right)\\
    &+\sum_{j=2}^{J-1} \sum_{m>j}^J 2\dfrac{n_{1j}}{N_i}\dfrac{n_{1m}}{N_i}Cov\left(\hat{s}_j,\hat{s}_m \right)+\mathbb{V}[\bar{r}_1]\\
    &\geq  v^* \left[ \sum_{j=2}^J \dfrac{n_{1j}^2}{N_i^2} + \sum_{j=2}^{J-1} \sum_{m>j}^J 2\dfrac{n_{1j}}{N_i}\dfrac{n_{1m}}{N_i} \right] +\dfrac{\sigma^2}{N_1}\\
    &=  v^* \left[ \sum_{j=2}^J \dfrac{n_{1j}}{N_i}  \right]^2 + \dfrac{\sigma^2}{N_1}
\end{align*}
As $J \rightarrow \infty$, we have $\left[ \sum_{j=2}^J \dfrac{n_{1j}}{N_1}  \right]^2 \rightarrow 1$. Hence the lower bound of  $\mathbb{V}[\hat{\mu}_1] \rightarrow v^*>0$.\\
 
 2) The variance of the difference between two mean estimators can be computed as; $\mathbb{V}[\hat{\mu}_1-\hat{\mu}_2 ]$
 \vspace{-1em}
\begin{align*}
    &=\mathbb{V}[\bar{r}_1]+\mathbb{V}[\bar{r}_2]+\sum_{j=2}^J  \mathbb{V}\left[\left( \dfrac{n_{1j}}{N_1}-\dfrac{n_{2j}}{N_2}\right)\hat{s}_j\right]\\
    &-\sum_{j=2}^J 2Cov\left(\bar{r}_1,\left( \dfrac{n_{1j}}{N_1}-\dfrac{n_{2j}}{N_2}\right)\hat{s}_j \right)\\
    &+\sum_{j=2}^{J-1} \sum_{m>j}^J 2Cov\left(\left( \dfrac{n_{1j}}{N_1}-\dfrac{n_{2j}}{N_2}\right)\hat{s}_j,\left( \dfrac{n_{1m}}{N_1}-\dfrac{n_{2m}}{N_2}\right)\hat{s}_m \right)\\
    &=\dfrac{\sigma^2}{N_1}+\dfrac{\sigma^2}{N_2}+\sum_{j=2}^J \left( \dfrac{n_{1j}}{N_1}-\dfrac{n_{2j}}{N_2}\right)^2 \mathbb{V}\left[\hat{s}_j\right]\\
    &-\sum_{j=2}^J \left( \dfrac{n_{1j}}{N_1}-\dfrac{n_{2j}}{N_2}\right) 2Cov\left(\bar{r}_1,\hat{s}_j \right)\\
    &+\sum_{j=2}^{J-1} \sum_{m>j}^J \left( \dfrac{n_{1j}}{N_1}-\dfrac{n_{2j}}{N_2}\right) \left( \dfrac{n_{1m}}{N_1}-\dfrac{n_{2m}}{N_2}\right) 2Cov\left(\hat{s}_j,\hat{s}_m \right)\\
    &\leq \dfrac{\sigma^2}{N_1}+\dfrac{\sigma^2}{N_2}+ w^*  \left( \sum_{j=2}^J  \left| \dfrac{n_{1j}}{N_1}-\dfrac{n_{2j}}{N_2}\right| \right)^2 
\end{align*}
To guarantee the absolute sum convergence goes to zero, we suppose the number of environments not to grow too fast $J\in o(N)$, and the number of samples for each arm to be quite similar $N_1,N_2 \in \Theta(N)$. Some algorithms, such as round-robin sampling, can satisfy such an assumption. Therefore  $\sum_{j=2}^J  \left| \dfrac{n_{1j}}{N_1}-\dfrac{n_{2j}}{N_2}\right| \rightarrow 0$. Hence $\mathbb{V}[\hat{\mu}_1-\hat{\mu}_2 ] \rightarrow 0$. 
\end{proof}

Apart from the theoretical results, we found that, empirically, such additional assumptions for the consistency of the ranking can be lifted. From our empirical studies, we conjecture that the value of $Cov(\hat{s}_j,\hat{s}_m)$ converges to one constant, specifically, $Cov(\hat{s}_j,\hat{s}_m) \rightarrow 1/\sum_{i=1}^K n_{i1}$ for all $j\neq m$. Provided that the conjecture holds, we can simply prove that $\mathbb{V}[\hat{\mu}_1-\hat{\mu}_2 ] \rightarrow 0$ without additional assumptions. Besides, the same convergence occurs with the variance of mean estimators, i.e., $\mathbb{V}[\hat{\mu}_i] \rightarrow 1/\sum_{i=1}^K n_{i1}$ for all $i\in [K]$. 

\section{Additional Empirical Results} \label{sec:appendix_supplement}
The code is available on  \href{https://github.com/S-Phurinut/ABtesing-with-warning}{github.com/S-Phurinut/ABtesing-with-warning}. We report the PICS and Expected Opportunity Cost (EOC) (or Simple regret) of the same policies as in Section \ref{sec:test} on MDM and SC configurations with $K=5$ and $K=10$. During the sequential allocation policy, the $j^{th}$ environment will last for $\Delta cp_j \sim \tilde{\mathcal{U}}(cp_{min},cp_{max})$ time steps. Depending on the environment length parameters $cp_{min}$ and $cp_{max}$, we split the experiments into 4 scenarios as follows:
\begin{enumerate}
\item \textit{Worst-case} scenario ($cp_{min}=cp_{max}=2$) 
\item \textit{Cannot-sample-all-arms} scenario ($cp_{min}=2, cp_{max}=N-1$)
\item \textit{Sample-1-to-10-per-arm} scenario ($cp_{min}=N, cp_{max}=10N$)
    \item  \textit{General}  scenario ($cp_{min}=2$, $cp_{max}=10N$).
\end{enumerate}

Overall, LinLUCB outperforms other policies in all configurations. For 0-1$_1$ policy, it performs better in short environments than long environments, i.e., Worst-case and Cannot-sample-all-arms scenarios, since samples of each arm from several environments can reduce the shift influence in choices of allocation and selection. Especially in the Worst-case scenario, the performance of 0-1$_1$ policy is quite competitive to round-robin sampling. For the EOC measures, all results are more or less similar to the PICS measures, and LinLUCB still outperforms the others.  

\subsection{Measure of PICS} 

\begin{figure}[h]
    \begin{subfigure}{0.45\textwidth}
        \centering
         \includegraphics[width=0.7\textwidth]{pic/ECAI-compare_label.pdf}
    \end{subfigure}
     \hfill
     \begin{subfigure}{0.24\textwidth}
         \centering
         \includegraphics[width=\textwidth,trim={2cm 0cm 0cm 1.5cm}]{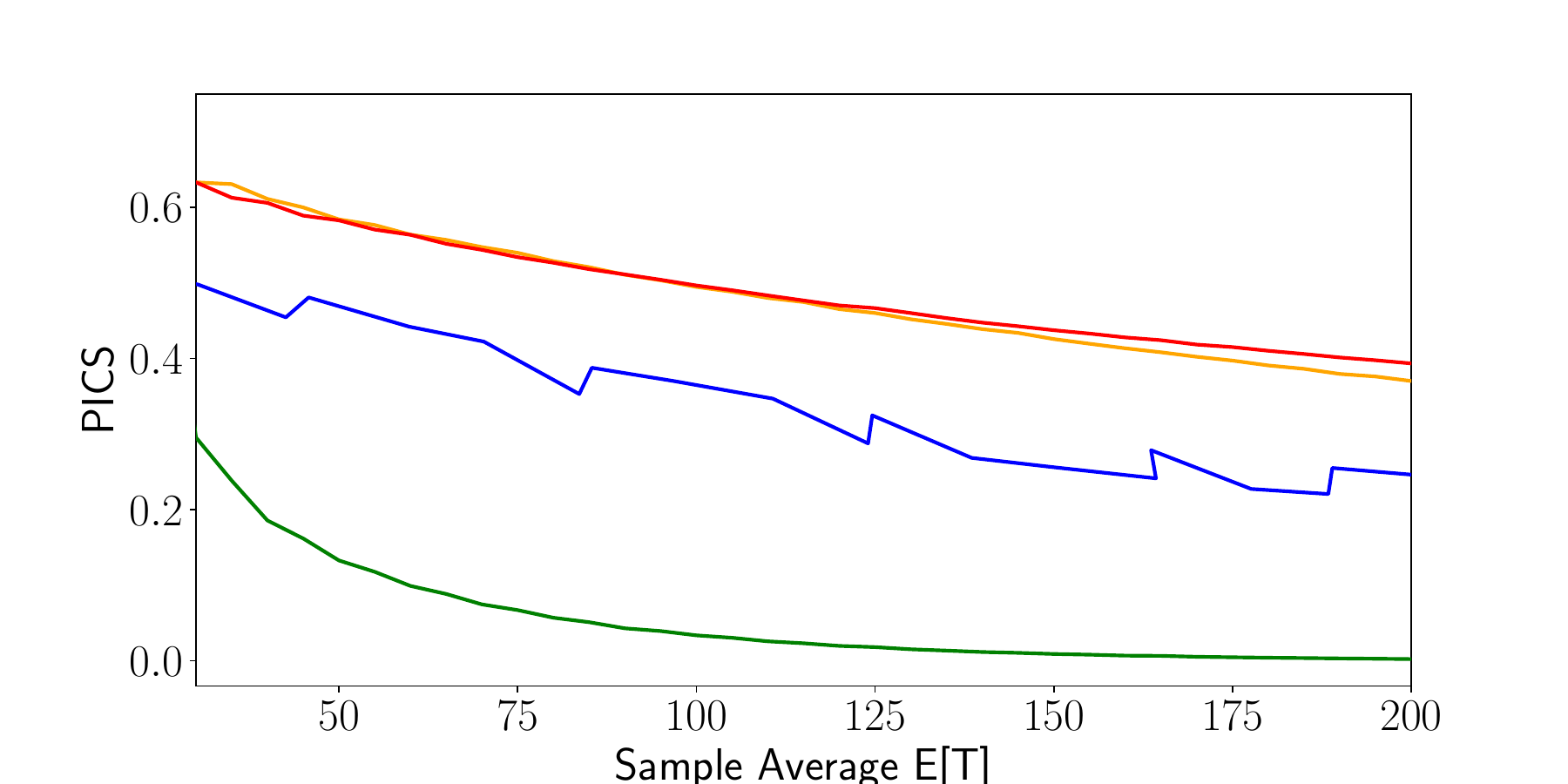}
         \caption{Worst-case}
         \label{fig-appendix:mdm5_worst_compare}
     \end{subfigure}
          \hfill
          \begin{subfigure}{0.24\textwidth}
         \centering
         \includegraphics[width=\textwidth,trim={2cm 0cm 0cm 1.5cm}]{pic/ECAI-compare_cannot_MDM_5arm_var1.pdf}
         \caption{Cannot-sample-all-arms}
         \label{fig-appendix:mdm5_cannot_compare}
     \end{subfigure}

     \vspace{1em}
     \begin{subfigure}{0.24\textwidth}
         \centering
         \includegraphics[width=\textwidth,trim={2cm 0cm 0cm 1.5cm}]{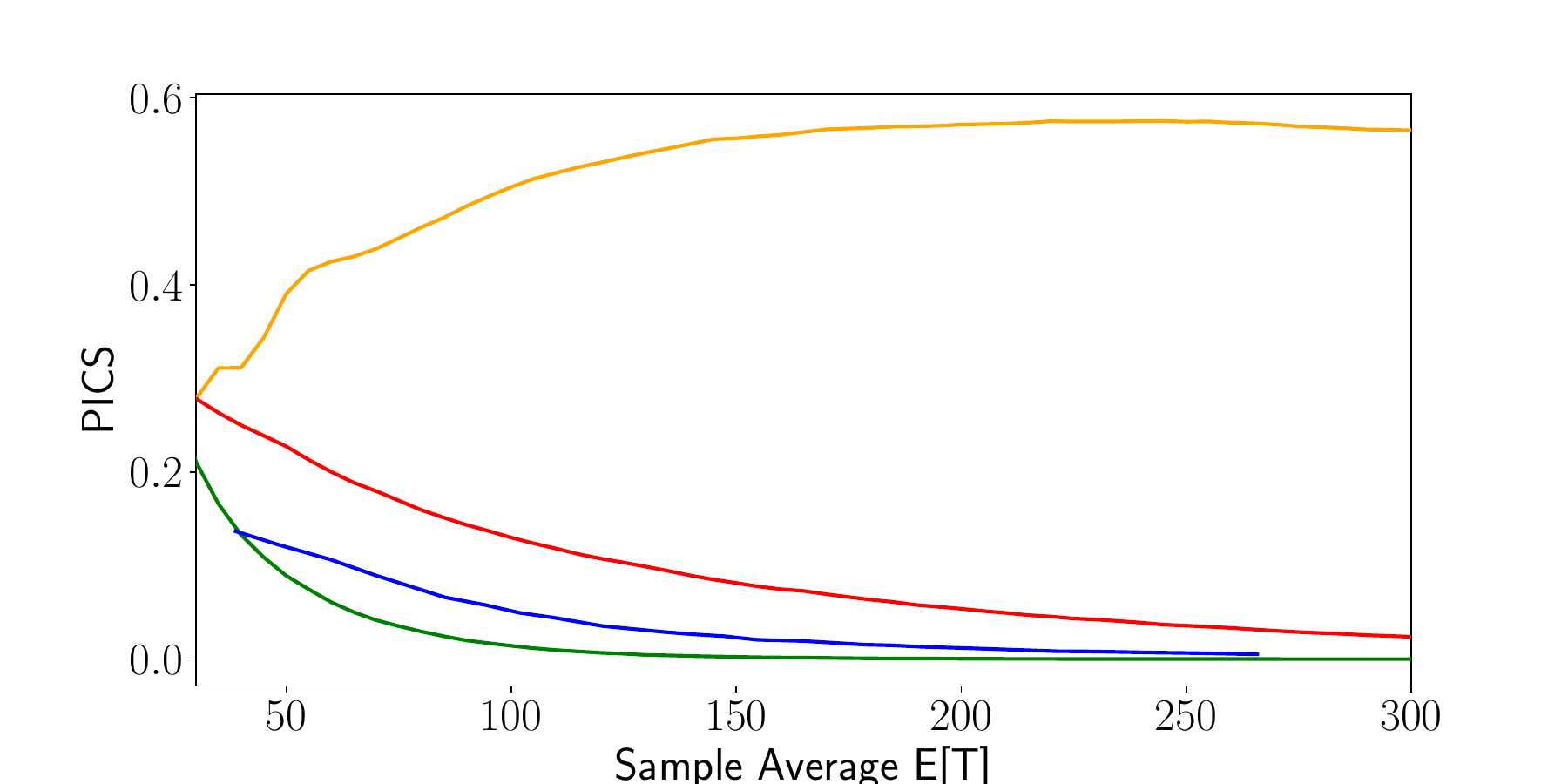}
         \caption{Sample-1to10-per-arm}
         \label{fig-appendix:mdm5_sam_compare}
     \end{subfigure}
     \hfill
     \begin{subfigure}{0.24\textwidth}
         \centering
         \includegraphics[width=\textwidth,trim={2cm 0cm 0cm 1.5cm}]{pic/ECAI-compare_general_MDM_5arm_var1.pdf}
         \caption{General}
         \label{fig-appendix:mdm5_gen_compare}
     \end{subfigure}
     \vspace{0.1em}
    \caption{MDM configurations with 5 arms}
    \label{fig-appendix:MDM_5arm_compare}
    \vspace{1em}
\end{figure}

\begin{figure}[h]
     \begin{subfigure}{0.24\textwidth}
         \centering
         \includegraphics[width=\textwidth,trim={2cm 0cm 0cm 1.5cm}]{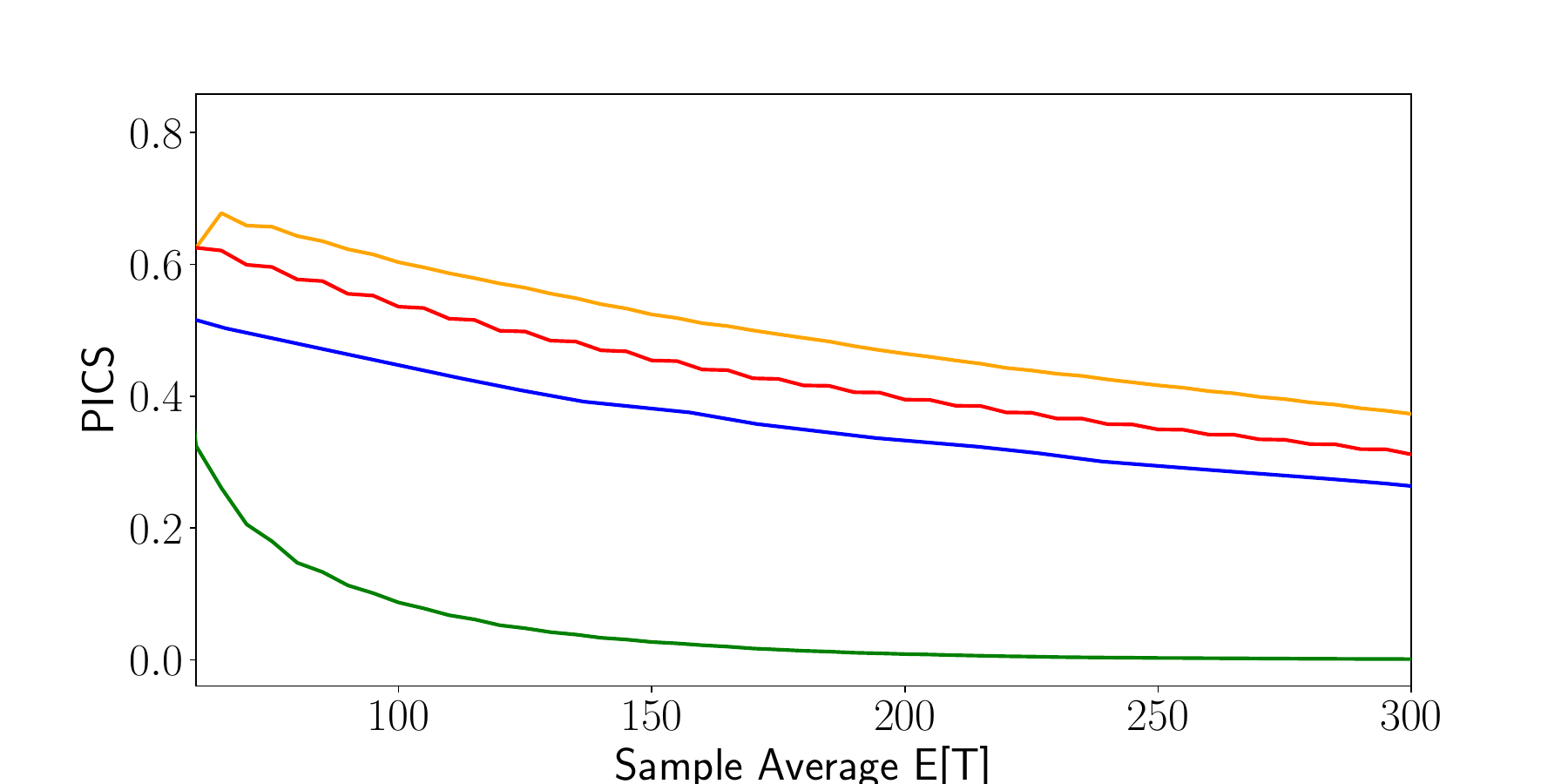}
         \caption{Worst-case}
         \label{fig-appendix:mdm10_worst_compare}
     \end{subfigure}
          \hfill
          \begin{subfigure}{0.24\textwidth}
         \centering
         \includegraphics[width=\textwidth,trim={2cm 0cm 0cm 1.5cm}]{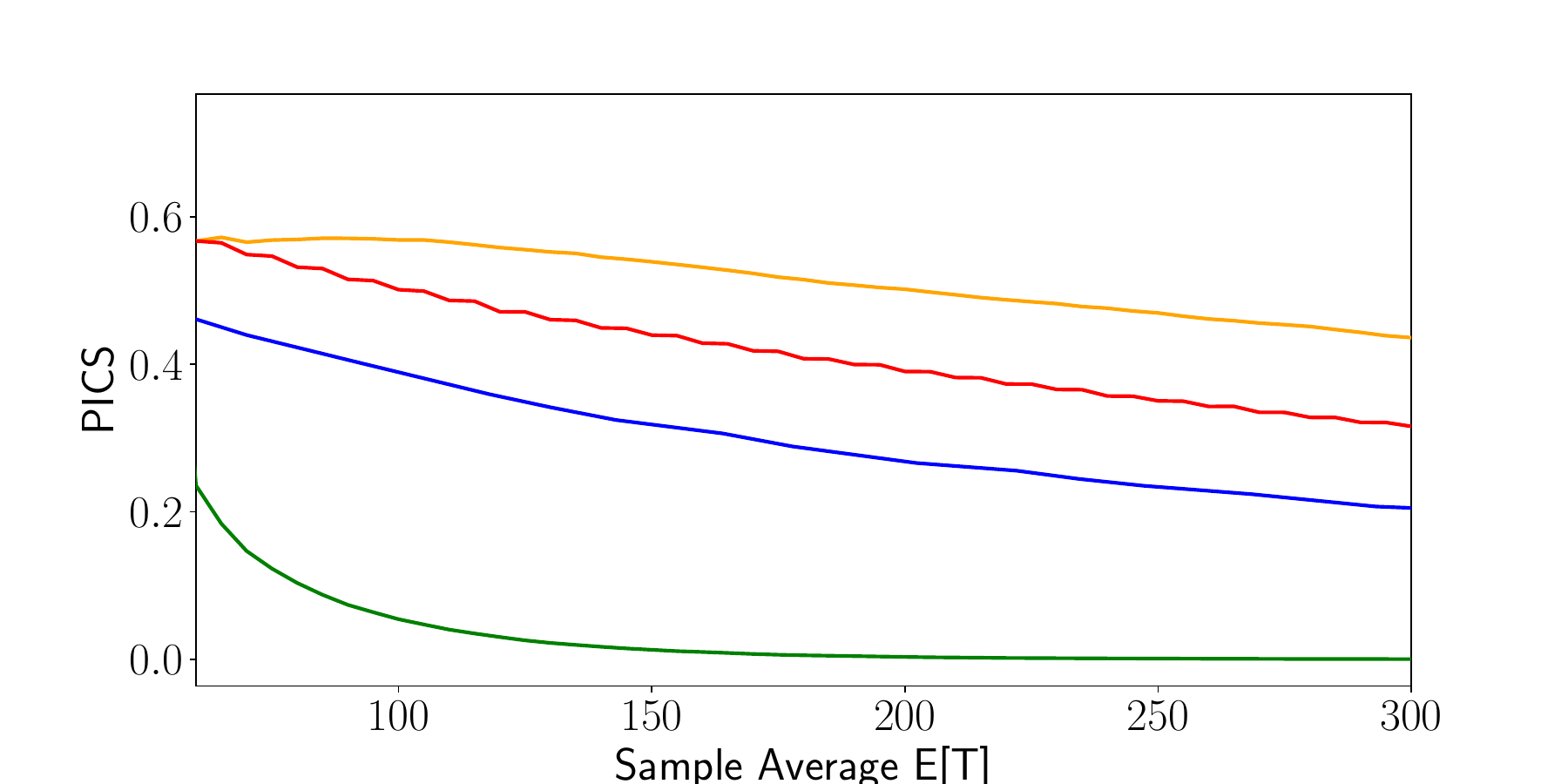}
         \caption{Cannot-sample-all-arms}
         \label{fig-appendix:mdm10_cannot_compare}
     \end{subfigure}

     \vspace{1em}
     \begin{subfigure}{0.24\textwidth}
         \centering
         \includegraphics[width=\textwidth,trim={2cm 0cm 0cm 1.5cm}]{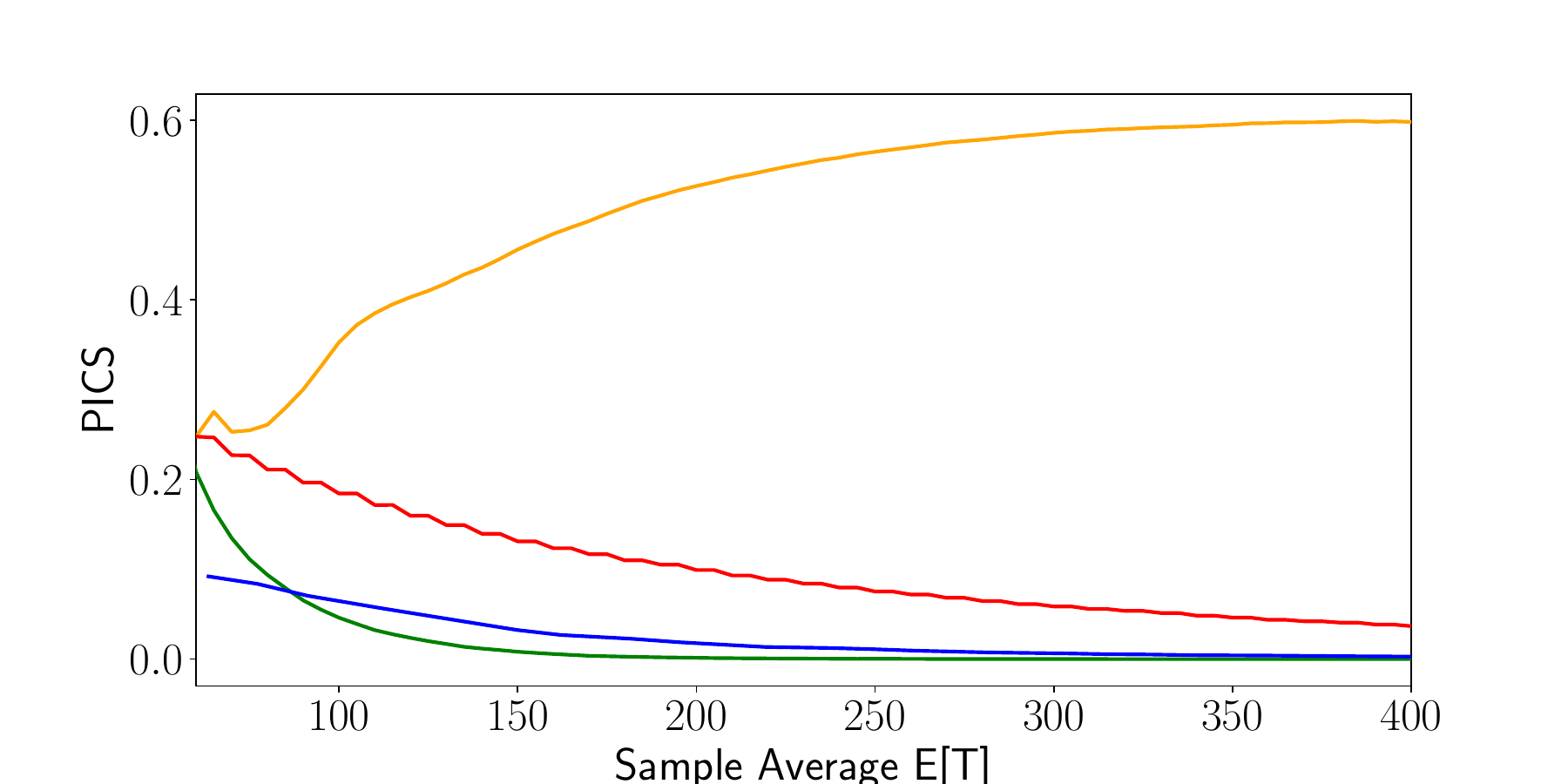}
         \caption{Sample-1to10-per-arm}
         \label{fig-appendix:mdm10_sam_compare}
     \end{subfigure}
     \hfill
     \begin{subfigure}{0.24\textwidth}
         \centering
         \includegraphics[width=\textwidth,trim={2cm 0cm 0cm 1.5cm}]{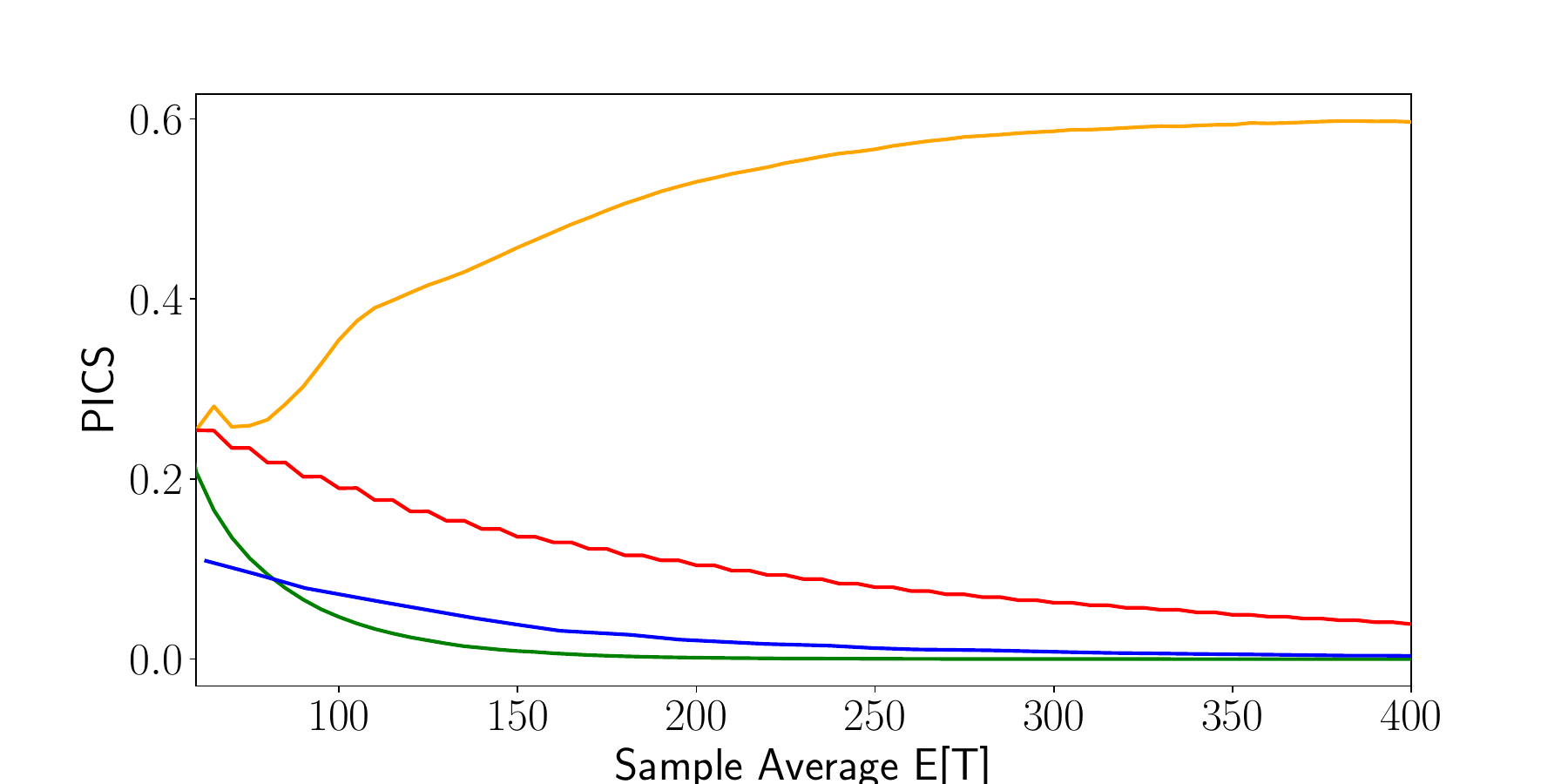}
         \caption{General}
         \label{fig-appendix:mdm10_gen_compare}
     \end{subfigure}
     \vspace{0.1em}
    \caption{MDM configurations with 10 arms}
    \label{fig-appendix:MDM_10arm_compare}
    \vspace{1em}
\end{figure}

\begin{figure}[h]
     \begin{subfigure}{0.24\textwidth}
         \centering
         \includegraphics[width=\textwidth,trim={2cm 0cm 0cm 1.5cm}]{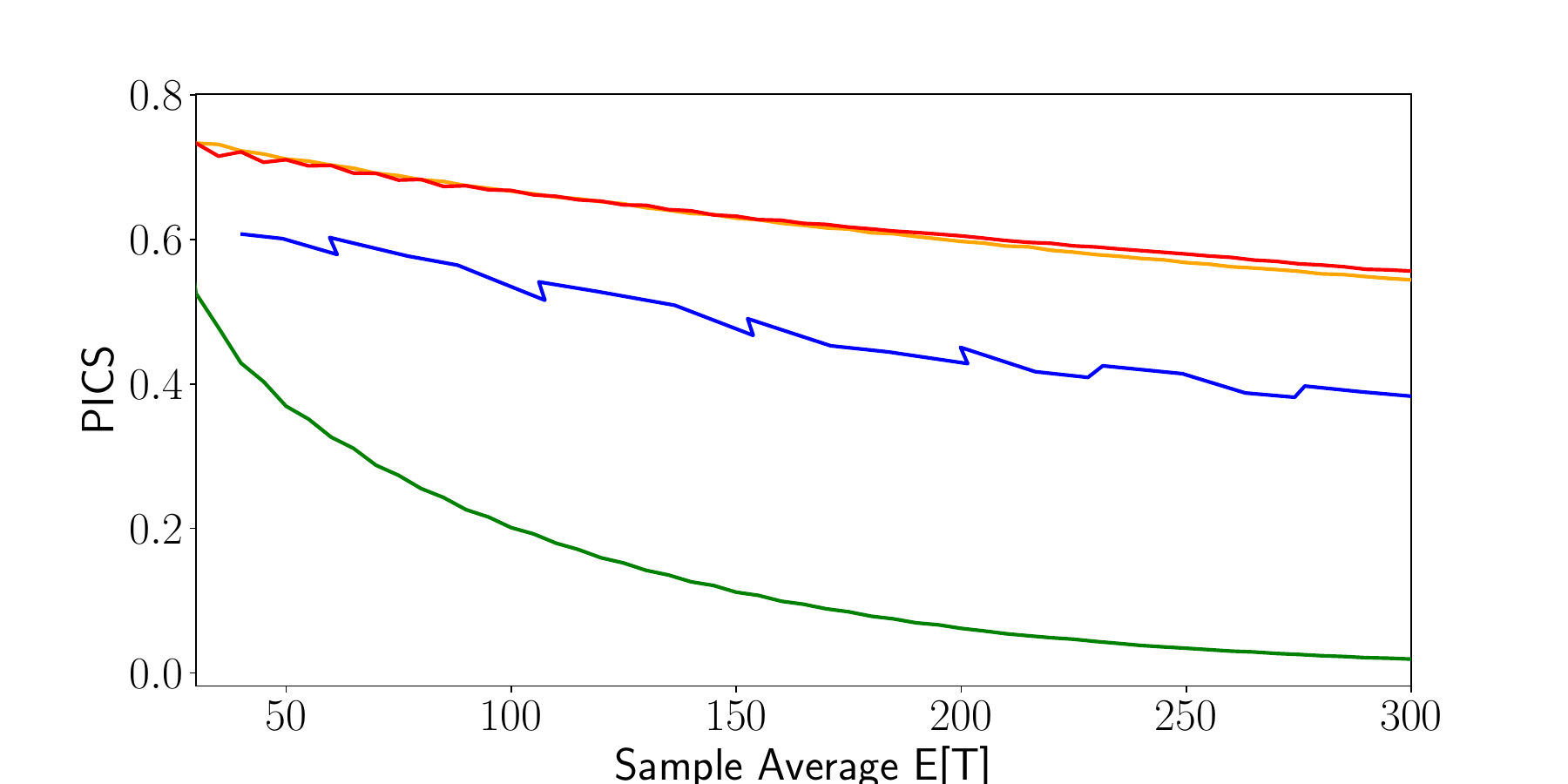}
         \caption{Worst-case}
         \label{fig-appendix:sc5_worst_compare}
     \end{subfigure}
          \hfill
          \begin{subfigure}{0.24\textwidth}
         \centering
         \includegraphics[width=\textwidth,trim={2cm 0cm 0cm 1.5cm}]{pic/ECAI-compare_cannot_SC_5arm_var1.pdf}
         \caption{Cannot-sample-all-arms}
         \label{fig-appendix:sc5_cannot_compare}
     \end{subfigure}

     \vspace{1em}
     \begin{subfigure}{0.24\textwidth}
         \centering
         \includegraphics[width=\textwidth,trim={2cm 0cm 0cm 1.5cm}]{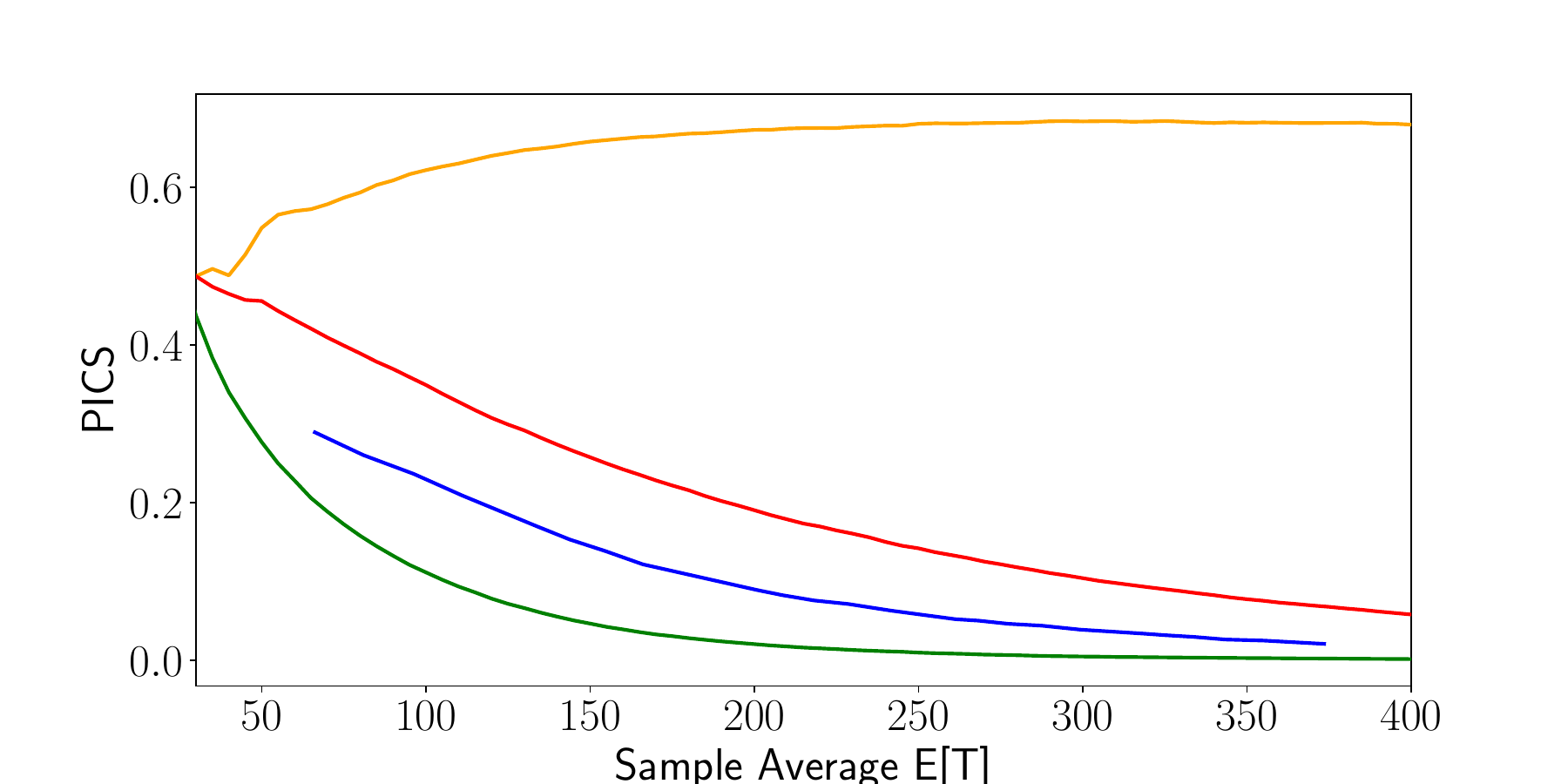}
         \caption{Sample-1to10-per-arm}
         \label{fig-appendix:sc5_sam_compare}
     \end{subfigure}
     \hfill
     \begin{subfigure}{0.24\textwidth}
         \centering
         \includegraphics[width=\textwidth,trim={2cm 0cm 0cm 1.5cm}]{pic/ECAI-compare_general_SC_5arm_var1.pdf}
         \caption{General}
         \label{fig-appendix:sc5_gen_compare}
     \end{subfigure}
     \vspace{0.1em}
    \caption{SC configurations with 5 arms}
    \label{fig-appendix:SC_5arm_compare}
    \vspace{1em}
\end{figure}

\begin{figure}[h]
     \begin{subfigure}{0.24\textwidth}
         \centering
         \includegraphics[width=\textwidth,trim={2cm 0cm 0cm 1.5cm}]{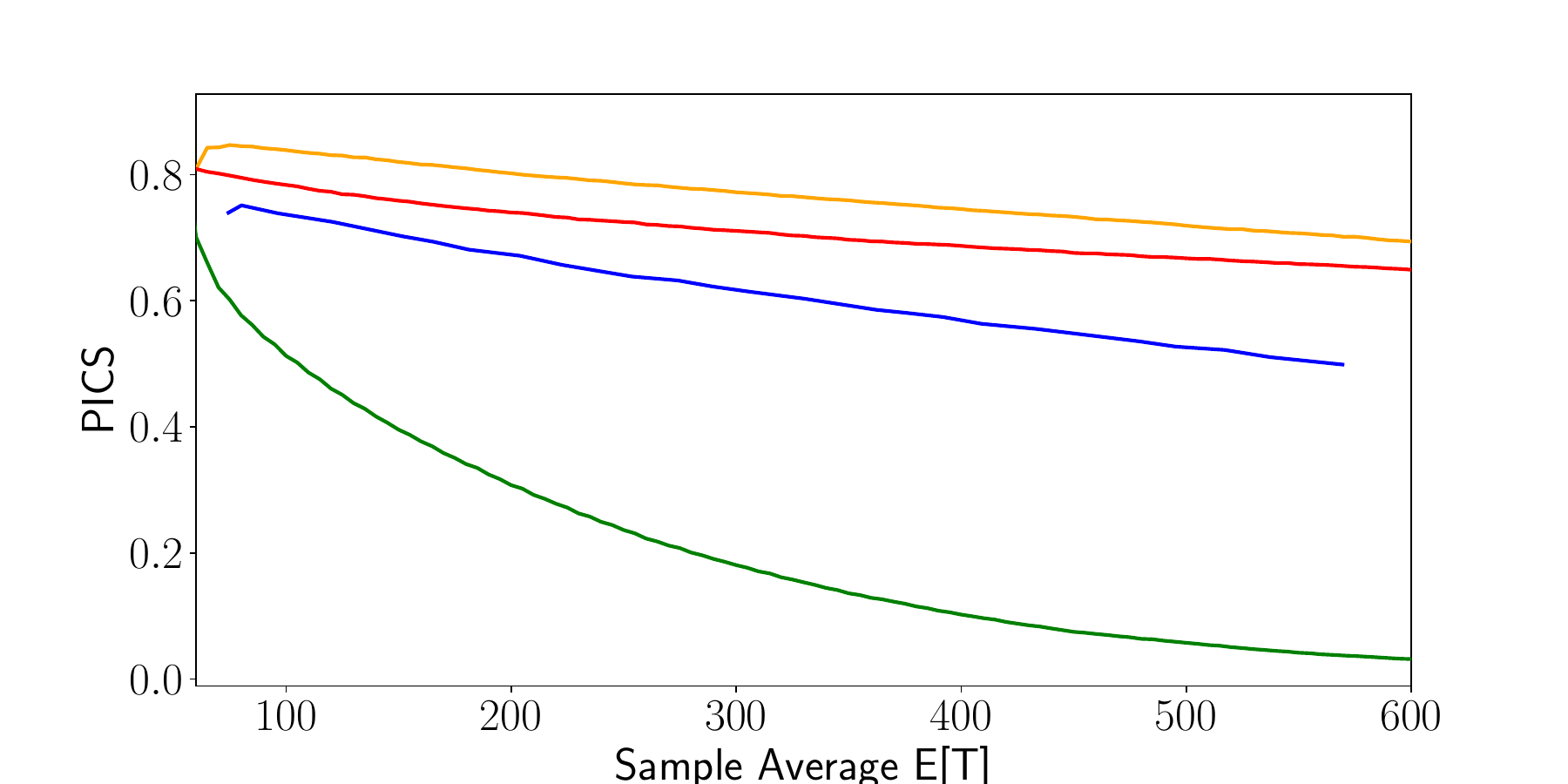}
         \caption{Worst-case}
         \label{fig-appendix:sc10_worst_compare}
     \end{subfigure}
          \hfill
          \begin{subfigure}{0.24\textwidth}
         \centering
         \includegraphics[width=\textwidth,trim={2cm 0cm 0cm 1.5cm}]{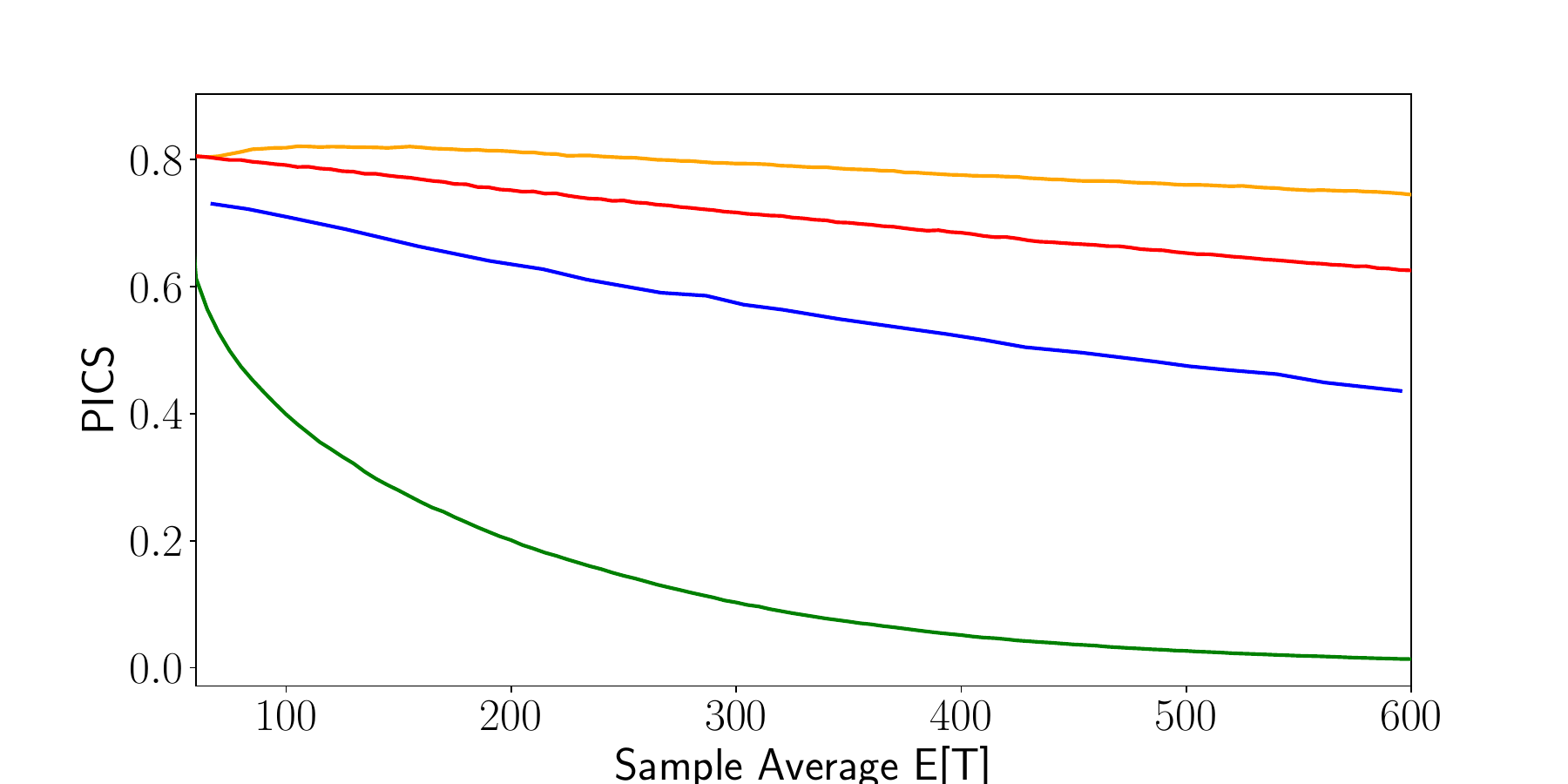}
         \caption{Cannot-sample-all-arms}
         \label{fig-appendix:sc10_cannot_compare}
     \end{subfigure}

     \vspace{1em}
     \begin{subfigure}{0.24\textwidth}
         \centering
         \includegraphics[width=\textwidth,trim={2cm 0cm 0cm 1.5cm}]{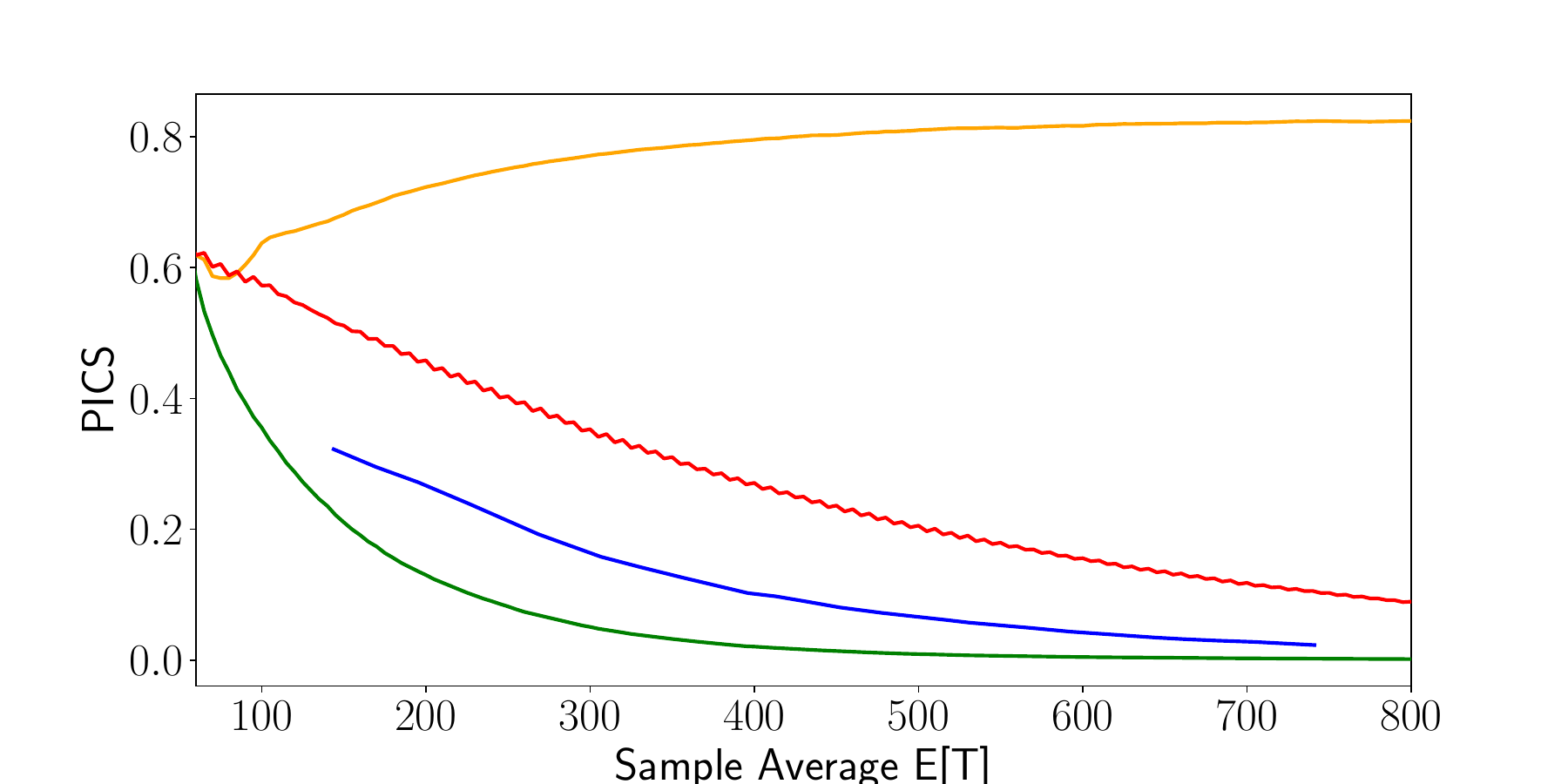}
         \caption{Sample-1to10-per-arm}
         \label{fig-appendix:SC10_sam_compare}
     \end{subfigure}
     \hfill
     \begin{subfigure}{0.24\textwidth}
         \centering
         \includegraphics[width=\textwidth,trim={2cm 0cm 0cm 1.5cm}]{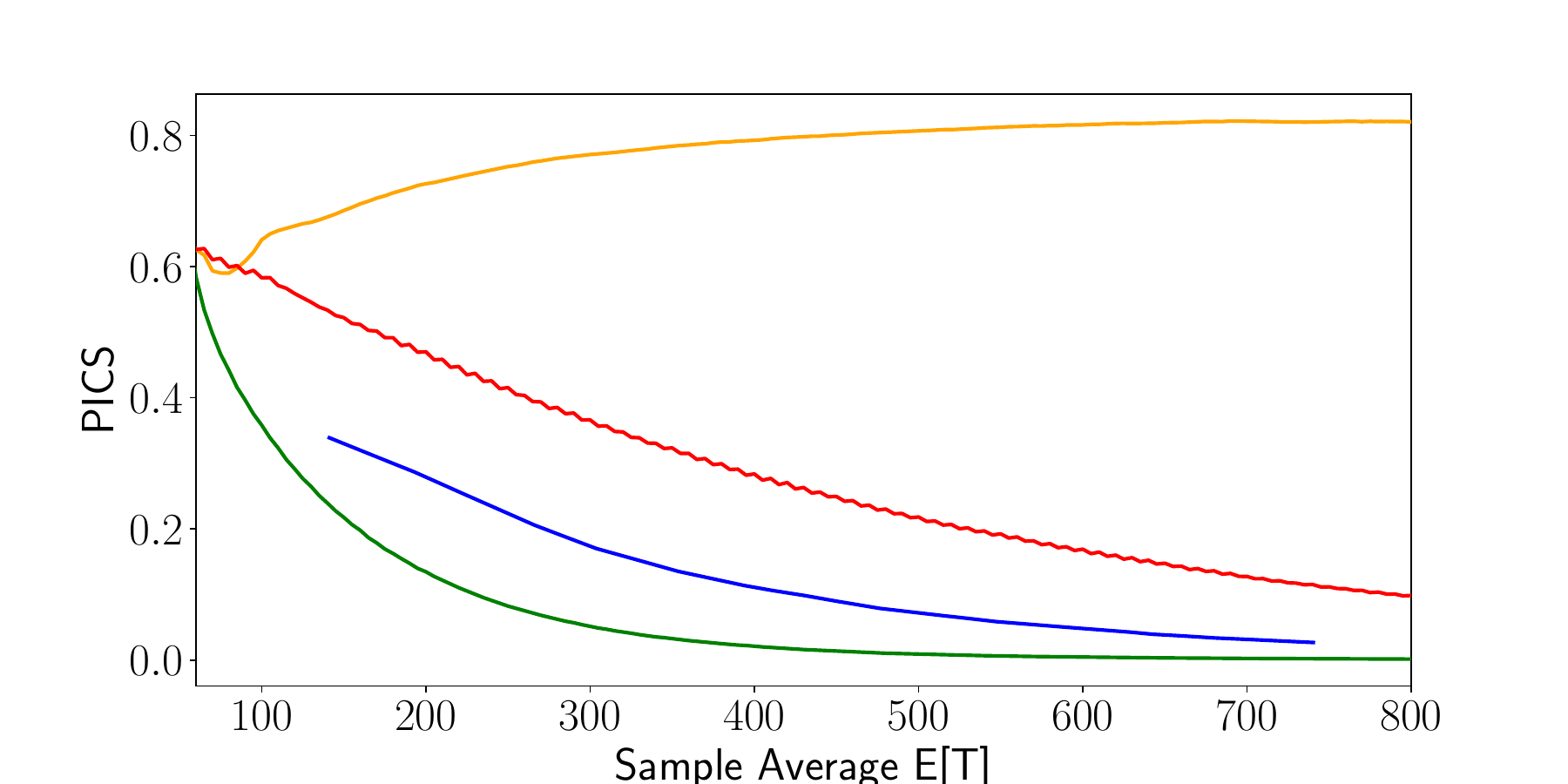}
         \caption{General}
         \label{fig-appendix:sc10_gen_compare}
     \end{subfigure}
     \vspace{0.1em}
    \caption{SC configurations with 10 arms}
    \label{fig-appendix:SC_10arm_compare}
    \vspace{2em}
\end{figure}

\vspace{0.5em}
\subsection{Measure of EOC} 
The EOC is defined by the expectation of linear loss function as follows 
\begin{align*}
     \mbox{EOC} & :=\mathbb{E}[\mu_{i^*}- \mu_{\hat{i}}]
\end{align*} where $\hat{i}$ is the arm recommended by the selection policy, and $i^*$ is the true best arm. The EOC values in all figures are approximated by the sample mean of linear loss instead. 
\begin{figure}[htbp]
    \centering
    \includegraphics[width=0.5\textwidth,trim={3.5cm 0cm 1.5cm 1.5cm}]{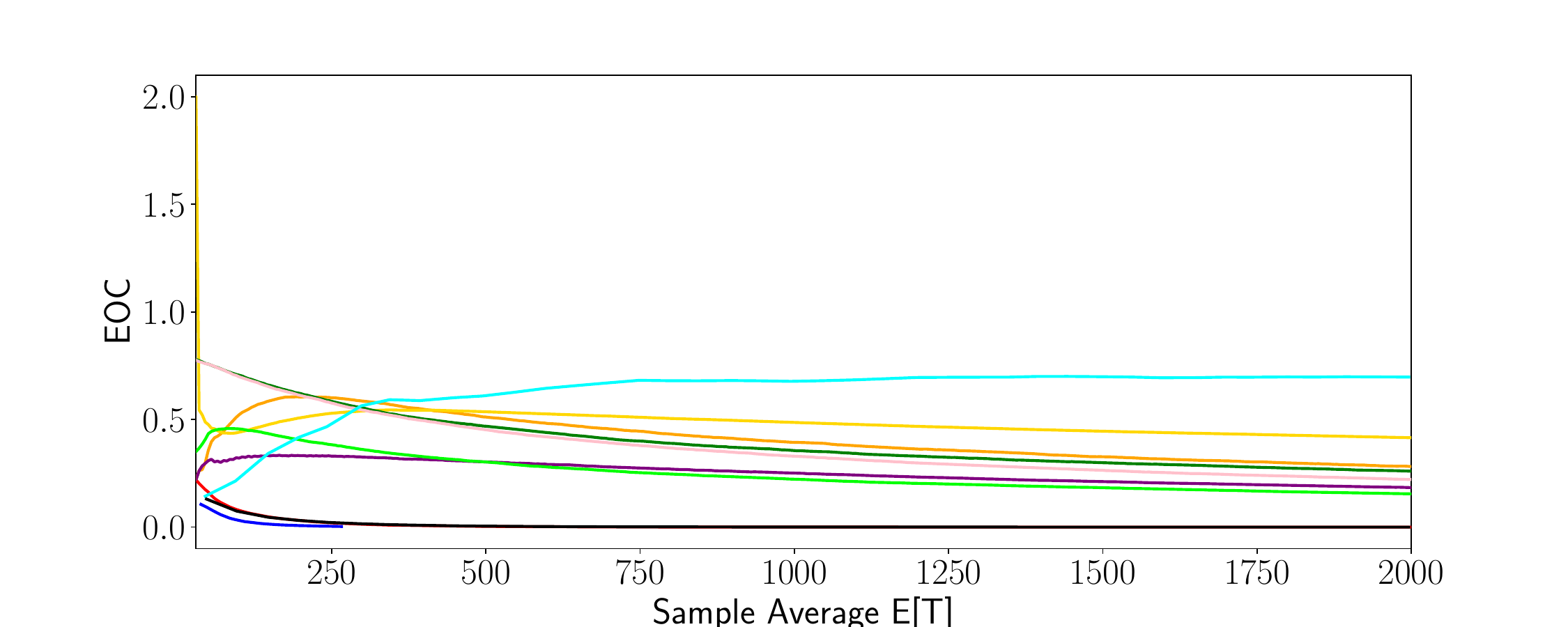}
    \includegraphics[width=0.7\textwidth,trim={3.5cm 1.5cm 2.5cm 1.5cm}]{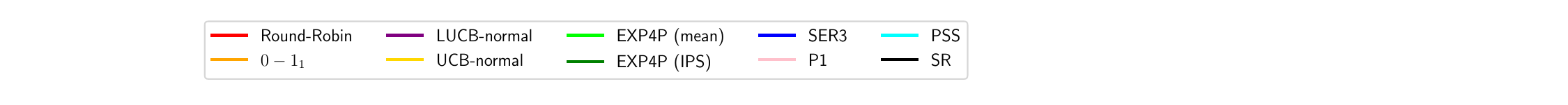}
    \vspace{0.5em}
    \caption{EOC of existing algorithms from $10^5$ replications on the Gaussian configuration of 5 arms where the gaps of ordered arms ($\delta=0.5$) are equally distributed and arms have equal variance ($\sigma=1$). The length of environment $j^{th}$ are random uniformly, $\Delta cp_{j} \sim \tilde{\mathcal{U}}(2,50)$ and the shift is a random variable, $s_j \sim \mathcal{U}(0,20)$.}
    \label{fig:homo_shift-original_alg_EOC}
    \vspace{2em}
\end{figure}

\begin{figure}[h]
     \begin{subfigure}{0.24\textwidth}
         \centering
         \vspace{-2em}\includegraphics[width=\textwidth,trim={2cm 0cm 0cm 1.5cm}]{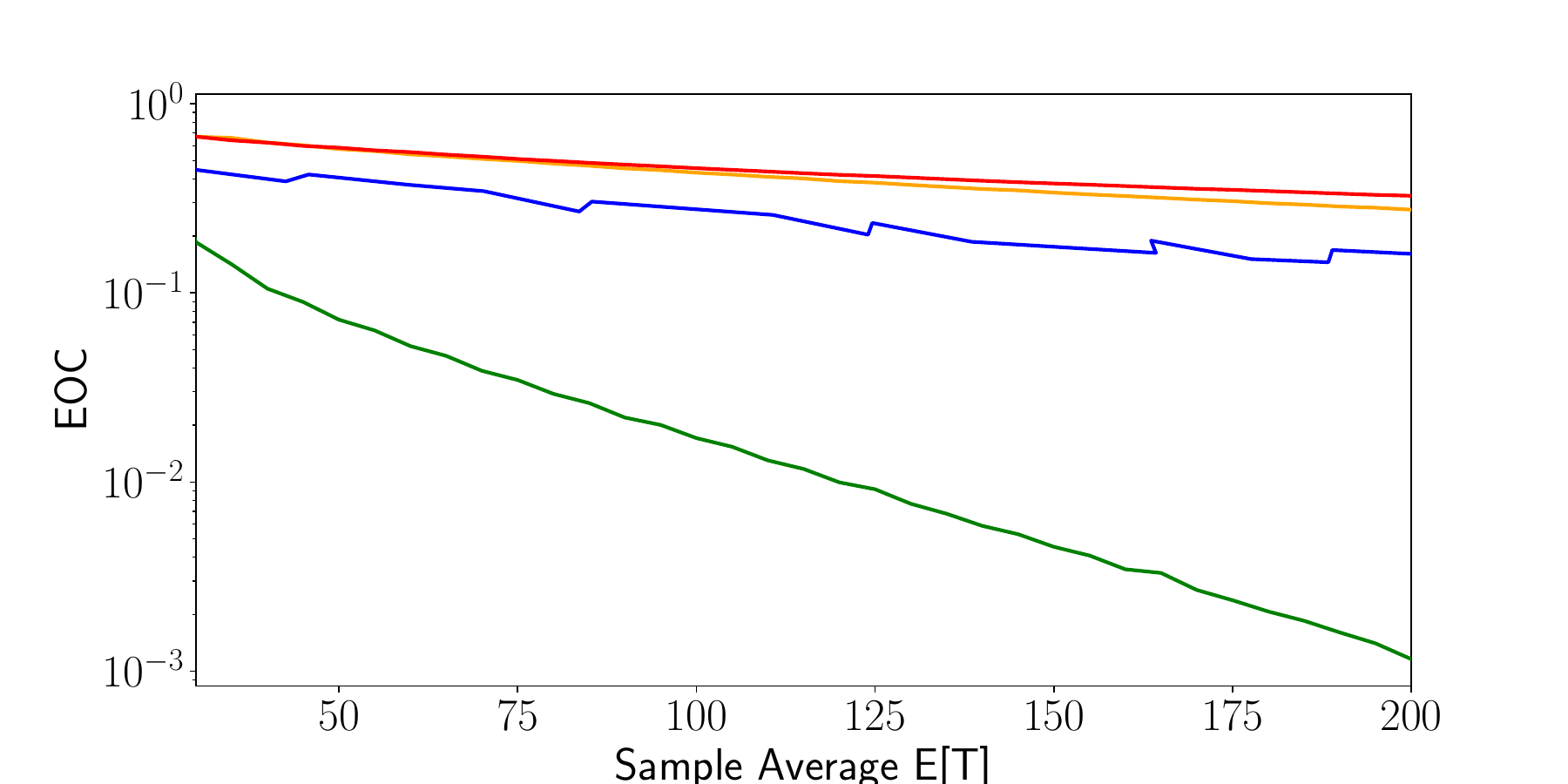}
         \caption{Worst-case}
         \label{fig-appendix:mdm5_worst_compare}
     \end{subfigure}
          \hfill
          \begin{subfigure}{0.24\textwidth}
         \centering
         \includegraphics[width=\textwidth,trim={2cm 0cm 0cm 1.5cm}]{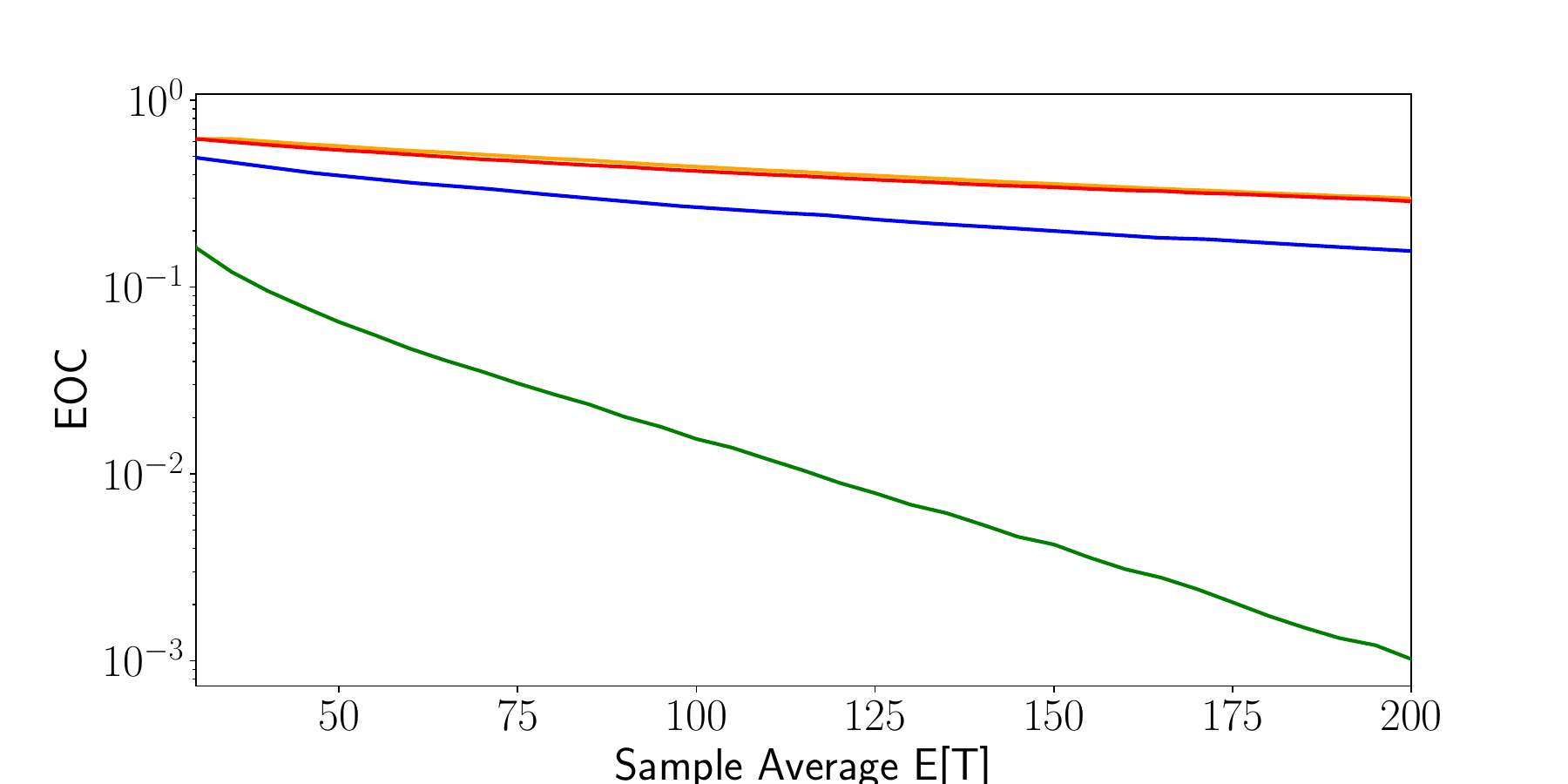}
         \caption{Cannot-sample-all-arms}
         \label{fig-appendix:mdm5_cannot_compare}
     \end{subfigure}

     \vspace{1em}
     \begin{subfigure}{0.24\textwidth}
         \centering
         \includegraphics[width=\textwidth,trim={2cm 0cm 0cm 1.5cm}]{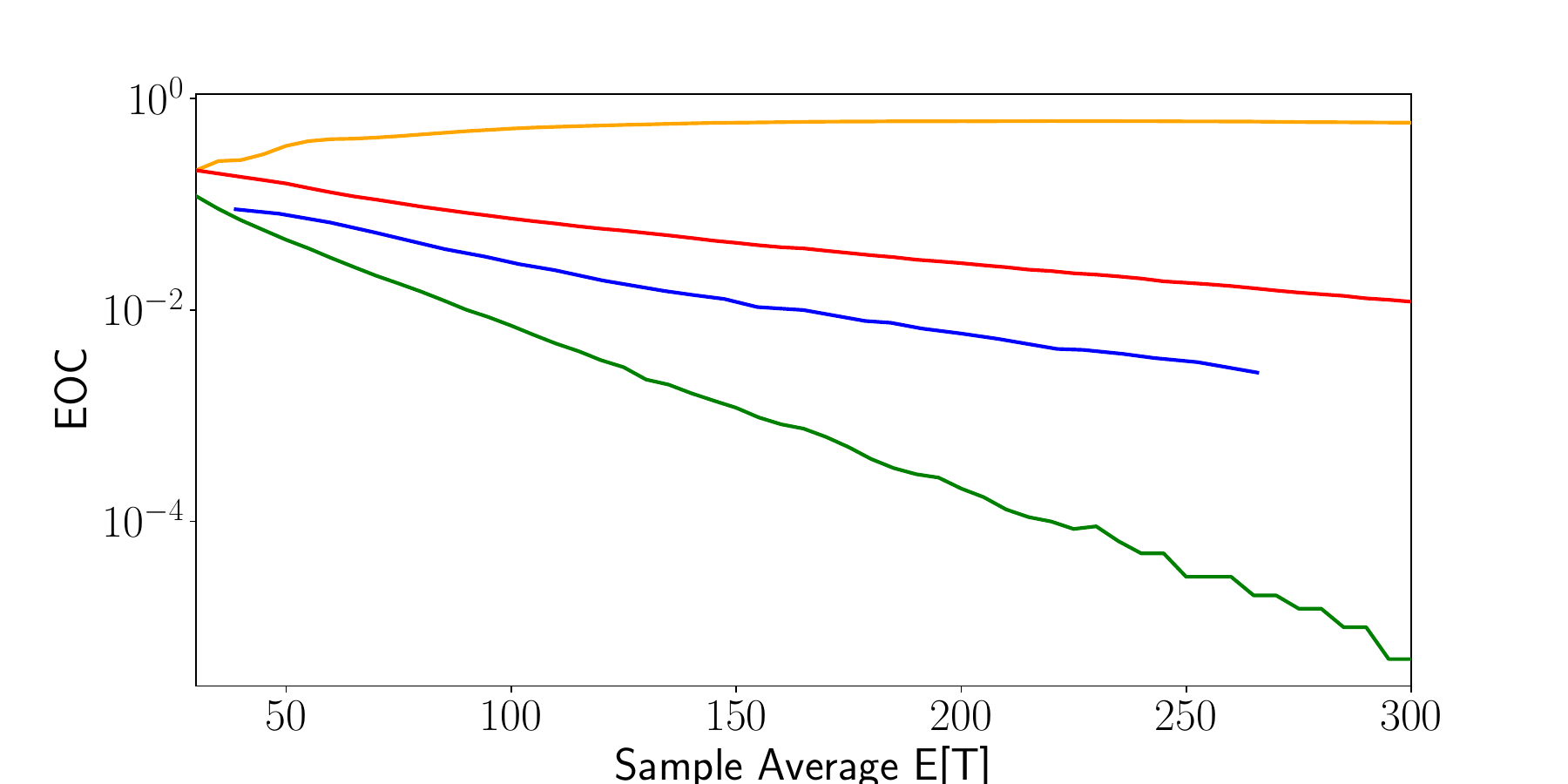}
         \caption{Sample-1to10-per-arm}
         \label{fig-appendix:mdm5_sam_compare}
     \end{subfigure}
     \hfill
     \begin{subfigure}{0.24\textwidth}
         \centering
         \includegraphics[width=\textwidth,trim={2cm 0cm 0cm 1.5cm}]{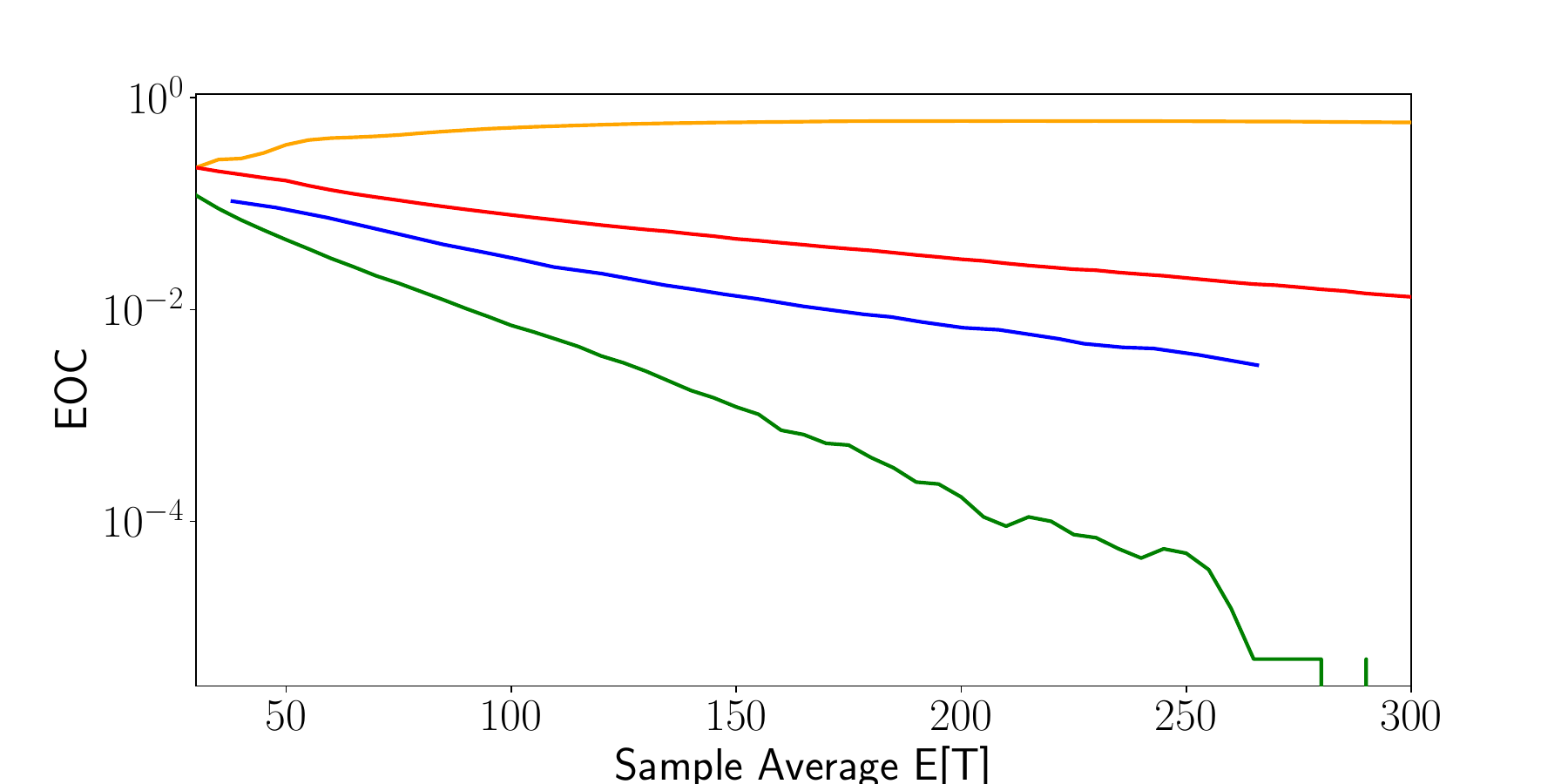}
         \caption{General}
         \label{fig-appendix:mdm5_gen_compare}
     \end{subfigure}
     \vspace{0.1em}
    \caption{MDM configurations with 5 arms}
    \label{fig-appendix:MDM_5arm_compare}
    \vspace{1em}
\end{figure}

\begin{figure}[h]
     \begin{subfigure}{0.24\textwidth}
         \centering
         \includegraphics[width=\textwidth,trim={2cm 0cm 0cm 1.5cm}]{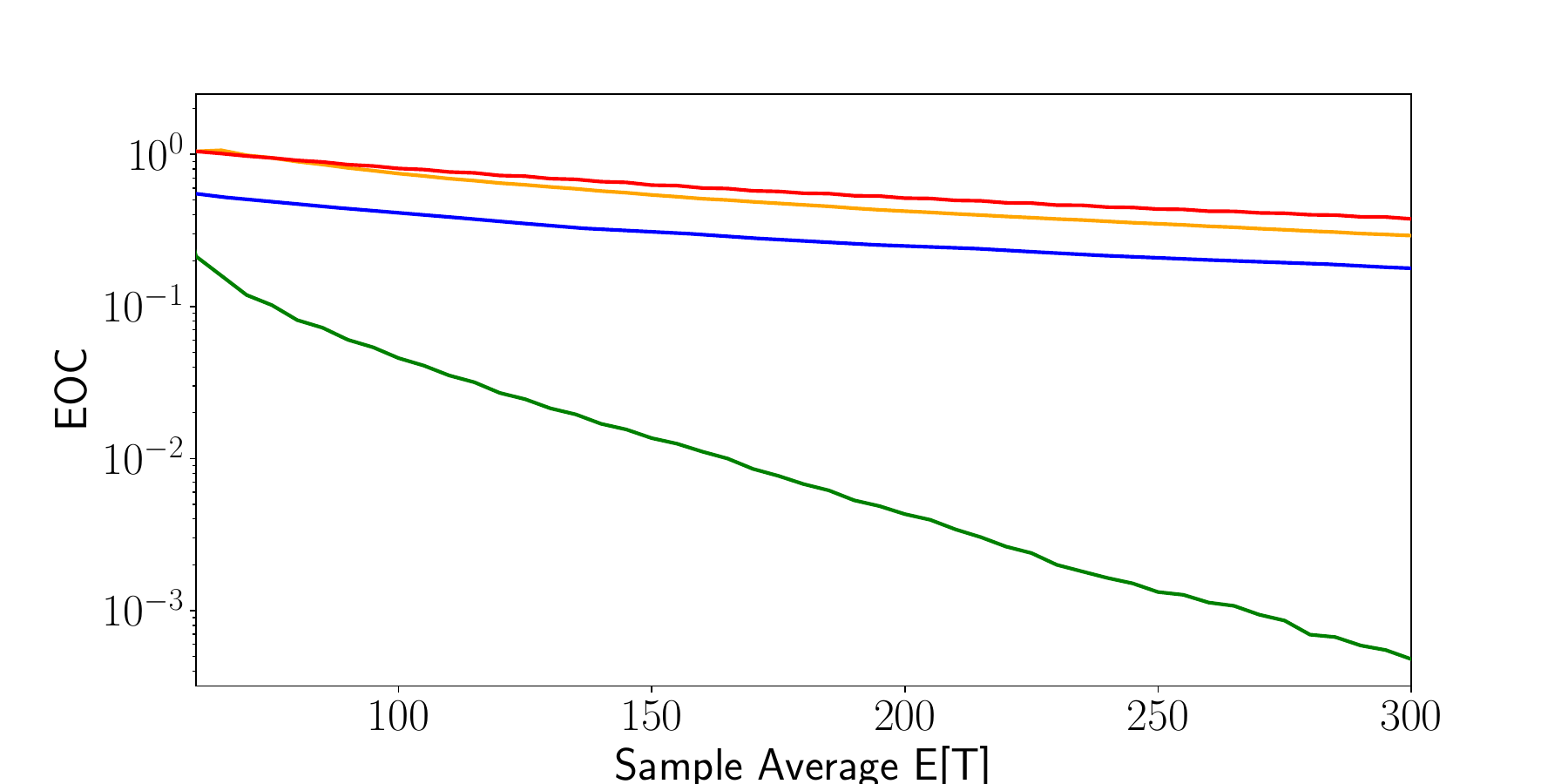}
         \caption{Worst-case}
         \label{fig-appendix:mdm10_worst_compare}
     \end{subfigure}
          \hfill
          \begin{subfigure}{0.24\textwidth}
         \centering
         \includegraphics[width=\textwidth,trim={2cm 0cm 0cm 1.5cm}]{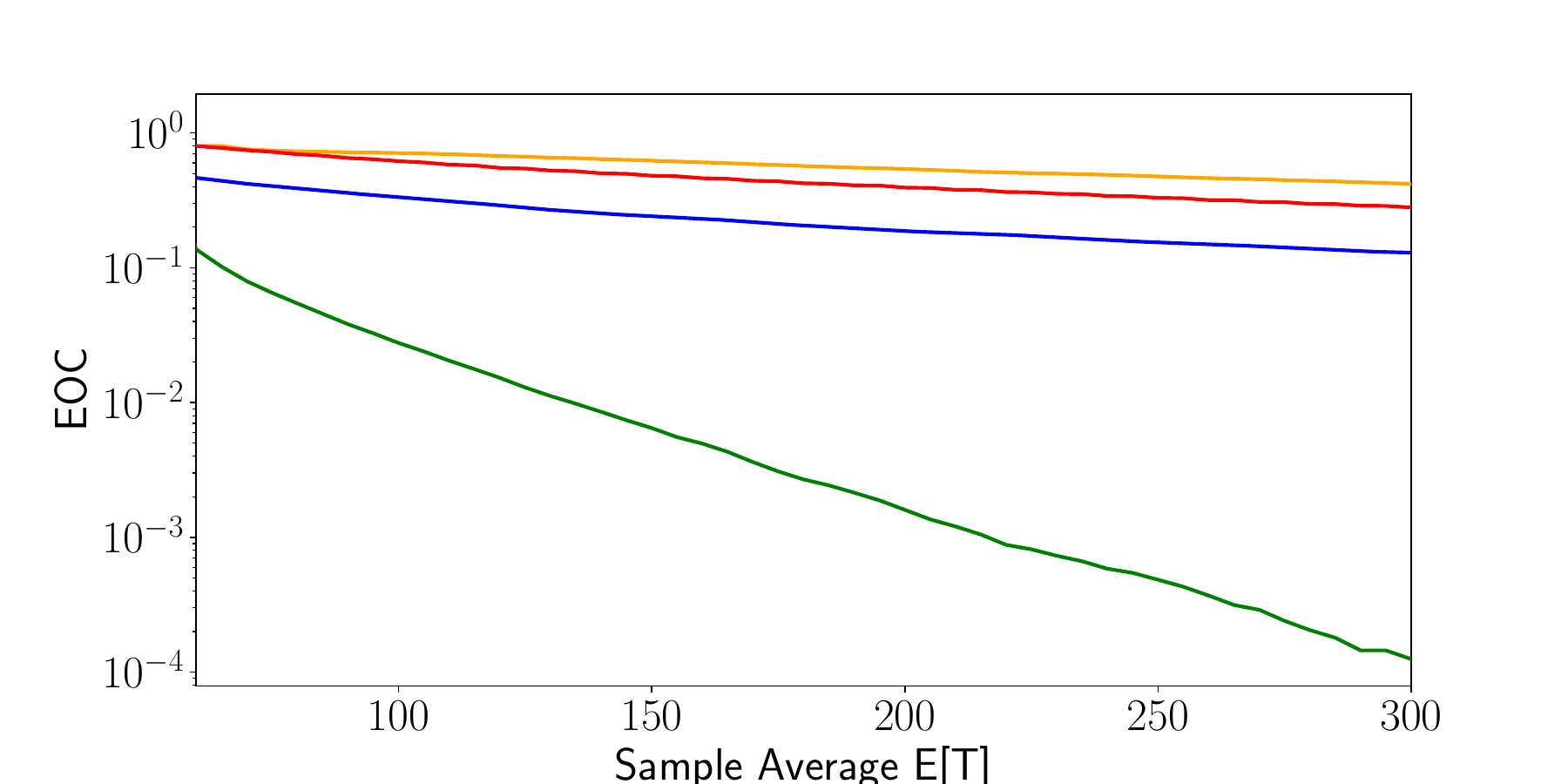}
         \caption{Cannot-sample-all-arms}
         \label{fig-appendix:mdm10_cannot_compare}
     \end{subfigure}

     \vspace{1em}
     \begin{subfigure}{0.24\textwidth}
         \centering
         \includegraphics[width=\textwidth,trim={2cm 0cm 0cm 1.5cm}]{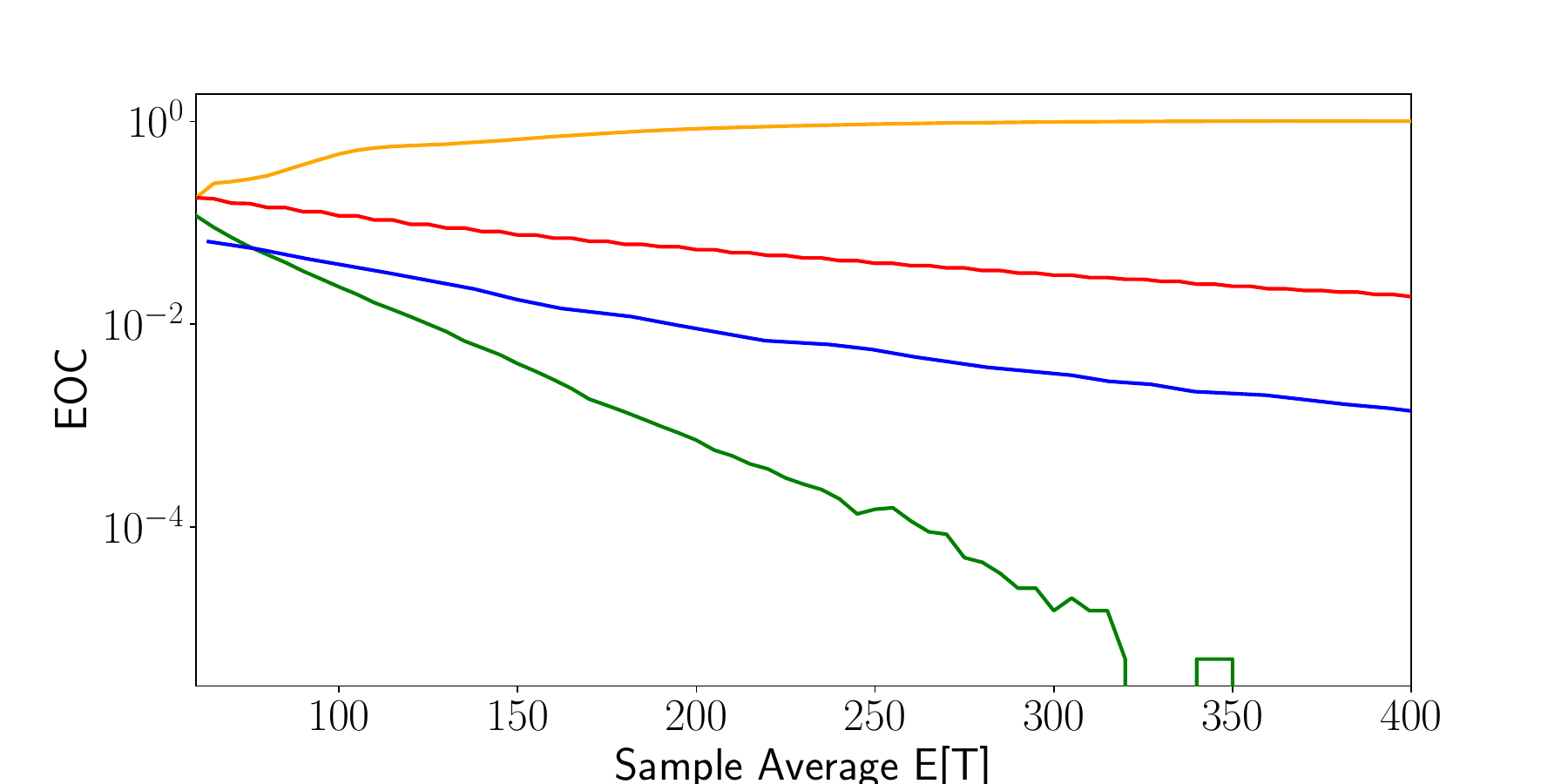}
         \caption{Sample-1to10-per-arm}
         \label{fig-appendix:mdm10_sam_compare}
     \end{subfigure}
     \hfill
     \begin{subfigure}{0.24\textwidth}
         \centering
         \includegraphics[width=\textwidth,trim={2cm 0cm 0cm 1.5cm}]{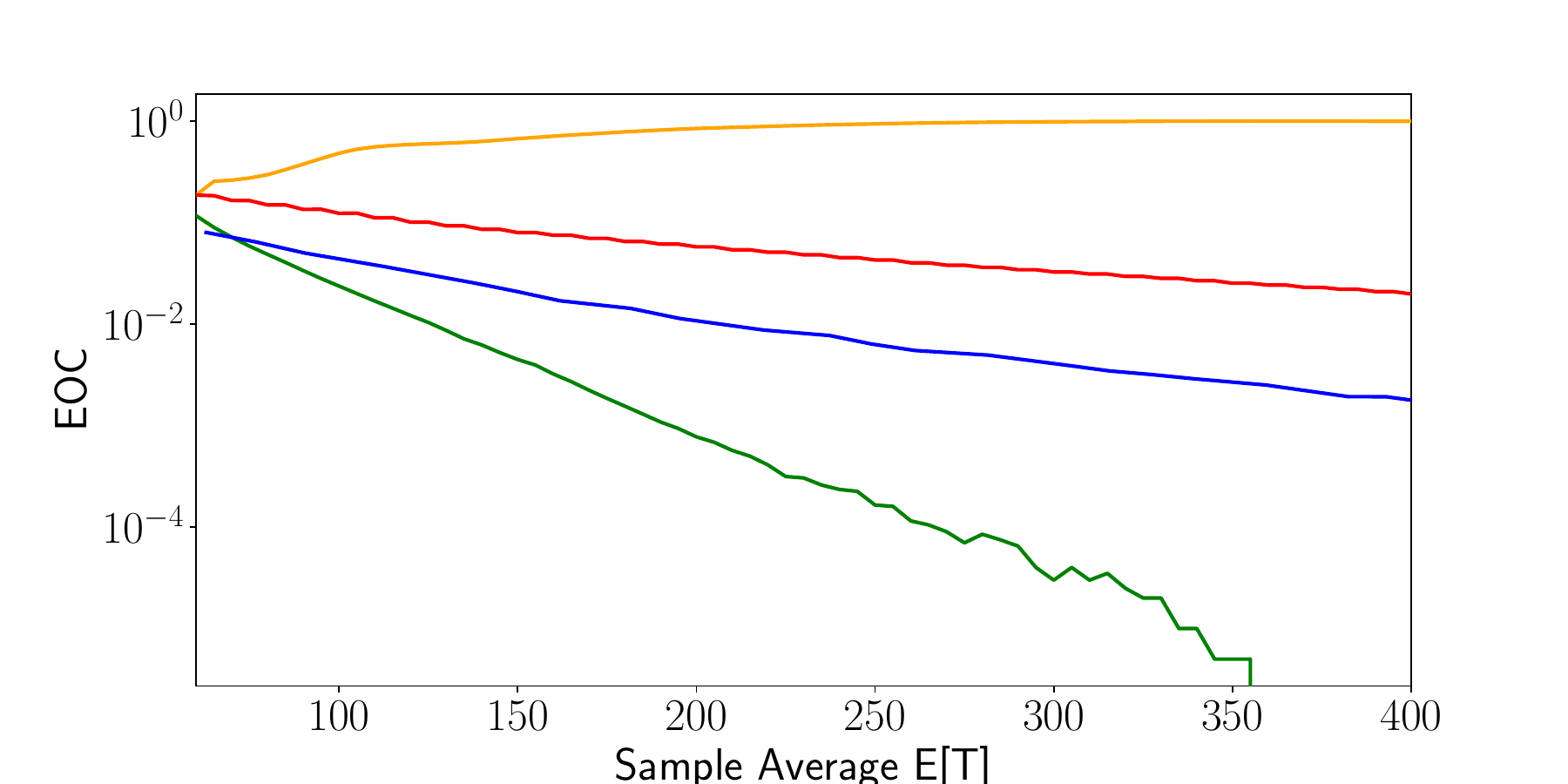}
         \caption{General}
         \label{fig-appendix:mdm10_gen_compare}
     \end{subfigure}
     \vspace{0.1em}
    \caption{MDM configurations with 10 arms}
    \label{fig-appendix:MDM_10arm_compare}
    \vspace{1em}
\end{figure}

\begin{figure}[h]
     \begin{subfigure}{0.24\textwidth}
         \centering
         \includegraphics[width=\textwidth,trim={2cm 0cm 0cm 1.5cm}]{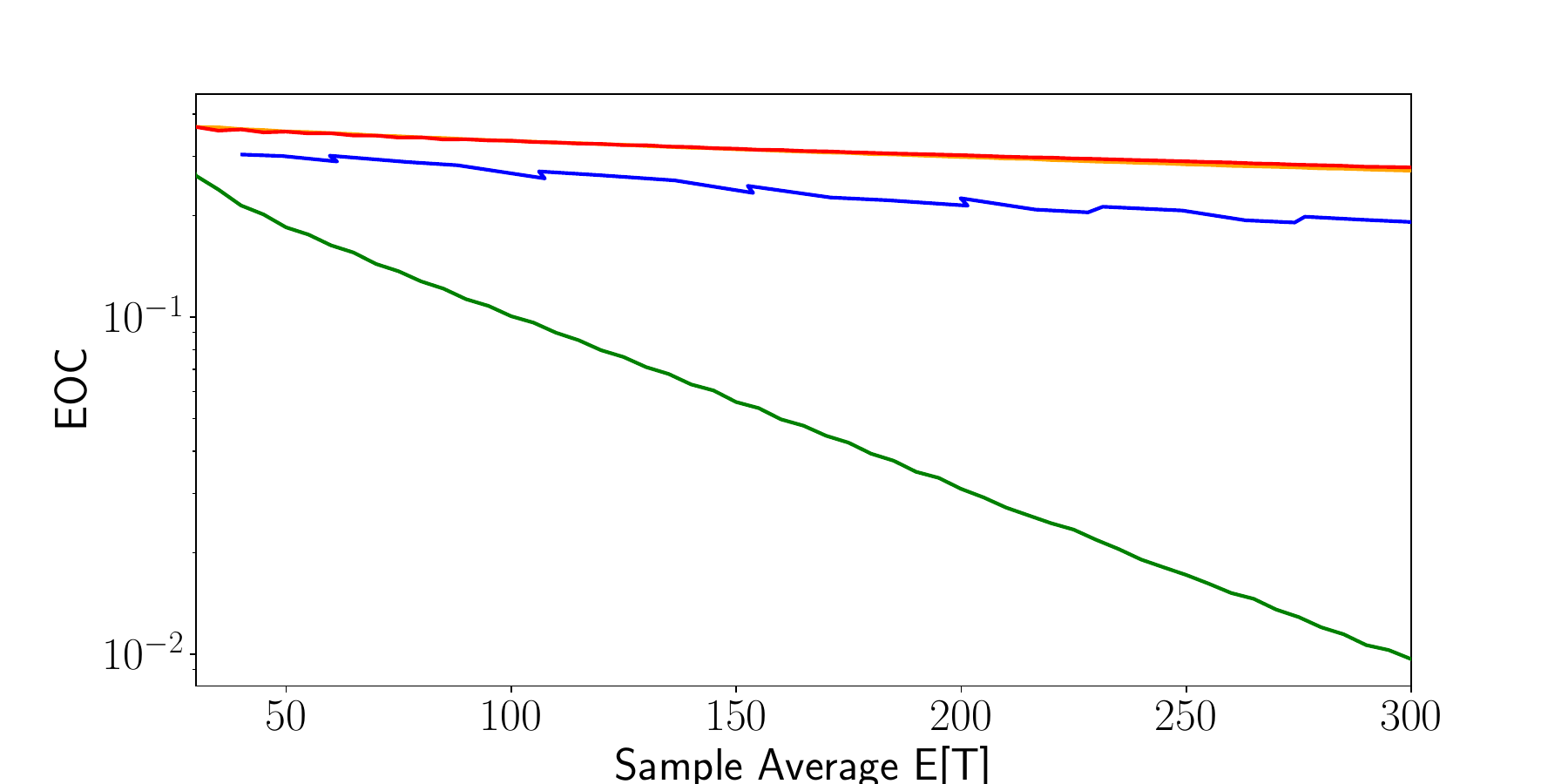}
         \caption{Worst-case}
         \label{fig-appendix:sc5_worst_compare}
     \end{subfigure}
          \hfill
          \begin{subfigure}{0.24\textwidth}
         \centering
         \includegraphics[width=\textwidth,trim={2cm 0cm 0cm 1.5cm}]{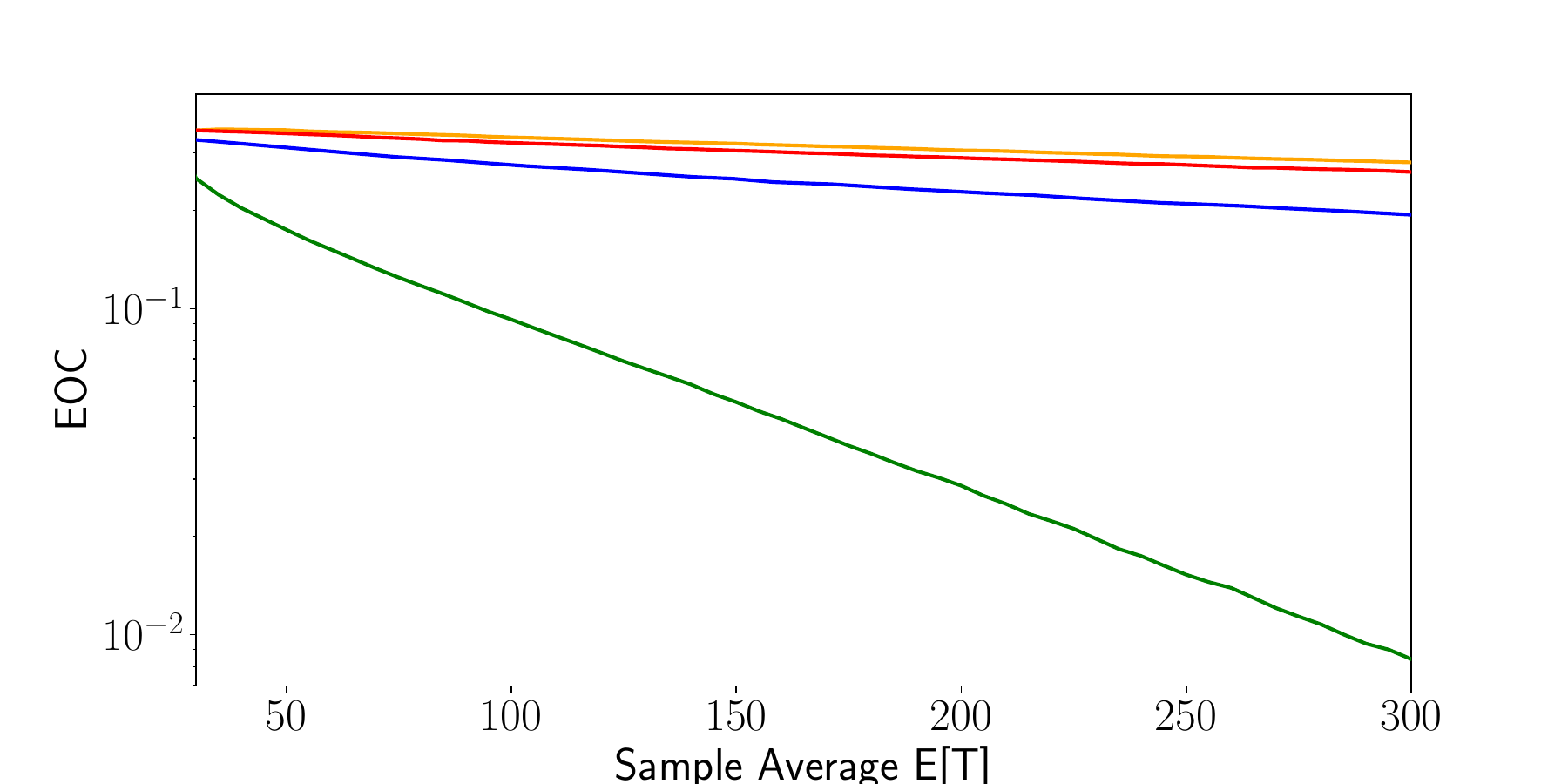}
         \caption{Cannot-sample-all-arms}
         \label{fig-appendix:sc5_cannot_compare}
     \end{subfigure}

     \vspace{1em}
     \begin{subfigure}{0.24\textwidth}
         \centering
         \includegraphics[width=\textwidth,trim={2cm 0cm 0cm 1.5cm}]{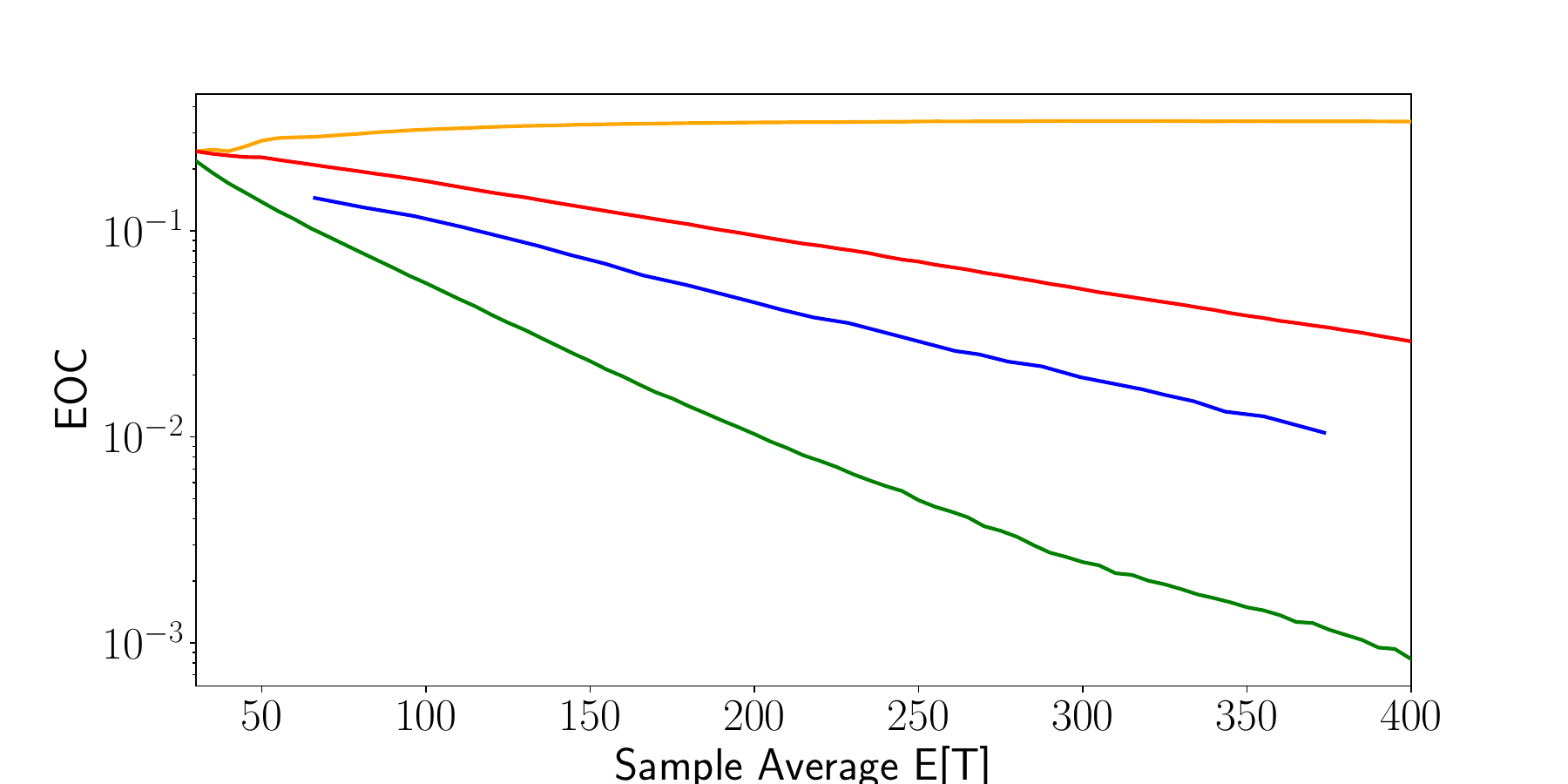}
         \caption{Sample-1to10-per-arm}
         \label{fig-appendix:sc5_sam_compare}
     \end{subfigure}
     \hfill
     \begin{subfigure}{0.24\textwidth}
         \centering
         \includegraphics[width=\textwidth,trim={2cm 0cm 0cm 1.5cm}]{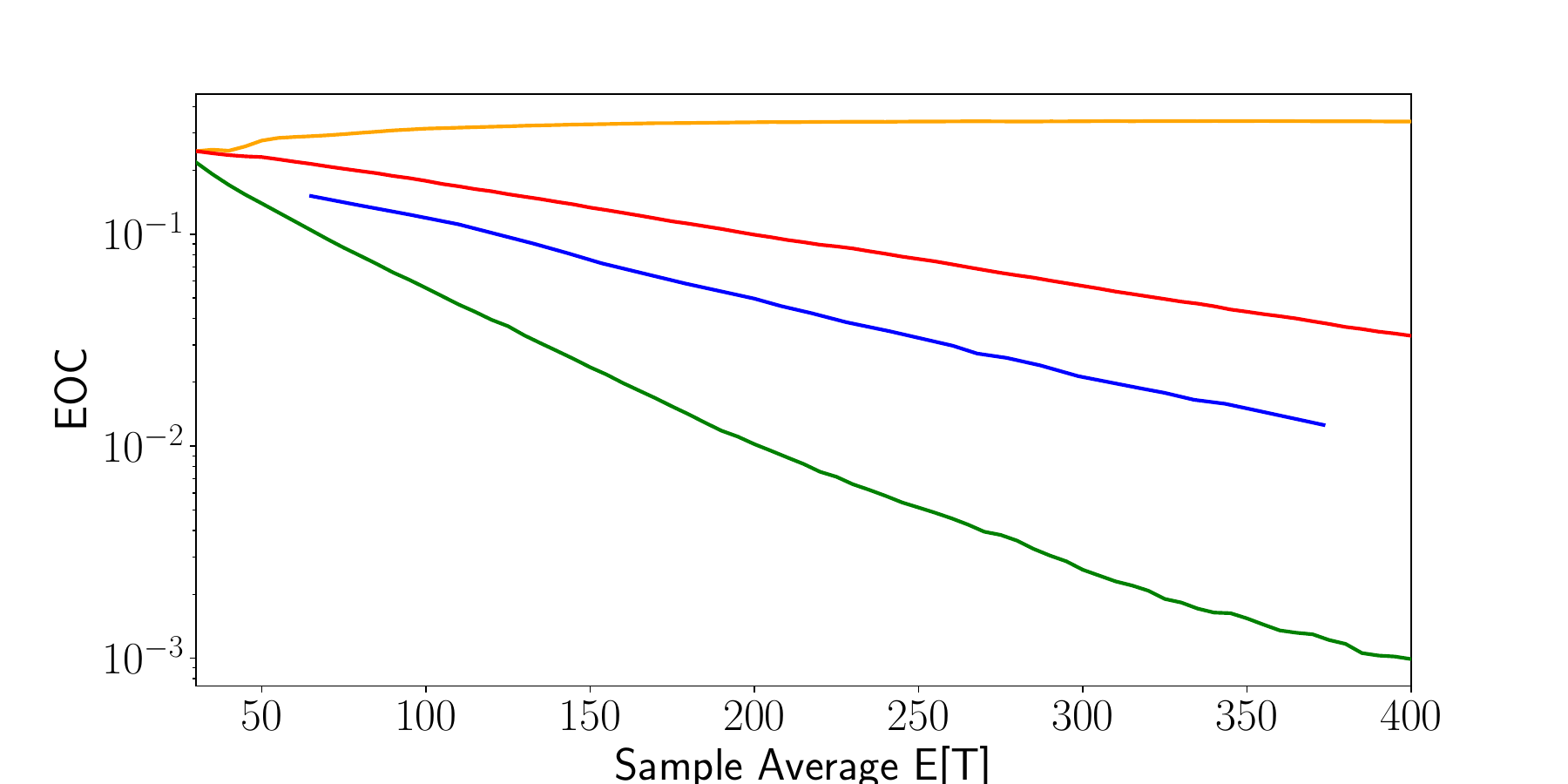}
         \caption{General}
         \label{fig-appendix:sc5_gen_compare}
     \end{subfigure}
     \vspace{0.1em}
    \caption{SC configurations with 5 arms}
    \label{fig-appendix:SC_5arm_compare}
    \vspace{1em}
\end{figure}

\begin{figure}[h]
     \begin{subfigure}{0.24\textwidth}
         \centering
         \includegraphics[width=\textwidth,trim={2cm 0cm 0cm 1.5cm}]{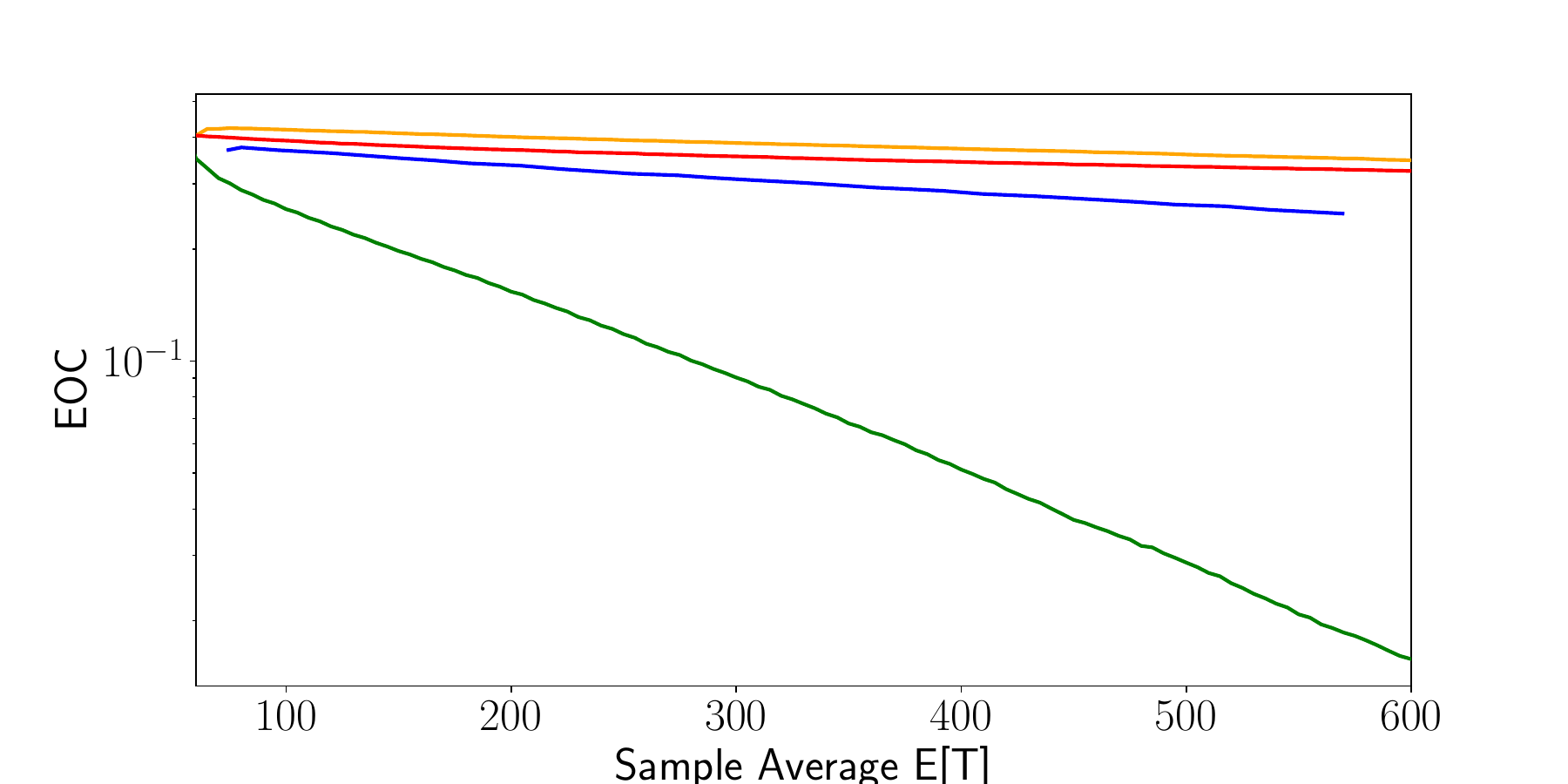}
         \caption{Worst-case}
         \label{fig-appendix:sc10_worst_compare}
     \end{subfigure}
          \hfill
          \begin{subfigure}{0.24\textwidth}
         \centering
         \includegraphics[width=\textwidth,trim={2cm 0cm 0cm 1.5cm}]{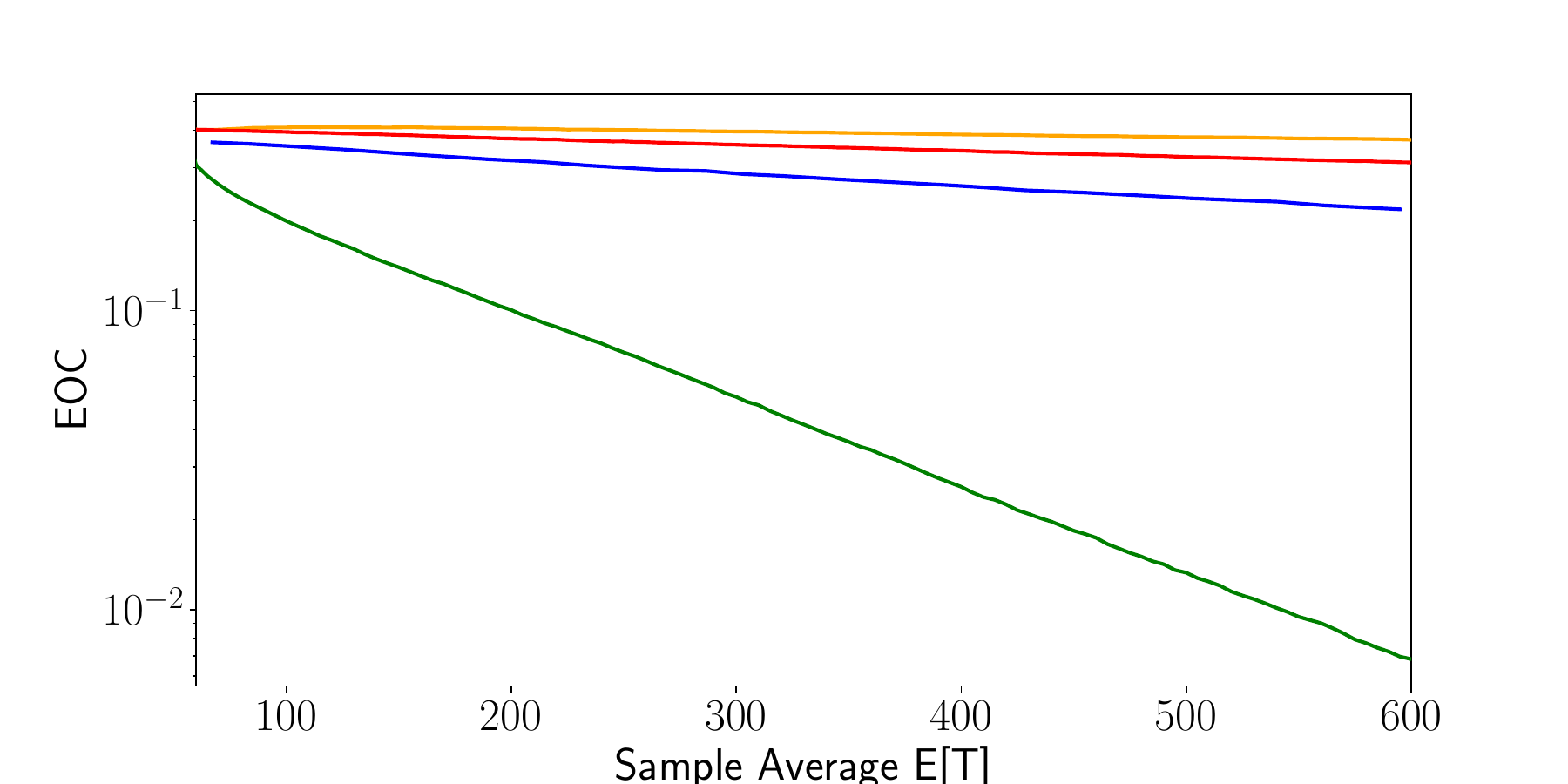}
         \caption{Cannot-sample-all-arms}
         \label{fig-appendix:sc10_cannot_compare}
     \end{subfigure}

     \vspace{1em}
     \begin{subfigure}{0.24\textwidth}
         \centering
         \includegraphics[width=\textwidth,trim={2cm 0cm 0cm 1.5cm}]{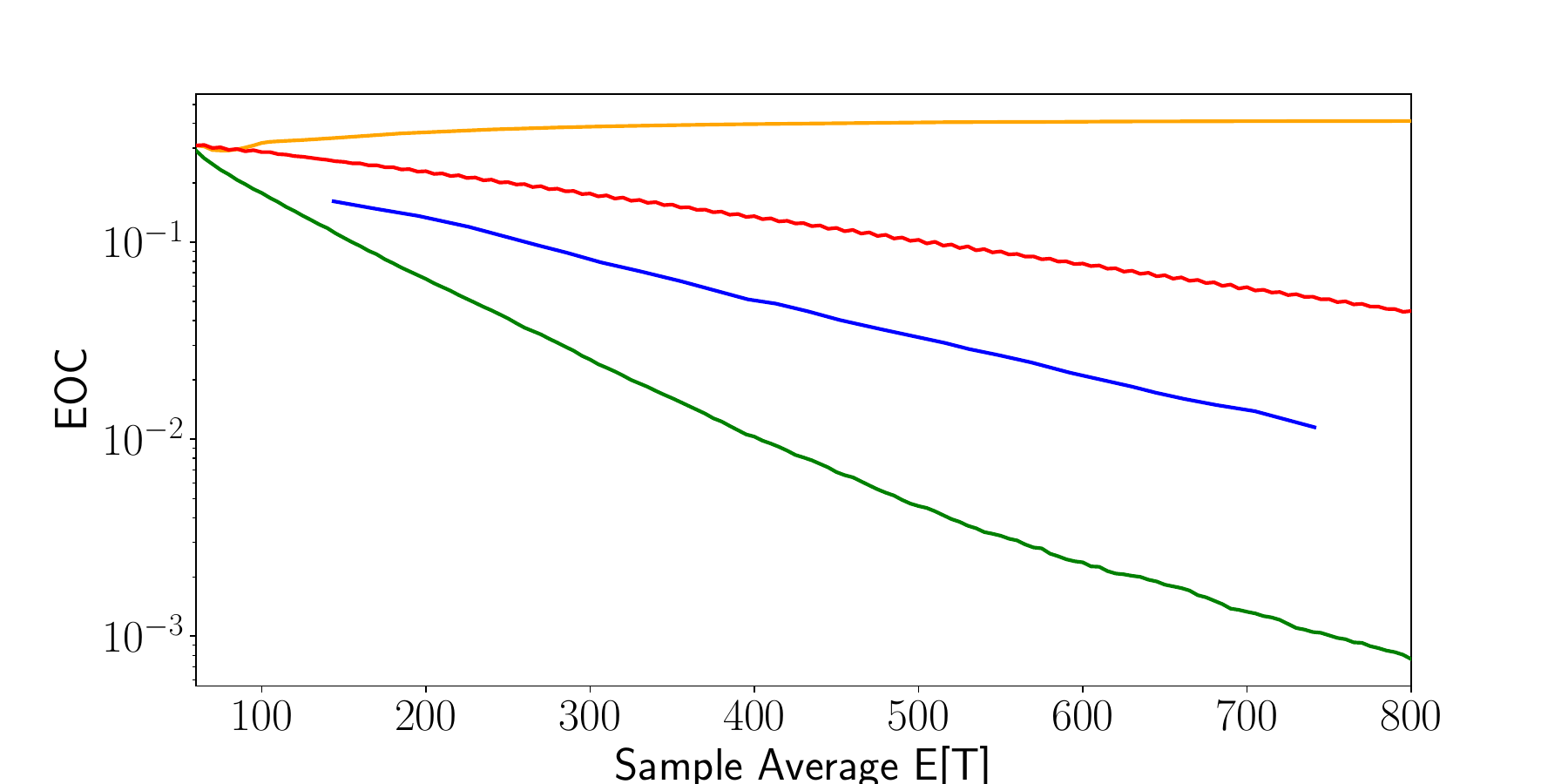}
         \caption{Sample-1to10-per-arm}
         \label{fig-appendix:SC10_sam_compare}
     \end{subfigure}
     \hfill
     \begin{subfigure}{0.24\textwidth}
         \centering
         \includegraphics[width=\textwidth,trim={2cm 0cm 0cm 1.5cm}]{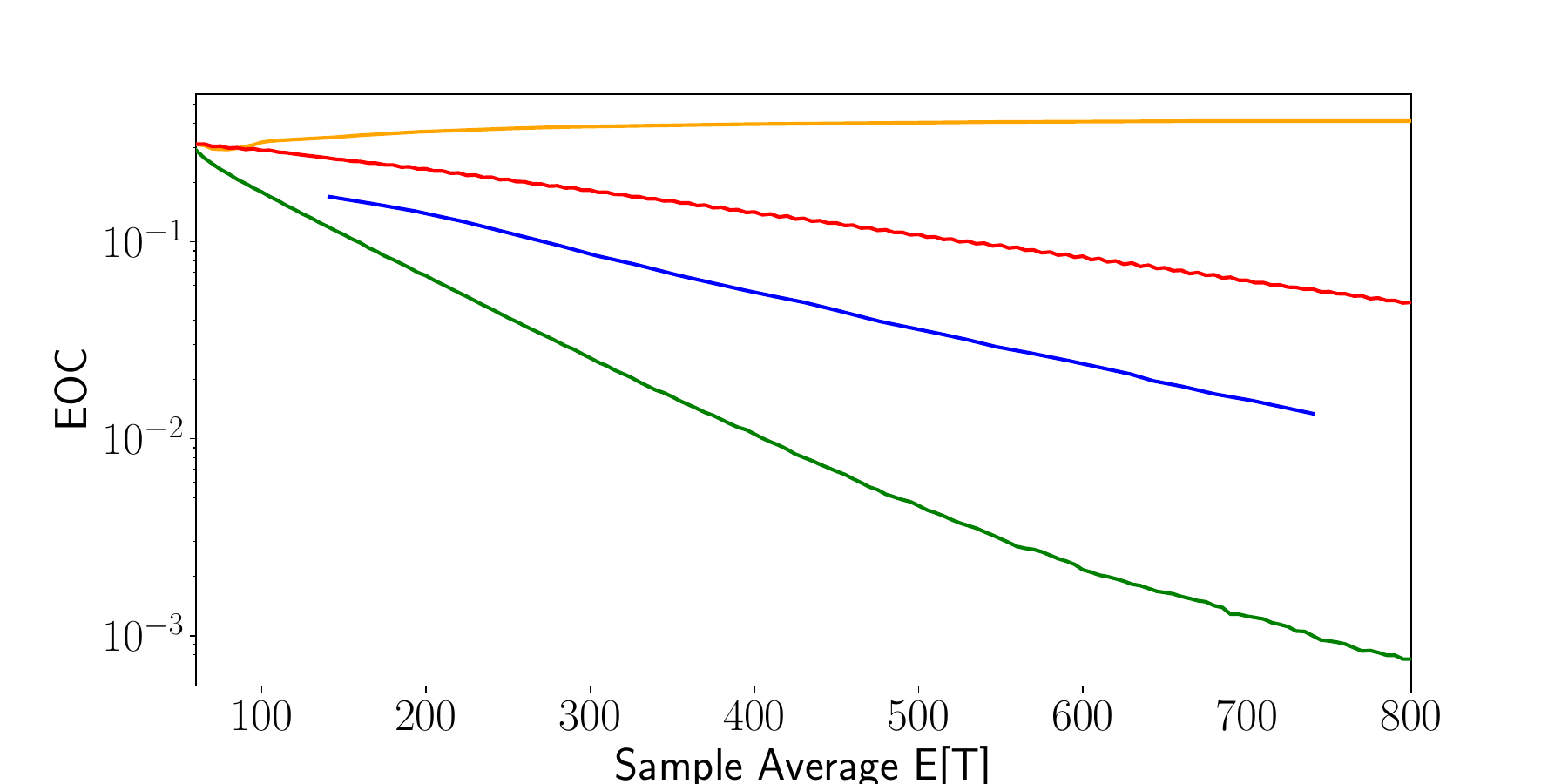}
         \caption{General}
         \label{fig-appendix:sc10_gen_compare}
     \end{subfigure}
     \vspace{0.1em}
    \caption{SC configurations with 10 arms}
    \label{fig-appendix:SC_10arm_compare}
    \vspace{2em}
\end{figure}

\end{document}